\documentclass{article}

\usepackage{arxiv}

\usepackage[utf8]{inputenc} 
\usepackage[T1]{fontenc}    
\usepackage{hyperref}       
\usepackage{url}            
\usepackage{booktabs}       
\usepackage{amsfonts}       
\usepackage{nicefrac}       
\usepackage{microtype}      
\usepackage{lipsum}		
\usepackage{graphicx}
\usepackage{doi}
\usepackage{amssymb}
\usepackage{graphicx}
\usepackage{epsfig}
\usepackage{amsmath}
\usepackage{floatrow}
\usepackage{enumitem}
\usepackage{bm}
\usepackage{mathtools}
\usepackage{algorithm}
\usepackage{multirow}
\usepackage{algpseudocode}
\newcommand{\specialcell}[2][c]{%
\begin{tabular}[#1]{@{}c@{}}#2\end{tabular}}

\title{Spectral Operator Learning for Parametric PDEs Without Data Reliance}

\author{ \href{https://orcid.org/0000-0000-0000-0000}{\includegraphics[scale=0.06]{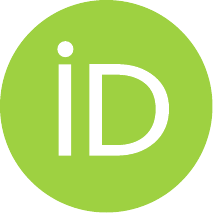}\hspace{1mm}Junho Choi} \\
	Department of Mathematical Sciences\\
	Korea Advanced Institute of Science and Technology\\
	\texttt{junho\_choi@kaist.ac.kr}
	\And
 \href{https://orcid.org/0000-0000-0000-0000}{\includegraphics[scale=0.06]
 {orcid.pdf}\hspace{1mm}Taehyun Yun} \\
	Department of Mechanical Engineering\\
	Gachon University\\
	\texttt{dbsxogus132@gachon.ac.kr}
        \And
	\href{https://orcid.org/0000-0000-0000-0000}{\includegraphics[scale=0.06]{orcid.pdf}\hspace{1mm}Namjung Kim} \\
	Department of Mechanical Engineering\\
	Gachon University\\
	\texttt{namjungk@gachon.ac.kr}
        \And
  \href{https://orcid.org/0000-0000-0000-0000}{\includegraphics[scale=0.06]{orcid.pdf}\hspace{1mm}Youngjoon Hong} \\
	Department of Mathematical Sciences\\
	Korea Advanced Institute of Science and Technology\\
	\texttt{hongyj@kaist.ac.kr}
	}

\renewcommand{\shorttitle}{\textit{} Spectral Operator Learning}

\hypersetup{
pdftitle={Spectral Operator Learning for Parametric PDEs Without Data Reliance},
pdfauthor={Choi. JH, Kim. NJ, Hong. YJ},
pdfkeywords={unsupervised learning, 
deep neural network, 
Legendre-Galerkin approximation,
spectral bias, 
boundary layer, 
singular perturbation.}
}

\begin{document}
\maketitle

\begin{abstract}
In this paper, we introduce the Spectral Coefficient Learning via Operator Network (SCLON), a novel operator learning-based approach for solving parametric partial differential equations (PDEs) without the need for data harnessing. 
The cornerstone of our method is the spectral methodology that employs expansions using orthogonal functions, such as Fourier series and Legendre polynomials, enabling accurate PDE solutions with fewer grid points. By merging the merits of spectral methods – encompassing high accuracy, efficiency, generalization, and the exact fulfillment of boundary conditions – with the prowess of deep neural networks, SCLON offers a transformative strategy. 
Our approach not only eliminates the need for paired input-output training data, which typically requires extensive numerical computations, but also effectively learns and predicts solutions of complex parametric PDEs, ranging from singularly perturbed convection-diffusion equations to the Navier-Stokes equations.
The proposed framework demonstrates superior performance compared to existing scientific machine learning techniques, offering solutions for multiple instances of parametric PDEs without harnessing data.
The mathematical framework is robust and reliable, with a well-developed loss function derived from the weak formulation, ensuring accurate approximation of solutions while exactly satisfying boundary conditions.
The method's efficacy is further illustrated through its ability to accurately predict intricate natural behaviors like the Kolmogorov flow and boundary layers. In essence, our work pioneers a compelling avenue for parametric PDE solutions, serving as a bridge between traditional numerical methodologies and cutting-edge machine learning techniques in the realm of scientific computation.
\end{abstract}


\section{Introduction}
\label{intro}
Achieving a comprehensive understanding of natural phenomena that occur within society and nature is a coveted and ambitious objective that has yet to be fully realized by humankind. The dynamics of natural phenomena, ranging from familiar everyday events such as fluid flows 
\cite{anderson1995computational, wilcox1998turbulence, landau2013fluid,harlow2004fluid},
sound propagation \cite{knj01},
microwave processing \cite{knj02},
weather forecasting \cite{knj03},
geophysics \cite{vallis2016geophysical,pedlosky2013geophysical,salmon1998lectures}
to extraordinary features like optical cloaking \cite{cai2007optical,xu2015conformal,qian2020deep}, Casimir quantum levitation \cite{knj04}, and rogue waves \cite{adcock2011did, onorato2013rogue} can be often comprehended by the mathematical framework, expressed as a system of partial differential equations (PDEs). Their significance extends beyond the realm of physics and finds applications in fields such as {finance \cite{broadie2004anniversary,jeanblanc2009mathematical}, epidemiology \cite{giordano2020modelling,aleta2020modelling}, ecology \cite{ecology}, and computer graphics \cite{vince2006mathematics,aubert2006mathematical}.} Solving PDEs of complex physical systems using traditional tools, such as finite difference method (FDM)\cite{smith1985numerical,strikwerda2004finite}, finite volume method (FVM)\cite{leveque2002finite}, finite element method (FEM)\cite{zienkiewicz2005finite}, and spectral method \cite{shen2011spectral} has gained attention for several decades by its wide usability without restriction in domains, their reasonable accuracy and interpretable mathematical framework. These techniques have been provided one of the major thrusts for understanding nature as well as advancing the state-of-the-art engineering achievements, such as Cahn-Hilliard equation\cite{novick2008cahn},  Navier Stokes equation \cite{temam2001navier,girault2012finite,fefferman2000existence},microfluidic-based bioprinting \cite{knj07}, granular hydrogel matrices for future matter \cite{knj08}, implantable multimodal biosensors \cite{knj09}, and density-variant nanolattices \cite{knj10}.
Despite previous achievements, solving a wide range of PDE systems, as their complexity increases, can present both theoretical and practical difficulties. In addition, the time-related PDEs are generally considered more challenging, as errors tend to accumulate over time when using traditional techniques.

Recently, rapid progress in deep neural networks and statistical learning theory, coupled with the advent of powerful computational resources, has paved the way for tackling intricate PDEs, thereby giving rise to a novel research field known as scientific machine learning. 
Among various novel ideas in scientific machine learning, physics-informed neural networks (PINN) and operator learning algorithms have gained huge attention recently. 
PINNs utilize a neural network to solve PDEs with little or no data by minimizing the PDE residual loss, enforcing physical constraints \cite{PINN009,PINN001}.
It offers several advantages compared to conventional methods, including mesh-free characteristics, relatively low computational cost, differentiable solutions through analytical gradients, and the ability to solve both forward and inverse problems using the same optimization scheme.
These advantageous features have contributed to the success of PINNs in various fields, such as fluid mechanics {\cite{raissi2020hidden,cai2021physics}}, materials science {\cite{lu2020extraction,chen2020physics,goswami2020transfer,huang2020learning}}, biophysics {\cite{kissas2020machine,sahli2020physics}}, finance {\cite{elbrachter2022dnn,han2018solving}}, and chemistry \cite{schweidtmann2021machine}. 
Several related variations have been proposed, including the self-adaptive PINN \cite{Nam01}, the variational hp-VPINN \cite{Nam02}, the conservative PINN (CPINN) \cite{Nam03}, and other collocation-based approaches \cite{Nam04, Nam05} as well as “data-free” neural network, called physics-constrained neural network (PCNN) \cite{Nam06, Nam07}.
Despite the huge success of PINNs, however, one of the significant limitations of the method is its validity on a single instance that includes initial conditions, boundary conditions, and external forcing terms {\cite{PIDON}}.
Consequently, the entire training process must be repeated when there exist changes to the input conditions, making it challenging to generate real-time predictions for varying input data.
In addition, the problems related to stiffness induced bias and the difficulty in predicting unseen parts of the domain from the boundary and initial conditions are also significant challenges of PINNs.

Compared to PINNs, operator learning finds the underlying patterns and relationships in the input-output pairs, allowing for real-time predictions for varying input data. Due to its capacity to acquire the relationships between the conditions of partial differential equations (PDEs), including coefficients, boundary and initial conditions, and the corresponding solutions, this approach facilitates the forecasting of multiple instances and expedites the evaluation process.
Moreover, its function-to-function mapping ensures theoretically infinite resolution of the solution, surpassing the limited resolution of vector-to-vector mapping found in conventional techniques. Several variants have been proposed to extend the capabilities of operator learning so that the remarkable achievements in the comprehension of natural phenomena fluid dynamics {\cite{lin2021operator}}, chemistry \cite{mao2021deepm}, electromagnetics \cite{cai2021deepm}, materials engineering \cite{goswami2022physics,liu2022learning}, and aerodynamics\cite{di2021deeponet}. 
Despite its success, operator learning confronts significant obstacles, including the requirement for substantial volumes of training data, limited generation capability, and the lack of extrapolating to previously unseen conditions {\cite{PIDON}}. 
Even though researchers are currently exploring novel strategies to overcome these challenges, such as the integration of physical principles or constraints {\cite{PIDON,PINO}}, as well as the development of more expressive models \cite{PINN008, Nam01, krishnapriyan2021characterizing}, the overall accuracy of these methods remains suboptimal, primarily due to numerical errors stemming from boundary conditions. Moreover, many of these strategies display inherent instability, especially when subjected to complex or unexpected scenarios. 

To address the aforementioned limitations, this paper proposes a novel approach, rooted in the operator learning paradigm, for solving diverse parametric PDEs. Traditional methods often necessitate a dataset of solutions; however, our approach shifts the paradigm to learning the coefficients directly, embodying the principles of operator learning. We introduce the \textit{Spectral Coefficient Learning via Operator Network} (SCLON) that seamlessly integrates spectral methods to learn the solution operator, obviating the need for exhaustive solution datasets for a range of parametric PDEs.
For the spectral methods in numerical analysis, the solution is traditionally approximated as a linear combination of spectral coefficients, \( \alpha_n \), and their associated spectral basis, \( \phi_n(\mathbf{x}) \). 
The potency of spectral methods emanates from their reliance on global and orthogonal polynomials as the spectral basis:
\begin{equation} \label{e:1}
u_N(\mathbf{x}) = \sum_{n=0}^{N-1} \alpha_n \phi_n(\mathbf{x}), \quad \mathbf{x} \in \mathbb{R}^d,
\end{equation}
where \( u_h \) serves as the approximated solution for the target PDEs. Drawing motivation from \eqref{e:1}, and in the true spirit of operator learning, SCLON is designed to predict numerical solutions to PDEs when furnished with initial conditions, external forcing functions, or PDE coefficients.
The main insight of our approach is in the targeted prediction approach: by focusing solely on predicting the spectral coefficients, SCLON avoids the need for labor-intensive numerical computations to generate massive datasets.
This lean methodology facilitates real-time predictions adaptable to diverse input data. Thus, SCLON not only exemplifies the versatility of operator learning but also emerges as a potent tool, adept at decoding solutions for a plethora of PDEs across varied scenarios.
The loss function of SCLON is designed based on the residual quantity of the spectral approximation, akin to spectral methods \cite{shen2011spectral}.
This design enables SCLON to accurately approximate the solutions of PDEs while ensuring that boundary conditions are exactly met.
For constructing an approximation of PDE solutions, SCLON draws inspiration from the spectral element and Fourier spectral methods.
This approach involves inferring coefficients, denoted as \( \widehat{\alpha}_k \), of the polynomial basis functions \( \phi_k \). 
These coefficients are then used in the linear combination \( \sum\widehat{\alpha}_k\phi_k \) to approximate the solution of PDEs. 
Since each basis function inherently satisfies the boundary condition, the predicted solution also adheres to the exact boundary condition. Furthermore, given that the spectral method underpins our framework as shown in \eqref{e:1}, the predicted numerical solution is anticipated to exhibit relatively smaller errors compared to other machine learning-based methodologies, such as Physics-Informed Neural Operator \cite{PINO} and Physics-Informed DeepONet \cite{PIDON}.
Importantly, owing to the inherent structure of the proposed SCLON scheme, there is no need for paired input-output training data. This is because we employ the weak residual as a loss, allowing the model to train in an unsupervised fashion.

The main contribution of this work is unsupervised, model-agnostic Spectral method-based framework that efficiently learns and predicts parametric PDEs including complex natural phenomena such as the Kolmogorov flow, Kuramoto–Sivashinsky equation, and boundary layer problems. The proposed spectral-based deep neural network model exhibits a multitude of advantageous characteristics, including reduced data dependency, computational efficiency, resilience, and straightforward implementation.
Another pivotal contribution of this study is the introduction of a novel learning architecture, specifically designed to accurately solve convection-dominated, singularly perturbed problems characterized by pronounced boundary layer phenomena.
Such problems present considerable challenges for both traditional numerical methods and scientific machine learning approaches, given the sharp transitions within thin layers induced by minuscule diffusive parameters. By leveraging theory-guided methods \cite{hong2014numerical}, the network adeptly captures the behavior of boundary layers—a domain that recent machine learning approaches have often overlooked.
Moreover, through benchmark experiments involving a spectrum of parametric PDEs — from linear to nonlinear, and parabolic to hyperbolic equations — we validate the efficacy of SCLON. 
These experiments demonstrate that our approach consistently surpasses existing scientific machine learning techniques in accuracy. Interestingly, SCLON attains these outcomes without any dependency on datasets, emphasizing its capability to address a wide array of parametric PDEs in an 
 unsupervised manner. 

\section{Time sequential approaches for time-dependent parametric PDEs}

This section presents our proposed method, SCLON, for solving time-dependent parametric PDEs. 
We start by providing a concise overview of spectral methods, which form the foundation of our approach. 
Subsequently, we introduce how the SCLON solves time-dependent parametric PDEs through a combination of spectral methods and coefficient learning techniques. Lastly, we provide a detailed description of the sequential method for handling time-dependent PDEs. It is important to note that this section presents general concepts and algorithms related to SCLON. For more specific explanations tailored to different types of PDEs, please refer to sections \ref{sec_DRE} through \ref{sec_NE}.
\subsection{Spectral methods}\label{sec_spectral}
We consider a generic form of time-dependent PDE:
\begin{align}\label{pde}
   u_t+\mathcal{N}(u)&=f,\quad (t,{\bf x})\in(0,T]\times \Omega\\
   u({\bf x},0) &=u_0({\bf x}),\quad t=0,\quad {\bf x}\in \Omega,
\end{align}
where $\mathcal{N}$ represents a linear or nonlinear differential operator and $f$ denotes an external forcing function. In this context, boundary data are provided, which can take forms such as Dirichlet, Neumann, or periodic boundary conditions.
To derive a weak formulation in a specified Hilbert space \( H \), we multiply equation \eqref{pde} by \( \phi \in H \) and integrate over \( \Omega \):
\begin{align}\label{weak_solution}
    \int_\Omega (u_t+\mathcal{N}(u))\phi \, dx = \int_\Omega f\phi \, dx,
\end{align}
for \( 0 \leq t \leq T \). 
By performing integration by parts on equation \eqref{weak_solution}, we obtain a weak formulation for the given PDE. Note that we provide a formal definition of the weak formulation for each equation in Sections \ref{sec_DRE} and \ref{sec_DCE}.
Thanks to the Galerkin method, a weak solution $u$ to \eqref{weak_solution} can be approximated as $\sum_{n=0}^{N-1}\alpha_n\phi_n,$
where $\{\phi_n\}_{n=0}^{N-1}$ is a finite set of basis or trial functions of $H$, and $N$ is a finite integer. 
In numerical analysis, depending on the choice of basis functions, various classes of numerical methods have been developed such as finite element, finite volume, and spectral methods. 
In the subsequent discussions, we will focus on spectral methods where the $\phi_n$ are global polynomials, typically of the Legendre, Chebyshev, or Fourier types.
While spectral methods were developed more recently than finite difference schemes and finite element schemes, they have proven their ability to tackle a wide variety of problems. Most importantly, they allow for high accuracy with only moderate computational resources. For more details, see, e.g., \cite{shen2011spectral,evans2022partial,trefethen2000spectral}.
In this paper, we employ the spectral method to approximate the weak solution $u$ in $H$ as
\begin{align}\label{sm}
    u_N(x) =\sum^{N-1}_{n=0} \alpha_n\phi_n(x),
\end{align} 
where $N$ is the number of the basis function on $\Omega$. In this work, to determine $\{\alpha_n\}$, we employ either the Legendre spectral element method (LS) or the Fourier spectral element method (FS) depending on prescribed boundary conditions. 

When imposing Dirichlet or Neumann conditions, we utilize the LS that employs a linear combination of Legendre polynomials as basis functions for solving \eqref{weak_solution}. This choice is motivated by the ability of Legendre Polynomials to represent these conditions as follows:
\begin{align}\label{Legendre_poly}
    \phi_n(x) = L_n(x) + a_n L_{n+1}(x) + b_n L_{n+2}(x),
\end{align}
where \( L_n \) represents the Legendre polynomials of degree \( n \), and the coefficients \( a_n \) and \( b_n \) are determined by the boundary conditions. 
Besides the upside to represent the boundary conditions, due to the mutual orthogonality among \( L_n(x) \), the stiffness and mass matrices generated by \eqref{Legendre_poly} are sparse.

In order to consider time-dependent problems, we combine time marching numerical methods with LS. Let $T$ be a final time and let $\Delta t > 0$ be a small time step such that $T/\Delta t$ is an integer, denoted by $R$.
Afterwards, we denote solutions for $r$-th time step as $\alpha^r_n := \alpha_n(r\Delta t)$ for an integer $r \geq 0$, and hence, $u^r_N:= u_N(r\Delta t) = \sum^{N-1}_{n=0} \alpha_n^r \phi_n(x).$ Then, time marching numerical methods enable us to first find \( u^1_N \) from a given initial condition \( u_0 \), and to recursively compute \( u^{r}_N \) from \( u^{r-1}_N \) for \( r \geq 1 \). 
More precisely, using \eqref{sm} and \eqref{Legendre_poly}, the weak form \eqref{weak_solution} can be approximated by
\begin{align}\label{weak_approx00}
   \int_\Omega (u_N)_t\phi_n +\mathcal{N}({u}_N)\phi_n  dx=\int_\Omega f\phi_n dx,
\end{align}
for $0 < t < T$.
In the context of time discretization, given \( u^{r-1}_N \) for an integer \( r\geq 1 \), time marching methods such as the Euler method or the Runge-Kutta method can be applied. Through these methods, the weak formulation \eqref{weak_approx00} is transformed into a linear system for the vector \( \{\alpha^r_n\}_{n=0}^{N-1} \). Once the vector \( \{\alpha^r_n\}_{n=0}^{N-1} \) is determined from the linear system, \( u^r_N \) can be obtained by a linear combination of \( \{\phi_n\}_{n=0}^{N-1} \) and \( \{\alpha^r_n\}_{n=0}^{N-1} \) as described by equation \eqref{sm}. Consequently, \( u^r_n \) can be recursively computed from \( u^{r-1}_N \) for \( r\geq 1 \).


The SCLON is motivated by the framework of the LS. 
Given an input function such as an initial condition, variable coefficient, or forcing term (see Table \ref{tab:1} to \ref{tab:3}), our neural network infers the spectral basis coefficients $\{\widehat{\alpha}_n^r\}_{n=0}^{N-1}$ for $1\leq r \leq R$.
Subsequently, the approximate solutions ${\widehat{u}^r_N}$ are reconstructed with $\{\widehat{\alpha}_n^r\}_{n=0}^{N-1}$ by \eqref{sm}. 
To facilitate the learning process, we define a loss function using \eqref{weak_approx00} for the LS:
\begin{align}\label{loss}
    loss(\{\widehat{u}^r\}_{r=1}^{R})&=\sum_{r=1}^{R}\sum_{n=0}^{N-1}\left|\int_\Omega \widehat{u}_t\phi_n dx+\int_\Omega \mathcal{N}(\widehat{u})\phi_n  dx-\int_\Omega f\phi_n dx\right|^2.
\end{align}
For the time integration in \eqref{loss}, we utilize the same time marching technique as that employed for computing \( u^r_N \) earlier in numerical computations. 
With the reduction in loss, the SCLON undergoes updates, steadily refining the neural network's predicted solution towards the true one. Comprehensive insights into the SCOLON with LS for particular PDEs are provided in Sections \ref{sec_DRE} and \ref{sec_DCE}.

On the other hand, the FS is employed in cases where a periodic boundary condition is imposed; see, for example, \cite{trefethen2000spectral}.
 Let $x_n=nh$ be a nodal point on $[0,2\pi)$ where $h=\frac{2\pi}{N}$ for $0\leq n\leq N-1$, and let $\mathrm{i}$ be the imaginary unit.
In addition, discrete Fourier transform (DFT) is defined by 
\begin{align} \label{FTT}
\mathcal{F}_{\xi}(u)=h\sum_ {n=0}^{N-1}e^{-\mathrm{i}\xi x_n}u(x_n),\quad \xi=-\frac{N}{2}+1\cdots,\frac{N}{2}, 
\end{align}
for a function $u$, and inverse discrete Fourier transform (IDFT) is defined by 
\begin{align}
\mathcal{F}^{-1}_n(\alpha_\xi)=\frac{1}{2\pi}\sum_ {\xi=-N/2+1}^{N/2}e^{\mathrm{i}\xi x_n}\alpha_\xi,\quad n=0,\cdots,N-1,
\end{align}
where $\alpha_\xi$ are complex values. Now, by utilizing Fourier functions as basis functions, an approximate solution to \eqref{weak_solution} in $H$ is set to 
\begin{align}\label{Fsm}
u_N(x_n)
:=\frac{1}{2\pi}\sum_ {\xi=-N/2+1}^{N/2}e^{\mathrm{i}\xi x_n}\alpha_\xi,\quad n=0,\cdots,N-1.
\end{align}
Subsequently, once substituting \eqref{Fsm} to \eqref{weak_solution} with $\phi=e^{-\mathrm{i}\xi x_n}$ and then applying the rectangular quadrature rule, it turns into
\begin{align}\label{weak_approx1}
    \mathcal{F}_\xi(u_N)_t +\mathcal{F}_\xi(\mathcal{N}({u}_N))=\mathcal{F}_\xi (f).
\end{align}
Let us define \( \alpha^r_\xi := \alpha_\xi(t=r\Delta t) \) for an integer \( r \geq 0 \). This definition is used to construct \( u^r_N:= u_N(t=r\Delta t) \) for the given time step, as in \eqref{Fsm}. Thus, \eqref{weak_approx1} can be transformed into a linear system for \( \{\alpha^r_\xi\}_{\xi=-N/2+1}^{N/2} \), from which \( \{\alpha^r_\xi\}_{\xi=-N/2+1}^{N/2} \) can be sequentially determined using \( \{\alpha^{r-1}_\xi\}_{\xi=-N/2+1}^{N/2} \) where \( r \geq 1 \).

We now introduce the SCLON, inspired by the FS approach. Given input functions such as initial conditions, variable coefficients, or external forcing terms (refer to Tables \ref{tab:1}-\ref{tab:3}), our neural network deduces the spectral basis coefficients at \( t = r\Delta t \), represented as \(\{\widehat{\alpha}^r_\xi\}_{\xi=-\frac{N}{2}+1}^{\frac{N}{2}}\) within the FS framework. Based on this, we reconstruct the predicted solutions \({\widehat{u}^r_N}\) for \(1\leq r\leq R\) using \eqref{sm}. To streamline the learning, we formulate an FS-specific loss function as follows:
\begin{align}\label{loss_FS}
    loss(\{\widehat{u}^r\}_{r=1}^{R})&=\sum_{r=1}^{R}\sum_{\xi=-N/2+1}^{N/2}\left|\mathcal{F}_\xi(\widehat{u})_t +\mathcal{F}_\xi(\mathcal{N}(\widehat{u}))-\mathcal{F}_\xi (f)\right|^2 .
\end{align}
For the time method in \eqref{loss_FS}, we make use of the same method as employed earlier for computing $u^r_N$. 
As the loss diminishes, the SCLON undergoes updates, refining the neural network's approximate solution to more closely align with the true solution.
Further insights into the SCLON's application for specific PDEs can be found in Sections \ref{sec_VBE}, \ref{sec_CE}, \ref{sec_KS}, and \ref{sec_NE}.

\subsection{Sequential method}\label{sec_sequential}
Conventional numerical schemes for time-dependent problems often employ a time-sequential method, solving the problems step by step starting from initial conditions. In contrast, modern scientific machine learning approaches, such as PINN \cite{PINN001, wang2022and, kharazmi2019variational,PINN009}, PIDoN \cite{PIDON}, and PINO \cite{PINO}, do not utilize time-sequential methods.
Instead, they employ methods that train over the entire temporal domain at once. This often leads to neural networks failing to predict accurate solutions \cite{PINN008,mattey2022novel,wight2020solving,krishnapriyan2021characterizing}. 
If the former prediction of the neural network for time \( t \) is not fully accurate, training the networks with this inaccurate data can lead to the even more inaccurate later prediction for time \( t + \Delta t \). As a result, while the predictions for \( t \) close to \( 0 \) might be relatively close to the exact solution, those for \( t \) significantly distant from \( 0 \) deviate substantially from the exact solutions.

To reduce errors that grow larger as time steps evolve, we propose the framework of time-sequential method for training SCLON, that is, fragments a single network into multiple networks, with each being accountable for a distinct time segment \cite{PINN008, SPINN}. To refine the process, we partition the time domain $[0, T]$ into $Q$ segments, expressed as 
\[ [t_0=0,t_1],[t_{1},t_2],\cdots,[t_{q-1},t_q],\cdots,[t_{Q-1},t_{Q}=T].\]
Corresponding to the $q$-th segment, we establish the $q$-th network, represented by $\mathcal{G}_1$, $\mathcal{G}_2$, $\cdots$, $\mathcal{G}_q$,$\cdots$, $\mathcal{G}_{Q}$. 
Starting with the initial condition $u_0$, the network $\mathcal{G}_1$ is trained to approximate the solution $u$ over the interval $(t_0,t_1]$. 
This training proceeds until the loss functions, either \eqref{loss} or \eqref{loss_FS}, plateau during the training phase.
Then, an approximation \(\widehat{u}\) of the solution at \(t = t_1\) is obtained. 
Subsequently, \(\mathcal{G}_2\) is initialized with \(\widehat{u}(t_{1})\) and is trained to determine the approximate solutions over the interval \((t_1, t_2]\). This iterative process extends until \(\mathcal{G}_Q\), initialized with \(\widehat{u}(t_{Q-1})\), has been adequately trained over the interval \((t_{Q-1}, T]\). Consequently, by linking all \(Q\) networks in sequence, the sequential method ensures coverage across the entire time domain \([0, T]\).

\begin{figure*}[ht] 
\vskip 0.2in
\begin{center}
\centerline{\includegraphics[width=1.0\columnwidth]{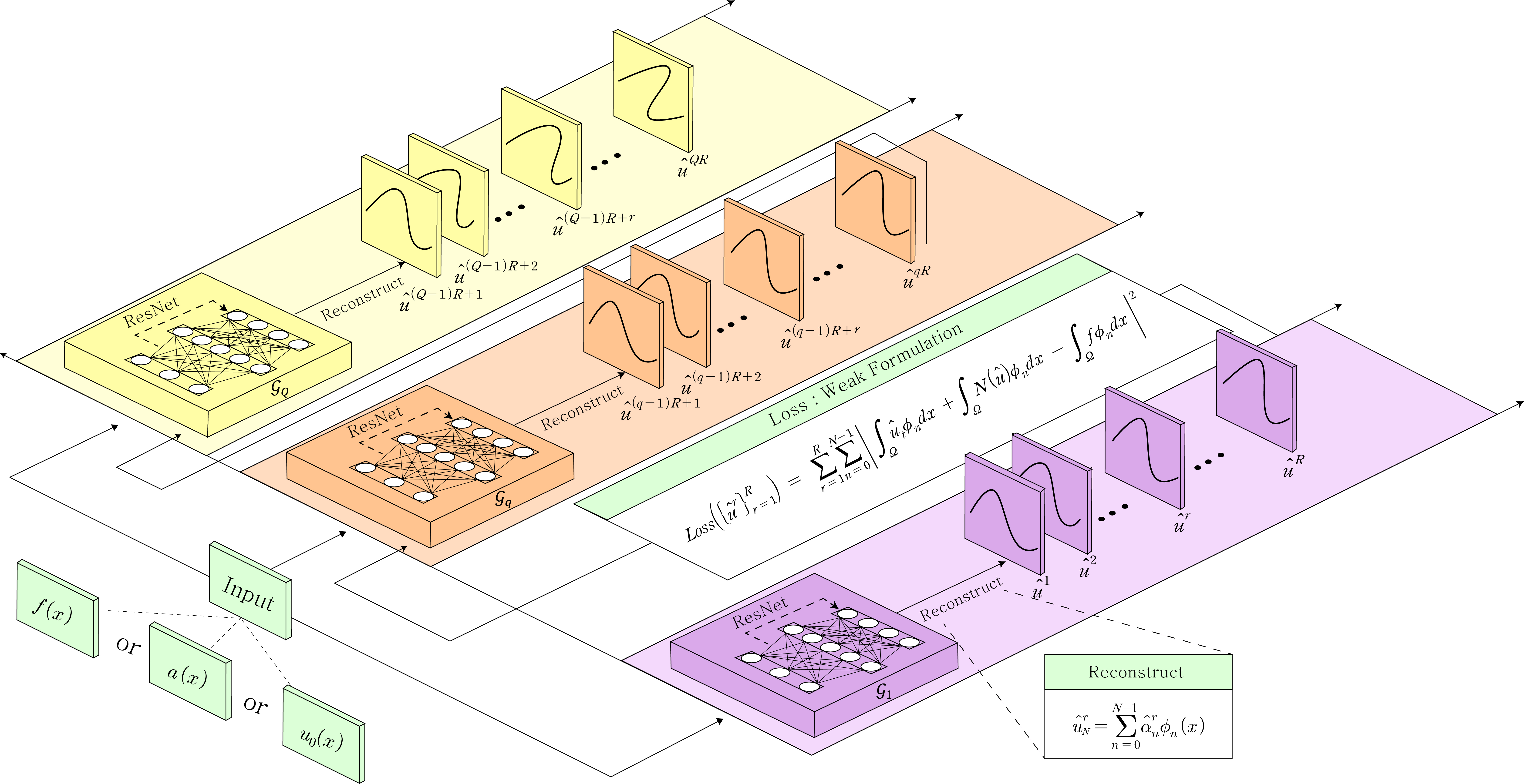}}
\caption{
\textbf{Schematic Diagram of the Structure of SCLON.}
The structure of SCLON consists of two parts: one part emulates the spectral method combined with a numerical method for time marching (refer to section \ref{sec_spectral}), while the other utilizes a sequential method (see section \ref{sec_sequential} and algorithm \ref{alg:cap}). Given input data—such as external forcing functions, variable coefficients, or initial conditions—to the network $\mathcal{G}_q$, it deduces a set of coefficients 
{{$\Big[\{\widehat{\alpha}_n^r\}_{r=(q-1)R+1}^{qR}\Big]_{n=0}^{N-1}$}}, constructing approximations: $\widehat{u}_N^{r}=\sum_{n=0}^{N-1}\widehat{\alpha}^{r}_n\phi_n(x)$ for $r=(q-1)R+1,\ldots,qR$. 
By allotting each network $\mathcal{G}_q$ for $q=1,\ldots,Q$ to a specific time segment $(t_{q-1},t_q]$, the approximations in the time direction are sequentially computed.}
\label{f:schematic}
\end{center}
\vskip -0.2in
\end{figure*}

We note that one of the advantages of the sequential method is that $\mathcal{G}_{q}$ starts to be trained with the input condition, $\widehat{u}$ at $t=t_{q-1}$ predicted by $\mathcal{G}_{q-1}$. 
As $\mathcal{G}_{q-1}$ becomes better trained, $\widehat{u}$ at $t=t_{q-1}$ approaches $u$ at $t=t_{q-1}$. Consequently, the subsequent prediction $\widehat{u}$ at $t=t_{q-1}+\Delta t$ becomes more accurate. Moreover, because the loss functions on each segments are smaller, it becomes much easier to find a local minimum than the loss functions on the whole domain. In addition, there is freedom to choose the size of segments, $t_q - t_{q-1}$. The choice of the size depends on tolerance about error. As more accurate approximations are required, a smaller size of segments with more segments should be set, which may result in a trade-off involving longer computational time. 
However, once the parameters of the network \( \mathcal{G}_{q-1} \) are trained, they can be used as the initial parameters for \( \mathcal{G}_{q-1} \). This approach not only saves training time but also enhances the performance of the neural network.
The architecture of the sequential approach is illustrated in Algorithm \ref{alg:cap} and Figure \ref{f:schematic}.


\begin{algorithm}
\caption{Training process of the SCOLON}\label{alg:cap}
\begin{algorithmic}[tb]
\State Refer to the notations of hyper-parameters in Appendix \ref{notation}.
\State {\bfseries Input:} 
Sample $f(x) \sim$ GRF.
\For{$q=1$ to $Q$} 
\State {\bfseries Step 1:} Take a segment $(t_{q-1},t_q]$ from $[0,T]$.
\State {\bfseries Step 2:} Create a $q$-th network for $(t_{q-1},t_q]$
 such that
   \begin{align}
       \mathcal{G}_q:(f,\theta)\longmapsto \Big[\{\widehat{\alpha}_n^r\}_{r=(q-1)R+1}^{qR}\Big]_{n=0}^{N-1}.
   \end{align}
\State {\bfseries Step 3:} Construct predictions for $1\leq r \leq R$ as
    \begin{align}\label{network}
       \widehat{u}^{(q-1)R+r}_N&:\Big[\widehat{\alpha}_n^{(q-1)R+r}\Big]_{n=0}^{N-1}\longmapsto\sum_{n=0}^{N-1}\widehat{\alpha}^{(q-1)R+r}_n\phi_n(x).
   \end{align}
 \State {\bfseries Step 4:} Define a loss function, $loss_q(\{u^r(\mathcal{G}^r_q(f,\theta))\}_{r=(q-1)R+1}^{qR})$ as in \eqref{loss} for the LS or \eqref{loss_FS} for the FS. Here, $\mathcal{G}^r_q(f,\theta)$ stands for the components in the $r$th row of $\mathcal{G}_q(f,\theta)$.

   
\State {\bfseries Step 5:} Train $\mathcal{G}_q$ to make $\widehat{u}^{(q-1)R+r}_N$ closer to the solution $u^{(q-1)R+r}$ at $t=((q-1)R+r)\Delta t$ by minimizing the loss function.



\EndFor
\end{algorithmic}
\end{algorithm}

\begin{table}[h!] 
\vskip 0.15in
\begin{center}
\begin{small}
\begin{tabular}{||cc|ccc|ccc||}
\hline
\multirow{3}{*}{Equation}&\multirow{3}{*}{Random input}&\multicolumn{6}{c||}{Test error}\\\cline{3-8}
&&\multicolumn{3}{c|}{SCLON (ours)} & \multicolumn{3}{c||}{PIDoN} \\
\cline{3-8}
      && MAE&  Rel.$L^2$& $L^\infty$ & MAE& Rel.$L^2$& $L^\infty$ \\
\hline
 \specialcell{Diffusion\\
 reaction\eqref{DRE}} & \specialcell{Forcing\\ functions} & 1.394e-03 & 5.150e-03& 4.894e-03& 2.232e-03
 & 7.060e-03& 2.943e-03 \\ 
 \hline
 Advection \eqref{CE} & \specialcell{Variable\\ coefficients} &9.472e-04  & 2.177e-03& 2.465e-03 &N/A& N/A&N/A \\
  \hline
 \specialcell{Convection diffusion\\with a boundary layer\eqref{DCE}} & \specialcell{Initial\\ conditions} &  6.349e-04 & 2.745e-03& 2.771e-03&1.155e-01& 5.359e-01&6.633e-02 \\
 \hline
 \specialcell{2D Kuramoto\\
 Sivashinsky \eqref{KSE}} & \specialcell{Initial\\ conditions} & 4.48e-03
 & 1.19e-02 & 1.52e-02 &N/A& N/A&N/A \\
 \hline
\end{tabular} 
\end{small}
\end{center}
\vskip 0.1in
\caption{Comparison of test errors between SCLON and Physics-informed DeepONet (PIDoN) \cite{PIDON} across various types of parametric differential equations. The test errors were measured in MAE, relative $L^2$, and $L^\infty$ errors (see definition of the metrics at Appendix \ref{metrics}), and subsequently averaged over new, unseen datasets. Notably, the results indicate that SCLON predicts solutions with greater accuracy than PIDoN. }
 \label{tab:1}
\end{table}

\begin{table}[h!] 
\vskip 0.15in
\begin{center}
\begin{small}
\begin{tabular}{||cc|ccc|ccc||}
\hline
\multirow{3}{*}{Equation}&\multirow{3}{*}{Random input}&\multicolumn{6}{c||}{Error}\\\cline{3-8}
&&\multicolumn{3}{c|}{Ours} & \multicolumn{3}{c||}{PINO} \\
\cline{3-8}
      && MAE&  Rel.$L^2$& $L^\infty$ & MAE& Rel.$L^2$& $L^\infty$ \\
\hline
 {\specialcell{2D Navier-Stokes \eqref{NSE}}}& {\specialcell{Initial\\ conditions}} & 6.349e-03& 2.225e-02& 2.974e-02& 
 1.434e-01& 4.757e-01&3.111e-01  \\
 \hline
\end{tabular} 
\end{small}
\end{center}
\vskip 0.1in
\caption{Comparison between SCLON and Physics Informed Neural Operator (PINO) \cite{PINO}. The test errors were measuered in MAE, Rel.$L^2$, and $L^\infty$ (see definition of the metrics at Appendix \ref{metrics}), and then averaged over 1,000 new, unseen data. Notably, the results show that SCLON can predict the solution more accurately compared to PINO.}
 \label{tab:2}
\end{table}

\section{Diffusion-reaction equation}\label{sec_DRE}
In this section, we present a practical example that illustrates the design of a network that emulates the LS for solving nonlinear parametric PDEs.
We begin with the nonlinear diffusion-reaction equation, subject to the homogeneous Dirichlet boundary condition:
\begin{align}\label{DRE}
\begin{split}
&u_t-\nu u_{xx}+\mu u^2=f,\quad \text{for}\quad t>0,x\in (-1,1)=:\Omega,\\ 
&u=u_0,\quad \text{for}\quad t=0,\quad x\in \Omega,\\
&u(\pm 1,t)=0, \quad t>0,
\end{split}
\end{align}
where $f$ is an external forcing term.
In our numerical simulations, we adopt $u_0=0$ and parameters \(\nu=0.01\) and \(\mu=-0.01\), as referenced in \cite{PIDON}. Our objective in this example is to train a neural operator that maps forcing terms \(f(x)\) to the corresponding PDE solutions \(u(x,t)\). To achieve this, we begin by detailing the numerical methods used to solve \eqref{DRE}, employing the LS in space and the implicit Euler method in time. We define the weak formulation of \eqref{DRE} as: find $u \in H^1_0(\Omega)$ such that
\begin{align}\label{DRE_weak0}
    \int_{\Omega}u_t\phi +\nu u_x\phi_x +\mu u^2\phi \, dx = \int_{\Omega}f\phi \, dx,
\end{align}
for all \(\phi \in H^1_0(\Omega)\).
To enforce the Dirichlet boundary condition in \eqref{Legendre_poly}, we adopt the basis function given by \cite{shen2011spectral}:
\begin{align}\label{Le_bases}
    \phi_n(x) = L_{n}(x) - L_{n+2}(x),\quad x \in \Omega,
\end{align}
where \( L_n \) denotes the Legendre polynomial of order \( n \).
The main idea of the LS is to approximate the weak solution $u^{r}:=u(t=r\Delta t)$ of \eqref{DRE_weak0} by 
\begin{align}\label{sm2}
  u_N^{r}:=\sum_{n=0}^{N-1}\alpha^r_n\phi_n    
\end{align}
where $N$ is the number of basis functions. 
Accordingly, after substituting \eqref{sm2} for $u^r$, the implicit Euler method transforms \eqref{DRE_weak0} into  
\begin{align} \label{e:DRE_weak}
    \int_{\Omega}\frac{u_N^{r}-u_N^{r-1}}{\Delta t}\phi_n dx+\nu (u_N^{r})_x(\phi_n)_x dx+\mu(u_N^{r-1})^2\phi_n-\int_{\Omega} f\phi_n dx= 0,
\end{align}
for $n=0,\cdots,N-1$ and $r\geq 1$.
The spatial integration in \eqref{e:DRE_weak} can be approximated using the Gauss-Lobatto quadrature rule, that is, if a function $v$ is integrable on $\Omega$, then
\begin{equation}
    \int_\Omega v  dx \approx \sum_{n=0}^{N+1} w_n v(x_n),
\end{equation}
where \( \{x_n\}_{n=0}^{N+1} \) are the Gauss-Lobatto points, and \( \{w_n\}_{n=0}^{N+1} \) are the corresponding Gauss-Lobatto weights \cite{golub1969calculation}. Consequently, \eqref{e:DRE_weak} becomes a linear system for the vector $\{\alpha^r_n\}_{n=0}^{N-1}$. Once solving the system, $\{\alpha^r_n\}_{n=0}^{N-1}$ is combined with $\{\phi_n\}_{n=0}^{N-1}$ as \eqref{sm2} to construct $u^r_N$. When repeating the process from $t=0$ to $R\Delta t$ for an integer $R>1$, one can obtain the numerical solution for $0\leq t\leq R\Delta t$. We note that the numerical solutions computed through the procedure above were employed as benchmark solutions to measure numerical errors.  

Based on the framework of LS above, we now describe the design of SCOLN. In order to apply the sequential method, upon segmenting the time domain \([0,1]\) into \(Q\) equal intervals, represented as $$
[t_0=0,t_1=1/Q],\ldots,[t_{q-1} = (q-1)/Q, t_q = q/Q],\cdots ,[t_{Q-1}=(Q-1)/Q,t_{Q}=1].$$
Accordingly, we introduce distinct networks, denoted \(\mathcal{G}_q\), for each \(q = 1, 2, \ldots, Q\). Each of these networks is trained over its corresponding time segment. For given a fixed time step size \(\Delta t > 0\), the number of time steps within each segment \([t_{q-1}, t_q ]\) is \(1/(\Delta t \times Q)\), represented as the positive integer \(R\).
In the experiment, we set $\Delta t=0.01$, $Q=10$, $R=10$. Then, we start training $\mathcal{G}_1$ for $(0,t_1]$. Given an input function \(f\) drawn from Gaussian random fields (GRF) with \(\mathcal{N}(0,25^2)\), the network \(\mathcal{G}_1\) produces a set $\Big[\{\widehat{\alpha}_n^r\}_{r=1}^{R}\Big]_{n=0}^{N-1}$, which reconstructs the predicted solutions for $1\leq r\leq R$ as 
\[
\widehat{u}_N^{r} = \sum_{n=0}^{N-1} \widehat{\alpha}^r_n \phi_n.
\]
After that, we define the loss function using the weak formulation \eqref{e:DRE_weak}
where $\widehat{u}_N^{0}$ employs $u^0_N$ computed through \eqref{sm2} for given $u_0$ as
\begin{align}\label{DRE_loss2}
    loss_{1}=\sum_{r=1}^{R}\sum_{n=0}^{N-1}\left|\int_{\Omega}\frac{\widehat{u}_N^{r}-\widehat{u}_N^{r-1}}{\Delta t}\phi_n dx+\nu \int_{\Omega}(\widehat{u}_N^{r})_x(\phi_n)_x dx + \mu\int_{\Omega} (\widehat{u}_N^{r-1})^2\phi_n - f\phi_n dx\right|^2.
\end{align}
We keep training the network \( \mathcal{G}_1 \) until \( loss_1 \) plateaus. As the loss diminishes, the neural network gets updated, and as a result, the prediction \(\widehat{u}_N^{r}\) converges towards $u^r_N$ on $(0,t_1]$. When the training finishes, we begin training $\mathcal{G}_2$ with the initial condition $\widehat{u}_N^{R}$ generated by $\mathcal{G}_1$. Thus, in the same manner, $\mathcal{G}_q$ for $q\geq 3$ is sequentially trained with the initial condition $\widehat{u}_N^{(q-1)R}$ after $\mathcal{G}_{q-1}$ has finished training. In addition, the loss function for $q$ is defined as
\begin{align}\label{DRE_loss2}
    loss_{q}=\sum_{r=(q-1)R}^{qR-1}\sum_{n=0}^{N-1}\left|\int_{\Omega}\frac{\widehat{u}_N^{r+1}-\widehat{u}_N^{r}}{\Delta t}\phi_n dx+\nu \int_{\Omega}(\widehat{u}_N^{r+1})_x(\phi_n)_x dx + \mu\int_{\Omega} (\widehat{u}_N^{r})^2\phi_n - f\phi_n dx\right|^2.
\end{align}
Therefore, $\mathcal{G}_q$ and $loss_q$ for $1\leq q \leq Q$ can cover the time domain $[0,1]$. 
As a consequence, considering $P$ input points on $f_p$ where $1 \leq p \leq P$, the losses become
\begin{align}\label{DRE_loss3}
    loss_{q}=\sum_{p=1}^{P} \sum_{r=(q-1)R}^{qR-1}\sum_{n=0}^{N-1}\left|\int_{\Omega}\frac{\widehat{u}^{r+1}_{N,p}-\widehat{u}^{r}_{N,p}}{\Delta t}\phi_n dx+\nu \int_{\Omega}(\widehat{u}_{N,p}^{r+1})_x(\phi_n)_x dx+\mu\int_{\Omega} (\widehat{u}_{N,p}^{r})^2\phi_n-f_p\phi_n dx\right|^2.
\end{align}
 This sequential process continues until \( q \) reaches \( Q \).

\begin{figure}[h!]
\begin{center}
\setlength{\tabcolsep}{0.00001pt}
\makebox[\textwidth][c]{\begin{tabular}{ ccc }
(a) Exact $u(x,t)$& (b) Predicted $\hat{u}(x,t)$& (c) Absolute error\\
\resizebox{0.35\columnwidth}{!}{\includegraphics{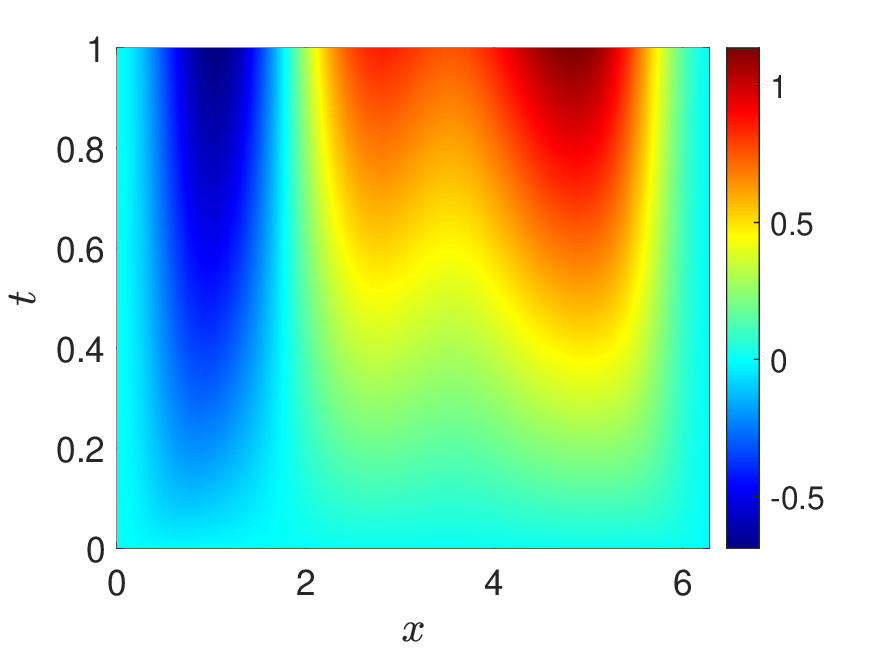}}&\resizebox{0.35\columnwidth}{!}{\includegraphics{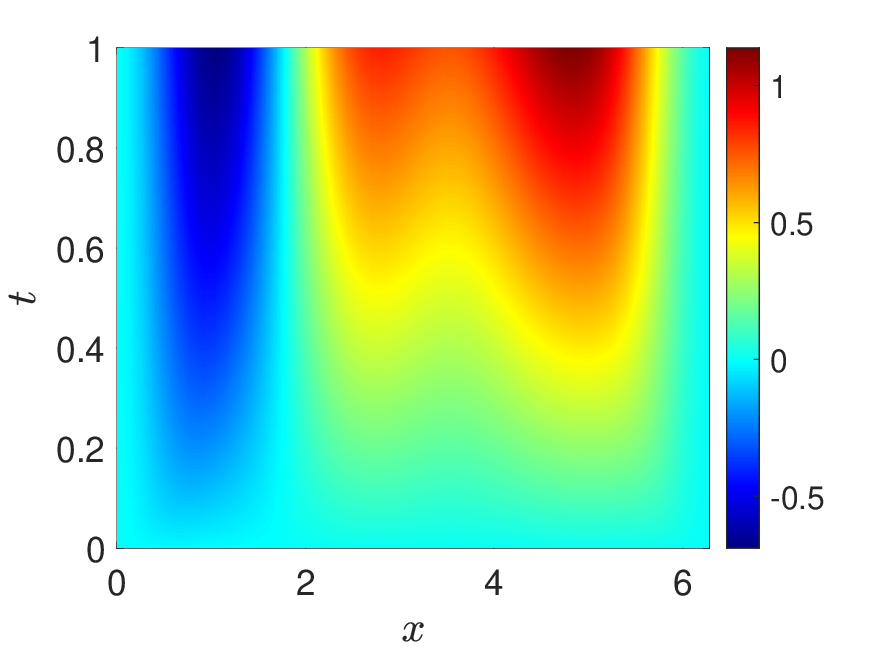}}&\resizebox{0.35\columnwidth}{!}{\includegraphics{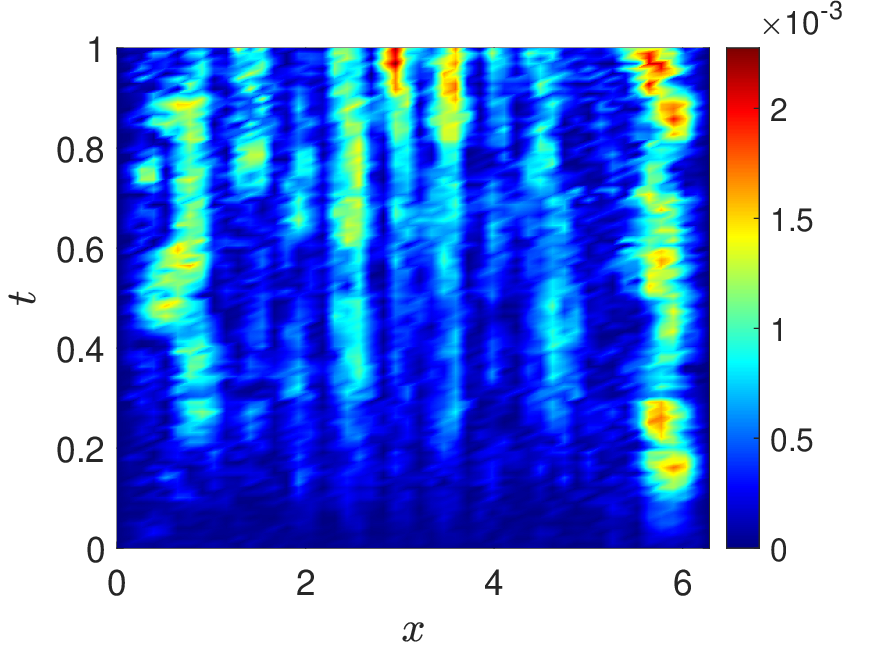}}\\
(d) $$ Exact vs Ours vs PIDoN$$ &&\\
for $t=0.25$& (e) for $t=0.5$& (f) for $t=1$\\
\resizebox{0.35\columnwidth}{!}{\includegraphics{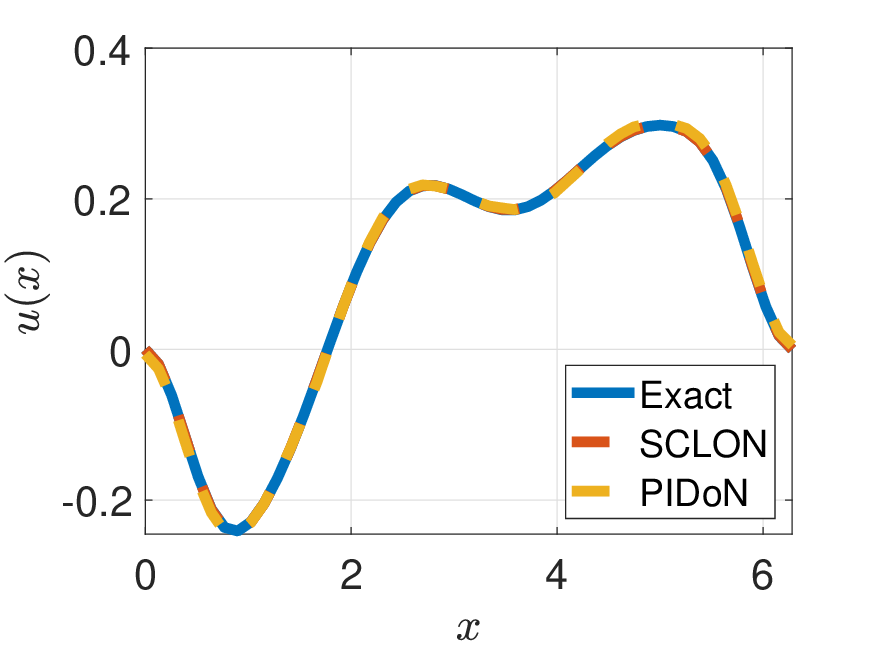}}&\resizebox{0.35\columnwidth}{!}{\includegraphics{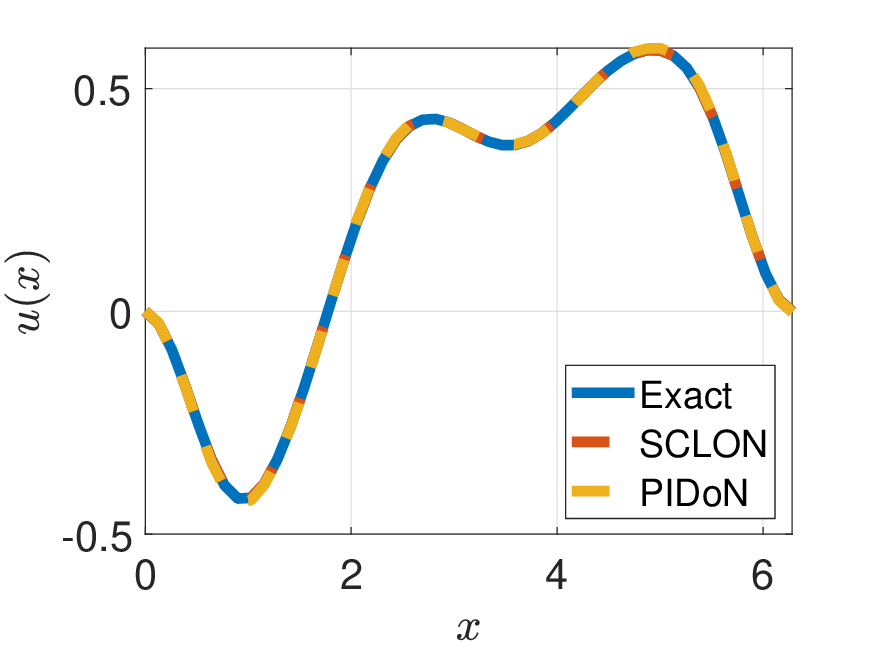}}&\resizebox{0.35\columnwidth}{!}{\includegraphics{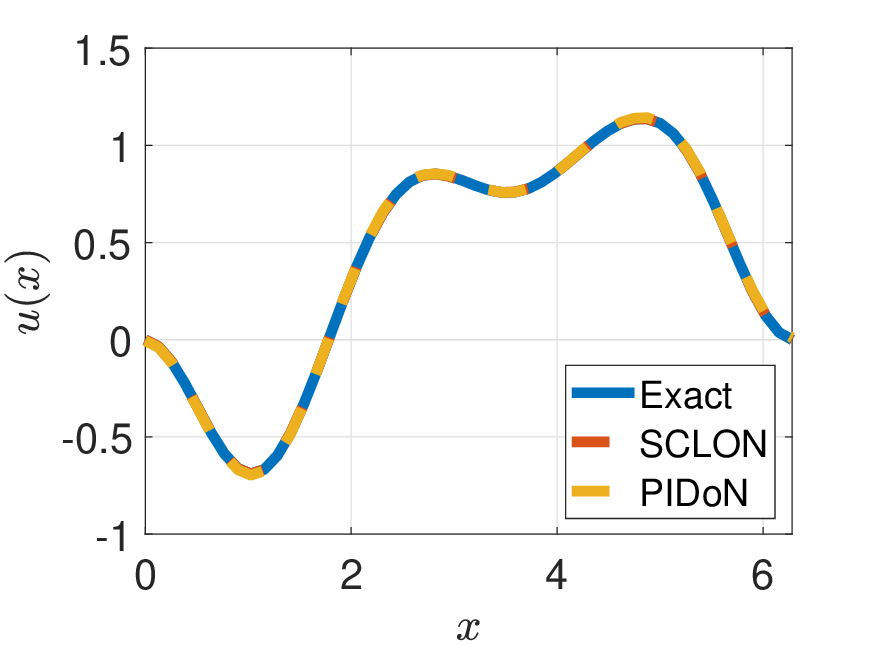}}\\
(g) ${L^\infty} $$error$ for $t=0.25$& (h) for $t=0.5$& (i) for $t=1$\\
\resizebox{0.35\columnwidth}{!}{\includegraphics{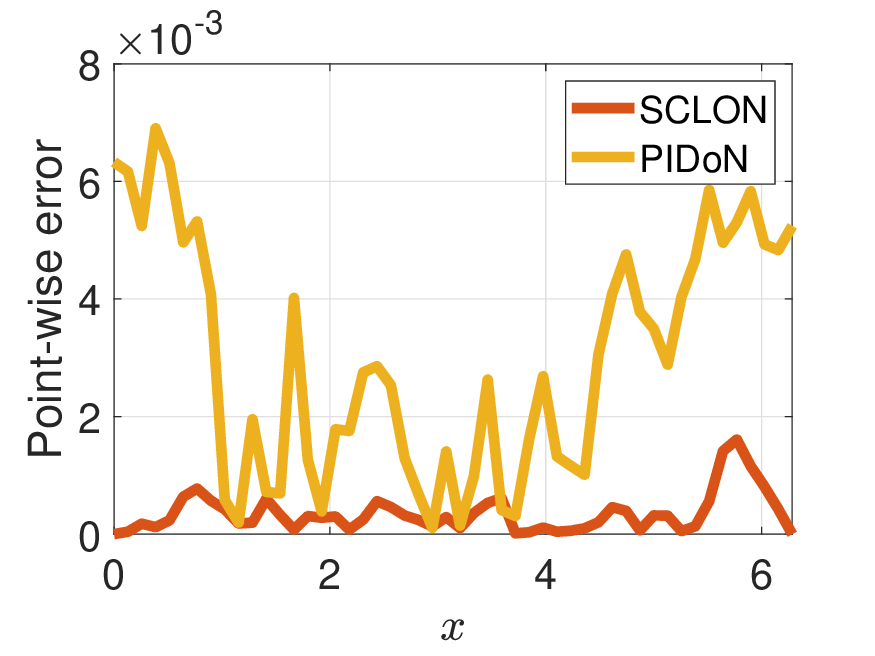}}&\resizebox{0.35\columnwidth}{!}{\includegraphics{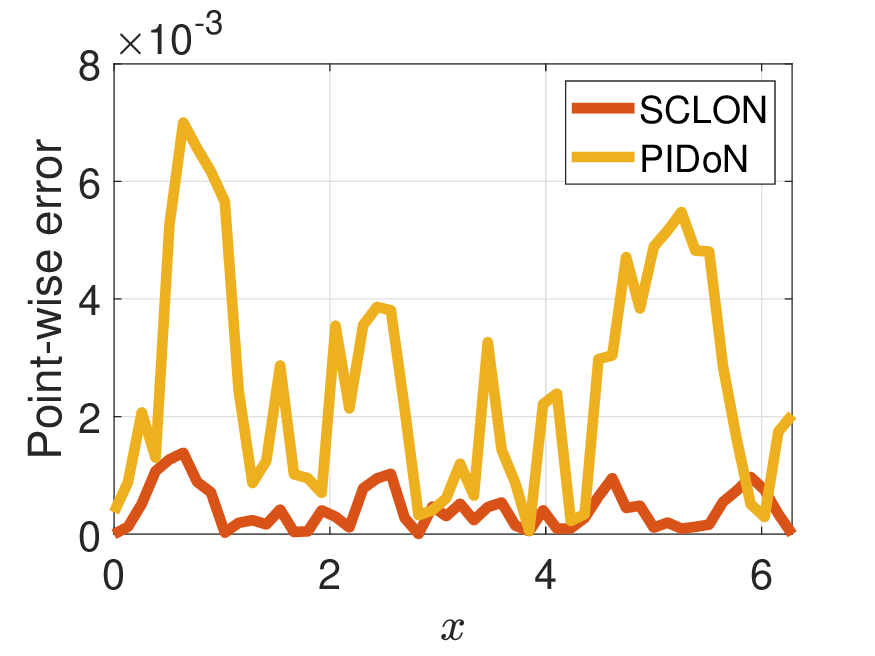}}&\resizebox{0.35\columnwidth}{!}{\includegraphics{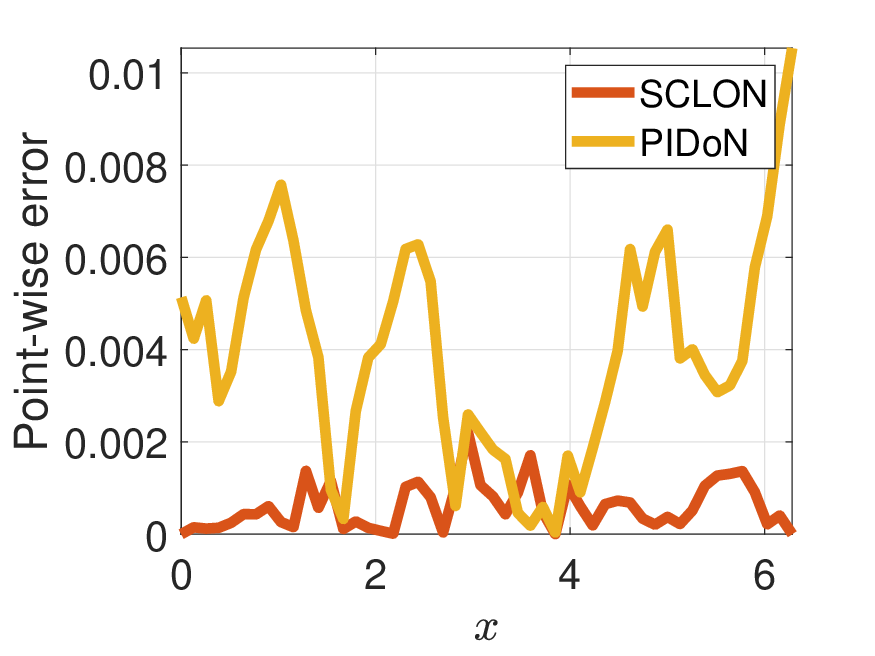}}
\end{tabular}}
\caption{\textbf{Solving the Diffusion-Reaction Equation \eqref{DRE}.} 
\textbf{Top:} (a) Exact solution and (b) SCLON's prediction for a representative example in the test case, with (c) the $L^\infty$ norm error. 
\textbf{Middle:} Time slices of the exact solution, SCLON's prediction, and PIDoN's prediction at (d) $t=0.25$, (e) $t=0.5$, and (f) $t=1$. 
\textbf{Bottom:} The $L^\infty$ norm error between the exact solution and predictions of both SCLON and PIDoN at (g) $t=0.25$, (h) $t=0.5$, and (i) $t=1$. 
For this instance, SCLON's error measures are 4.114e-04 in MAE, 1.322e-03 in Rel.$L^2$, and 1.309e-03 in $L^\infty$. In contrast, PIDoN's errors are 2.994e-03 in MAE, 8.905e-03 in Rel.$L^2$, and 7.837e-03 in $L^\infty$.
}
\label{f:RD}
\end{center}
\end{figure}

We evaluated the accuracy of SCLON's predictions using new and unseen forcing terms that were not in the training process. 
Furthermore, we compared its performance with predictions made by PIDoN \cite{PIDON}. 
As a result, the average relative \(L^2\) error for SCLON is \(0.51\%\), whereas for PIDoN it is \(0.71\%\) using the same PDEs (see table \ref{tab:1}).
This observation suggests that SCLON offers a more accurate approach for predicting solutions to the diffusion-reaction equation compared to PIDoN. Figure \ref{f:RD} demonstrates a predicted solution from SCLON in a test scenario, further emphasizing that its predictions align more closely with the reference solution than those of PIDoN.
Figure \ref{f:RD} (a)-(c) presents both the exact solution and the predictions from SCLON across the entire temporal and spatial domain. Additionally, it depicts the absolute errors between the SCLON predictions (our method) and the reference solution. These errors approximate to around \(0.2\%\).
Solution profiles at \(t=0.25\), \(0.5\), and \(1\) are displayed in Figure \ref{f:RD} (d)-(f), with a detailed view provided in Figure \ref{f:RD} (g)-(i).
Upon direct comparison in \(L^\infty\) errors, it becomes evident that SCLON predictions offer superior accuracy when compared with those of PIDoN.

\section{Viscous Burgers equation}\label{sec_VBE}
In this section, we delineate the architecture of a network designed to replicate the FS. The primary objective of the proposed SCLON method is to learn the solution operator that maps the initial conditions \( u_0(x) \) to the spatio-temporal solution \( u(x, t) \) for the 1D Burgers' equation without the need for a paired input-output dataset.
For this, the one-dimensional viscous Burgers' equation reads
\begin{align}\label{VBE}
\begin{split}
&u_t-\nu u_{xx}+\mu uu_x=0,\quad\text{for}\quad t>0, x\in (0,2\pi)=:\Omega,\\ 
&u=u_0(x),\quad \text{for}\quad t=0,\quad x\in \Omega,
\end{split}
\end{align}
with the periodic boundary condition $u(0,t) = u(2 \pi, t)$. 
For numerical computations, we set $\nu=0.5$ and $\mu=5$.

Since the periodic boundary condition is imposed,
we look for a numerical solution to \eqref{VBE} represented by
\begin{align}\label{VBE_sol}
    u_N^r(x_n)
    =\frac{1}{2\pi}\sum_ {\xi=-N/2+1}^{N/2}e^{\mathrm{i}\xi x_n}\alpha_\xi^r.
\end{align}
Thus, once substituting  \eqref{VBE_sol} and taking DFT to \eqref{VBE} as in \eqref{FTT}, it turns into
\begin{align}\label{VBE_weak2}
     \mathcal{F}_\xi(u_N)_t-\mathcal{F}_\xi\left((\nu u_N)_{xx}+\mu u_N (u_N)_x\right)=0,
\end{align}
for $\xi=-\frac{N}{2}+1,\cdots,\frac{N}{2}$ and $0\leq t\leq T$  .

For a time scheme in \eqref{VBE_weak2}, we employ the fourth-order Runge-Kutta method, which transforms \eqref{VBE_weak2} into the following equation:
\begin{align}\label{VBE_approxi}
\frac{\mathcal{F}_\xi(u_N^{r+1}) - \mathcal{F}_\xi(u_N^{r})}{\Delta t} - \frac{1}{6}(\eta_1 + 2\eta_2 + 2\eta_3 + \eta_4) = 0,
\end{align}
where
\begin{align}
\eta_1&= -\nu \xi^2 \mathcal{F}_\xi(u_N^{r}) -G_\xi(u_N^{r}, \mathcal{F}_\xi(u_N^{r})),\\
\eta_2&= -\nu \xi^2 (\mathcal{F}_\xi(u_N^{r})+\frac{\Delta t}{2}\eta_1)-G_\xi(u_N^{r},\mathcal{F}_\xi(u_N^{r})+\frac{\Delta t}{2}\eta_1), \\
\eta_3&= -\nu \xi^2 (\mathcal{F}_\xi(u_N^{r})+\frac{\Delta t}{2}\eta_2) -G_\xi(u_N^{r},\mathcal{F}_\xi(u_N^{r})+\frac{\Delta t}{2}\eta_2), \\
\eta_4&= -\nu \xi^2(\mathcal{F}_\xi(u_N^{r})+\Delta t\eta_3) -G_\xi(u_N^{r},\mathcal{F}_\xi(u_N^{r})+\eta_3),
\end{align}
for $\xi=-\frac{N}{2}+1,\cdots,\frac{N}{2}$, and where
\begin{align}
G_\xi(u_1,u_2)=\mu\mathcal{F}(u_1\mathcal{F}^{-1}(\mathrm{i}\xi u_2)).
\end{align}
Given that \( u^r_N(x_n) \) in \eqref{VBE_sol} is composed of \( N \) basis functions and \eqref{VBE_approxi} provides \( N \) equations, an \( N \times N \) linear system is established for the unknowns \( \{\mathcal{F}_\xi(u_N^{r+1}) \}_{\xi=-N/2+1}^{N/2} \).
Therefore, after solving the system, we can determine \( u^{r+1} \) using \eqref{VBE_sol}. As a note, we provide reference solutions for \eqref{VBE} using the aforementioned methods to measure errors.

Now, we describe how SCLON replicates the procedures of both the FS and the Runge-Kutta method using a sequential approach.
Let us segment \([0,1]\) into \(Q\) equal intervals, represented as 
$$
[0=t_0,1/Q=t_1],\ldots,[t_{q-1} = (q-1)/Q, t_q = q/Q],\cdots ,[(Q-1)/Q=t_{Q-1},1=t_{Q}].
$$ 
Afterwards, we create $Q$ distinct networks, denoted by \(\mathcal{G}_q\), which correspond to each $(t_{q-1} = (q-1)/Q, t_q = q/Q]$. 
We note that, given a fixed time step size \( \Delta t \), the number of time steps within the interval \( [t_{q-1}, t_q ] \) is given by \( \frac{1}{\Delta t \times Q} \), which we represent as the positive integer \( R \).
In the experiment, we set $\Delta t=0.01$, $Q=10$, $R=10$. Moreover, we draw $P$ initial conditions, $u_{0p}(x)$, as an input from GRF following a distribution $\sim \mathcal{N}(0,25^2 (-\Delta + 5^2 I)^{-2})$, which satisfies periodic boundary conditions. 
Once inserting $u_{0p}$ as an input, the neural networks $\mathcal{G}_q$ for $1 \leq q \leq Q$ yields a set of coefficients $\big[\{\widehat{\alpha}^{r}_{\xi,p}\}_{r=(q-1)R}^{qR-1}\big]_{\xi=-N/2+1}^{N/2}\subset \mathbb{C}$. 
Consequently, the prediction for $(q-1)R\leq r\leq qR-1$ is constructed as follows:
\begin{align}\label{VBE_predic}
\widehat{u}^{r}_p(x_n)=\frac{1}{2\pi}\sum_{\xi=-N/2+1}^{N/2}e^{\mathrm{i}\xi x_n}\widehat{\alpha}_{\xi,p}^{r}.
\end{align}
Following \eqref{VBE_approxi}, the loss for $\mathcal{G}_q$ are written as
\begin{align}\label{loss_VBE}
 loss_q=  \sum_{p=1}^{P}\sum_{r=(q-1)R}^{qR-1}\sum_{\xi=-N/2+1}^{N/2} \left| \widehat{\alpha}_{\xi, p}^{r+1}-\widehat{\alpha}_{\xi, p}^{r}-\frac{\Delta t}{6}(\widehat{\eta}_{1,p}+2\widehat{\eta}_{2, p}+2\widehat{\eta}_{3,p}+\widehat{\eta}_{4,p})   \right|^2
\end{align}
where 
\begin{align}\label{loss_VBE2}
\begin{split}
 \widehat{\eta}_{1,p}&= -\nu \xi^2 \widehat{\alpha}_{\xi, p}^{r} -G_\xi(\mathcal{F}^{-1}(\widehat{\alpha}_{\xi, p}^{r}), \widehat{\alpha}_{\xi, p}^{r}),\\
\widehat{\eta}_{2,p}&= -\nu \xi^2 (\widehat{\alpha}_{\xi, p}^{r}+\frac{\Delta t}{2}\widehat{\eta}_{1,p})-G_\xi(\mathcal{F}^{-1}(\widehat{\alpha}_{\xi, p}^{r}),\widehat{\alpha}_{\xi, p}^{r}+\frac{\Delta t}{2}\widehat{\eta}_{1,p}), \\
\widehat{\eta}_{3,p}&= -\nu \xi^2 (\widehat{\alpha}_{\xi, p}^{r}+\frac{\Delta t}{2}\widehat{\eta}_{2,p}) -G_\xi(\mathcal{F}^{-1}(\widehat{\alpha}_{\xi, p}^{r}),\widehat{\alpha}_{\xi, p}^{r}+\frac{\Delta t}{2}\widehat{\eta}_{2,p}), \\
\widehat{\eta}_{4,p}&= -\nu \xi^2(\widehat{\alpha}_{\xi, p}^{r}+\widehat{\eta}_{3,p}) -G_\xi(\mathcal{F}^{-1}(\widehat{\alpha}_{\xi, p}^{r}),\widehat{\alpha}_{\xi, p}^{r}+\widehat{\eta}_{3,p}).
\end{split}
\end{align}
We halt the training of the model \( \mathcal{G}_q \) if \( loss_q \) plateaus during the process. Afterwards, the sequential method continues until the training of \( \mathcal{G}_Q \) is complete.

\begin{figure}[t!]
\setlength{\tabcolsep}{0.00001pt}
\makebox[\textwidth][c]{\begin{tabular}{ cc }
\resizebox{0.45\columnwidth}{!}{\includegraphics{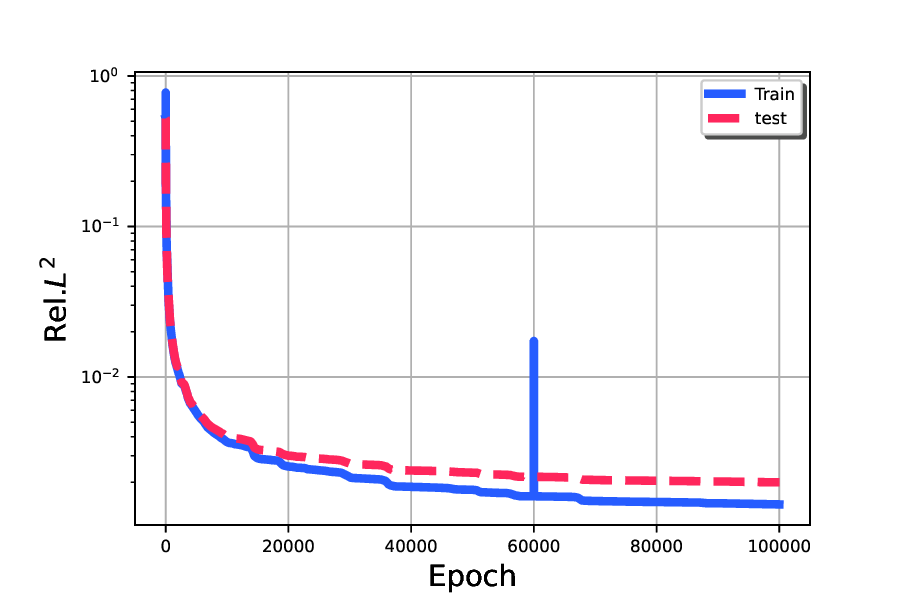}}&\resizebox{0.6\columnwidth}{!}{\includegraphics{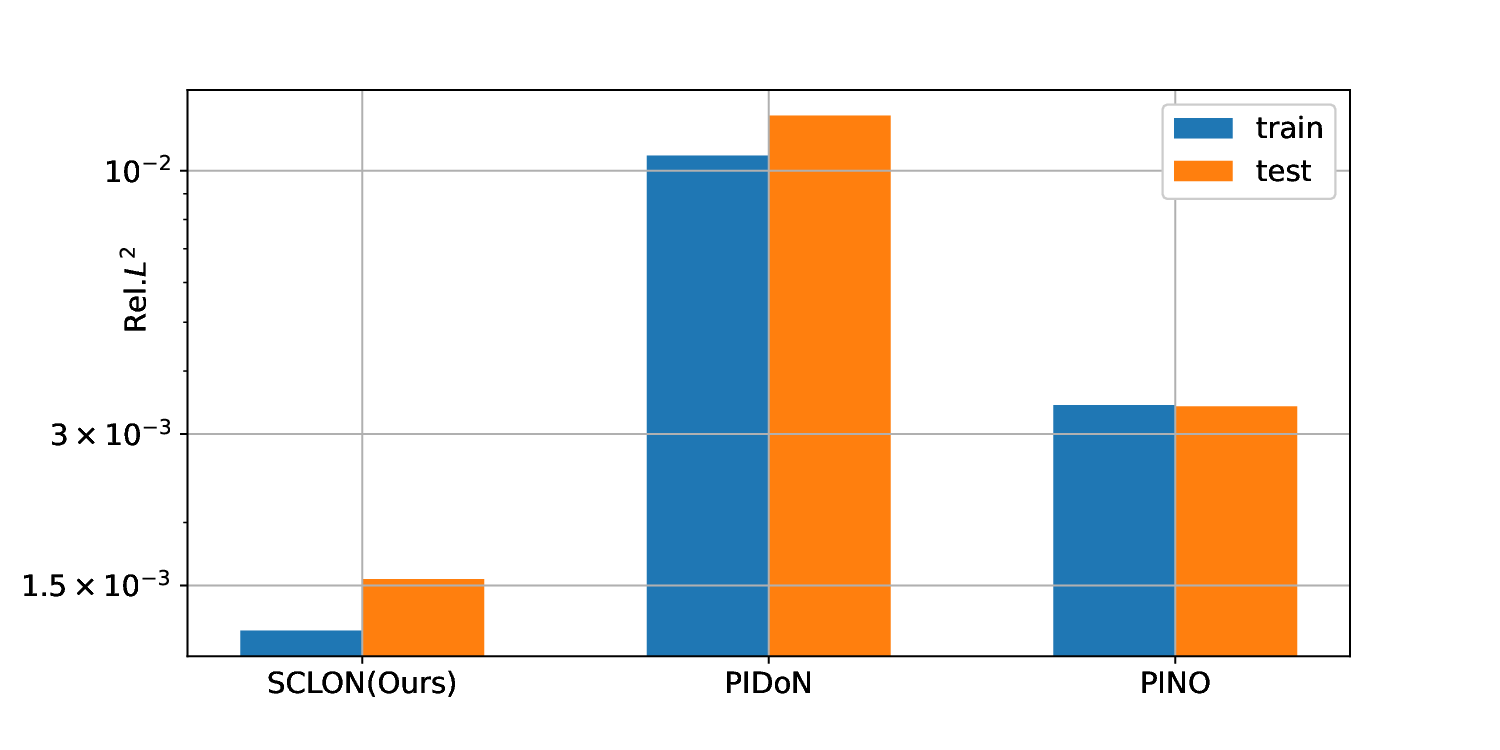}}\\(a) The training trajectory against epochs& (b) The training/test Rel.$L^2$ error
\end{tabular}}
\caption{Burgers' equations: In panel (a), training and test loss curves of SCLON are plotted on a semi-log scale against 100,000 epochs. Panel (b) depicts train error, and test error in the relative $L^2$ of SCLON, PIDoN and PINO (cf. table \ref{tab:3}). The SCLON has smallest train error and smallest test error in the three networks.}
\label{f:burgers_error}
\end{figure}

\begin{figure}[h]
\setlength{\tabcolsep}{0.00001pt}
\makebox[\textwidth][c]{\begin{tabular}{ ccc }
(a) Exact $u(x,t)$& (b) Predicted $\hat{u}(x,t)$& (c) Absolute error\\
\resizebox{0.35\columnwidth}{!}{\includegraphics{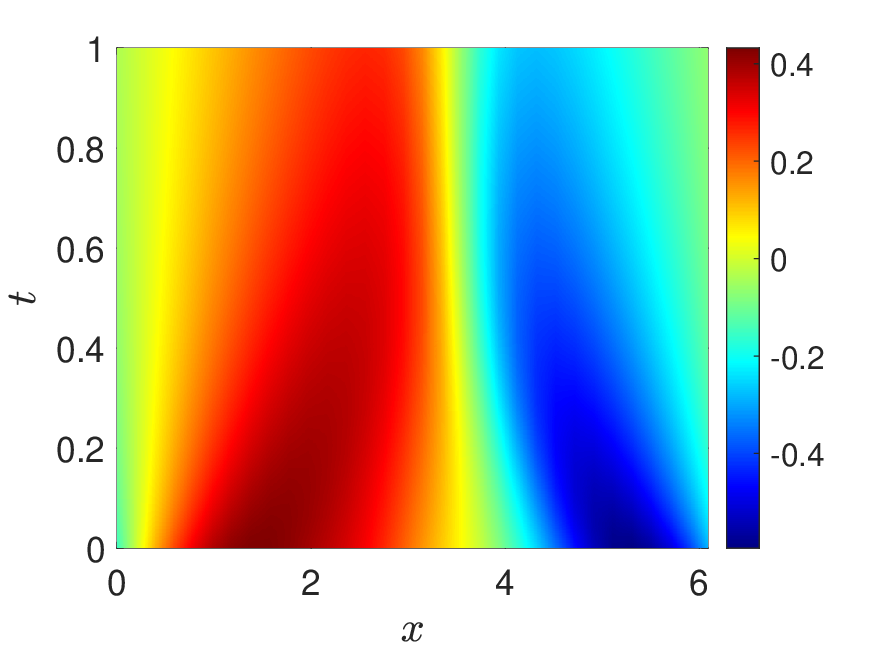}}&\resizebox{0.35\columnwidth}{!}{\includegraphics{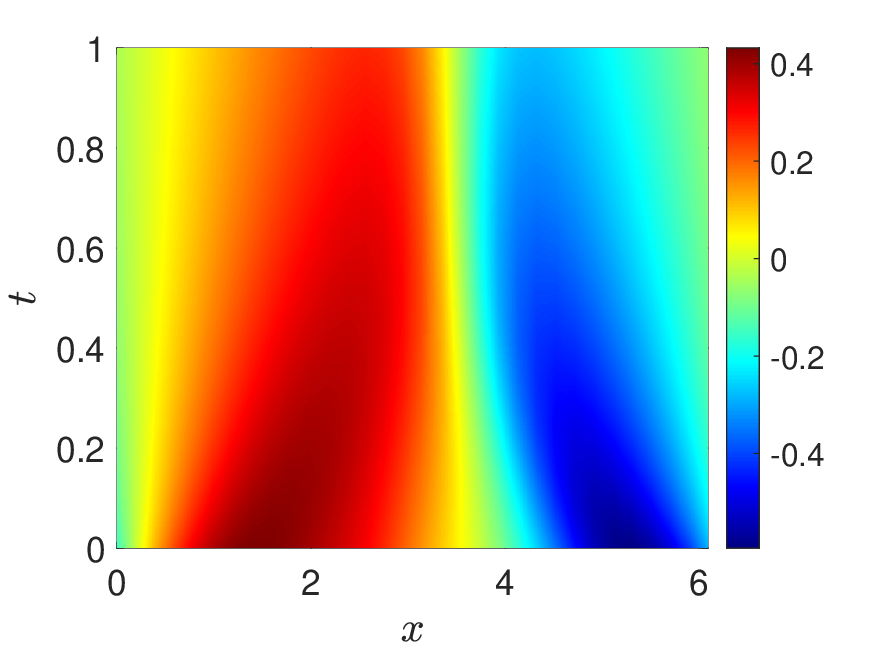}}&\resizebox{0.35\columnwidth}{!}{\includegraphics{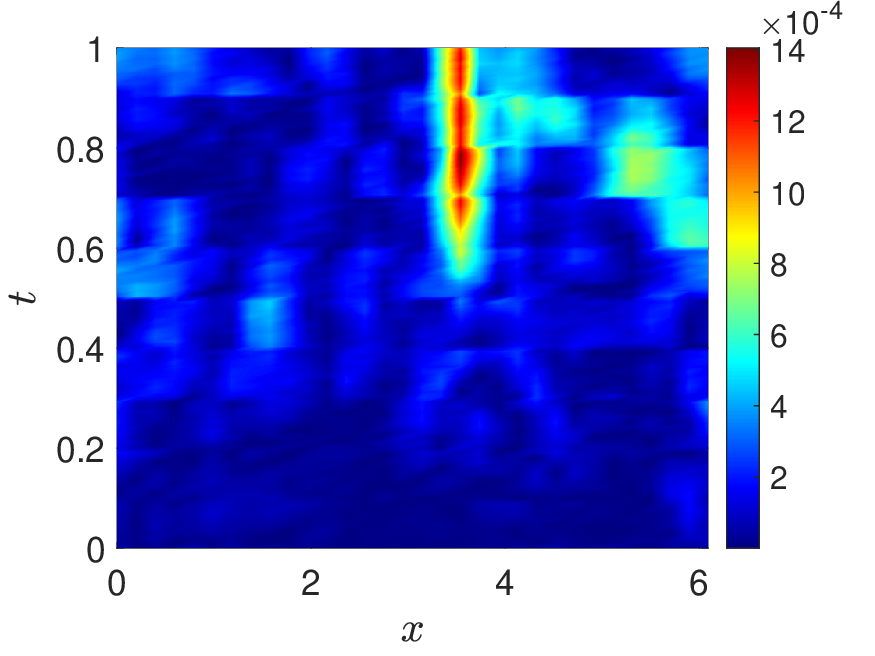}}\\
(d) $$ Exact vs Ours vs PINO vs PIDoN$$ &&\\
for $t=0.25$& (e) for $t=0.5$& (f) for $t=1$\\
\resizebox{0.35\columnwidth}{!}{\includegraphics{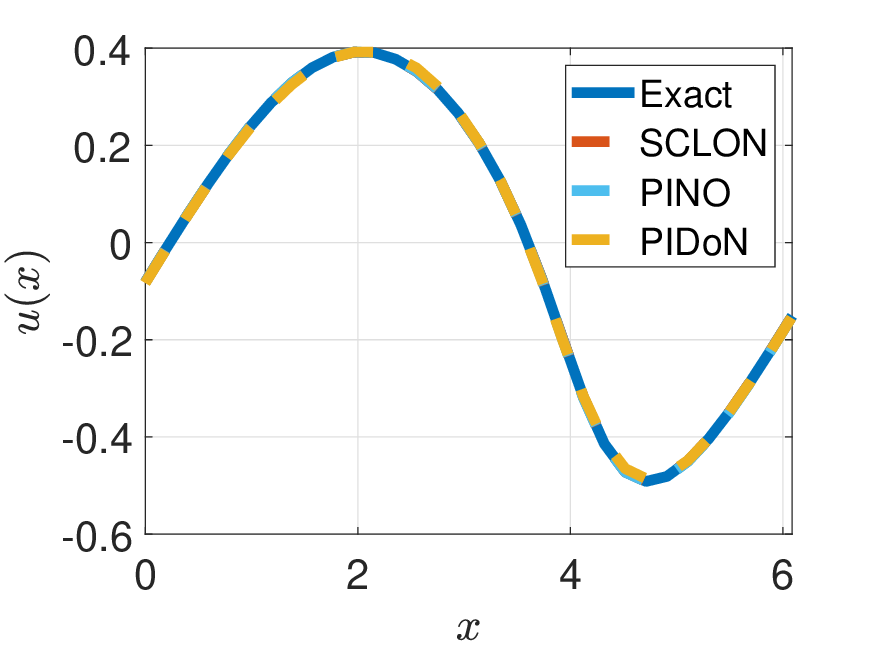}}&\resizebox{0.35\columnwidth}{!}{\includegraphics{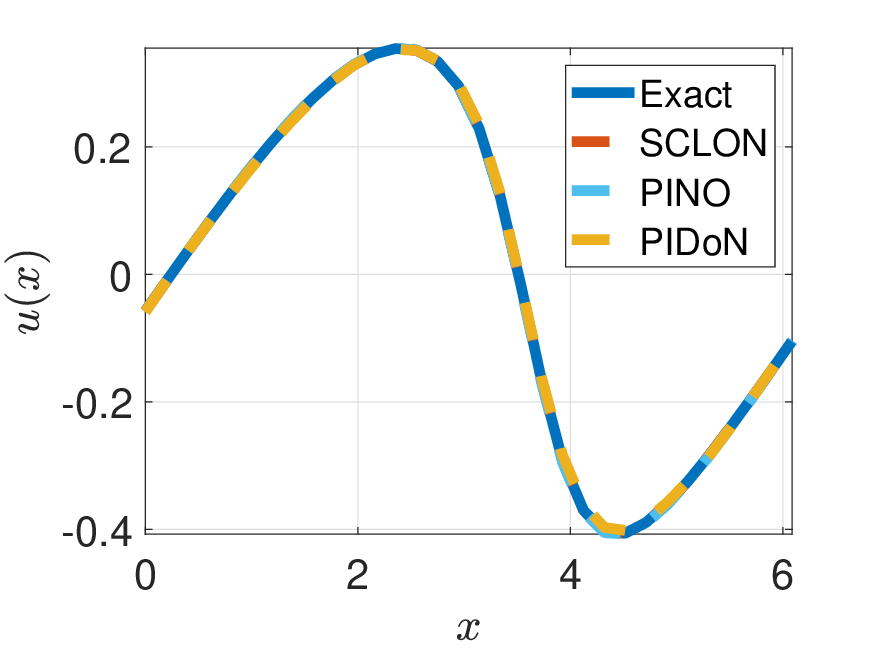}}&\resizebox{0.35\columnwidth}{!}{\includegraphics{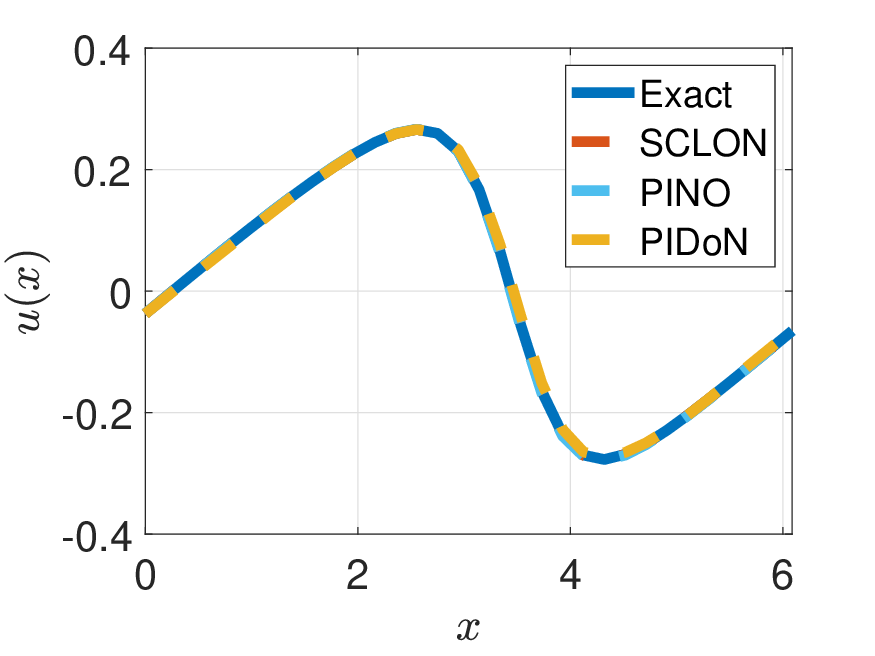}}\\
(g) ${L^\infty} $$error$  for $t=0.25$& (h) for $t=0.5$& (i) for $t=1$\\
\resizebox{0.35\columnwidth}{!}{\includegraphics{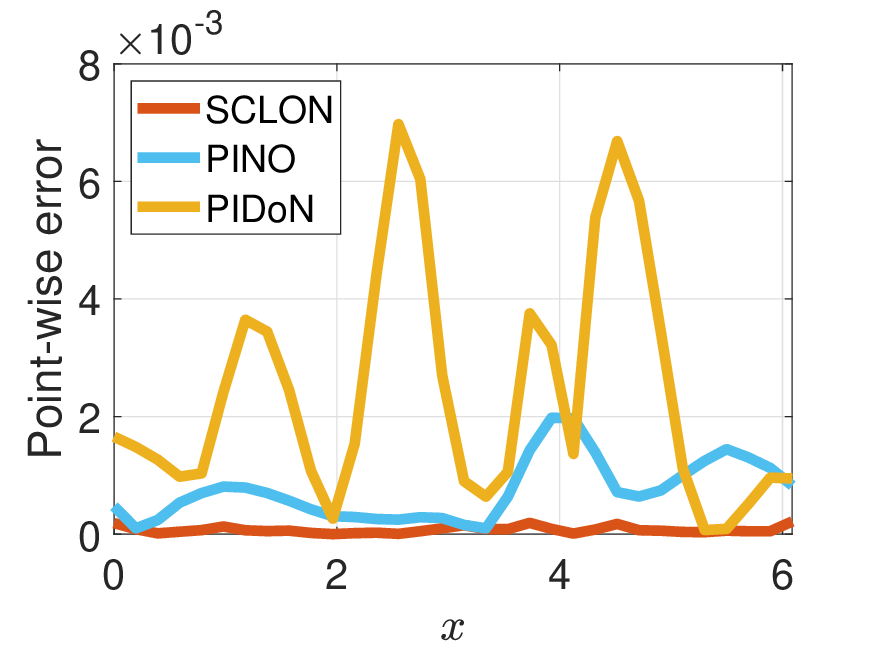}}&\resizebox{0.35\columnwidth}{!}{\includegraphics{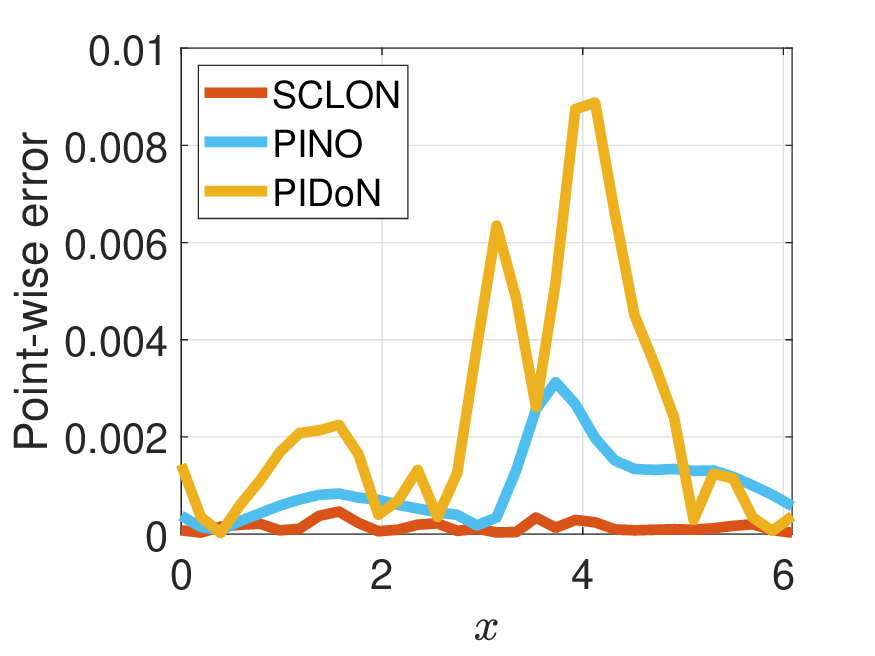}}&\resizebox{0.35\columnwidth}{!}{\includegraphics{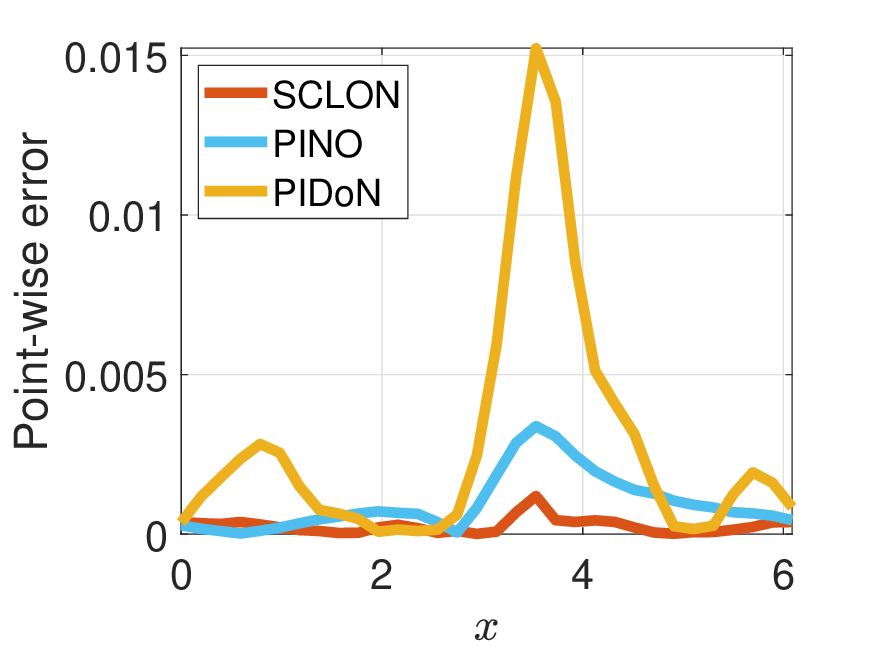}}
\end{tabular}}
\caption{{\bf{Solving a viscous Burgers' Equation \eqref{VBE}. (Top)}} (a) Exact solution versus (b) the prediction of a trained SCLON for a representative example
in the test dataset, and (c) the error in $L^\infty$ norm.  {\bf{(Middle)}} Time slices for Exact solution, the prediction of SCLON (ours), and the one of PINO \cite{PINO} and PIDoN \cite{PIDON} at (d) $t=0.25$, (e) $t=0.5$, and (f) $t=1$. {\bf{(Bottom)}} The error in $L^\infty$ norm between the exact solution and the prediction of SCLON(ours), and between the exact solution and the one of PIDoN at (g) $t=0.25$, (h) $t=0.5$, and (i) $t=1$. For this instance, the error of SCLON is measured as 1.581e-04 in MAE, 9.292e-04 in Rel.:$L^2$, 6.523e-04 in $L^\infty$ whereas the error of PIDoN is 2.486e-03 in MAE, 1.410e-02 in Rel.$L^2$, and 1.021e-02 in $L^\infty$. In addition, the error of PINO \cite{PINO} is 8.342e-04 in MAE, 4.163e-03 in Rel.$L^2$, and 2.643e-03 in $L^\infty$.}
\label{f:burgers}
\end{figure}

We evaluate the accuracy of our predictions using a set of 2,000 new and previously unseen inputs (initial conditions) generated by GRF. These inputs were not utilized during the model training process. Table \ref{tab:3} offers a performance comparison between SCLON, PIDoN, and PINO for the Burgers' equation, using the same equations and initial conditions. Test errors were measured in terms of MAE, Rel.$L^2$, and $L^\infty$, and then averaged over the unseen initial data. 
Notably, the results indicate that SCLON predicts the solution with greater accuracy compared to both PIDoN and PINO. 
Specifically, our SCLON model achieved a commendable average relative $L^2$ error of approximately $0.154\%$, as detailed in Table \ref{tab:3}, surpassing the numerical errors of state-of-the-art models such as PIDoN and PINO.
In Figure \ref{f:burgers_error}, Panel (a) displays the training and test loss curves of SCLON plotted on a semi-log scale against epochs. Panel (b) illustrates the training and test errors in the relative \(L^2\) errors for SCLON, PIDoN, and PINO. Notably, SCLON exhibits the smallest training and test errors among the three models.
Figure \ref{f:burgers} (a)-(c) demonstrates both the exact solution and the predictions from SCLON across the entire temporal and spatial domain for a given initial condition. It also illustrates the absolute errors between the SCLON predictions (our method) and the reference solution, which are approximately \(0.14\%\).
For a performance comparison, solution profiles from various models---SCLON, PINO, and PIDoN---at \(t=0.25\), \(0.5\), and \(1\) are displayed in Figure \ref{f:RD} (d)-(f). A more detailed view is provided in Figure \ref{f:burgers} (g)-(i).
A direct examination of the \(L^\infty\) errors in panels (g)-(i) reveals that the SCLON predictions are notably more accurate than those of PIDoN and PINO.

\begin{table*}[t]
\vskip 0.15in
\begin{center}
\resizebox{\columnwidth}{!}{
\begin{tabular}{||cc|ccc|ccc|ccc||}
\hline
\multirow{3}{*}{Equation}&\multirow{3}{*}{Random input}&\multicolumn{9}{c||}{Error}\\\cline{3-11}
&&\multicolumn{3}{c|}{Ours} & \multicolumn{3}{c|}{PIDoN}& \multicolumn{3}{c||}{PINO} \\
\cline{3-11}
      && MAE&  Rel.$L^2$& $L^\infty$ & MAE& Rel.$L^2$& $L^\infty$ & MAE& Rel.$L^2$& $L^\infty$\\
\hline
 \specialcell{Burgers} & \specialcell{Initial\\ conditions} & 2.358e-04 & 1.544e-03 & 1.010e-03&2.345e-03&1.286e-02&8.989e-03 &7.470e-04 &3.402e-03&1.956e-03 \\ \hline
\end{tabular}
}
\end{center}
\caption{Performance comparison with SCLON, PIDoN, and PINO of the Burgers' equation. The test errors were measuered in MAE, Rel.$L^2$, and $L^\infty$ (see definition of the metrics at Appendix \ref{metrics}), and then averaged over 2,000 new, unseen data. Notably, the results show that SCLON can predict the solution more accurately compared to PIDoN, and PINO.}
\label{tab:3}
\vskip -0.1in
\end{table*}



\section{Advection equation}\label{sec_CE}
In this section, our goal is to train a neural operator capable of mapping variable coefficients to their corresponding PDE solutions. To achieve this, we examine the advection equation with variable coefficients:
\begin{align}\label{CE}
\begin{split}
&u_t+a(x)u_x=0,\quad\text{for}\quad t>0, x\in [0,2\pi),\\ 
&u(x,0)=u_0(x),\quad \text{for}\quad t=0,\quad x\in [0,2\pi),
\end{split}
\end{align}
where $u_0(x)=\frac{1}{2}(1-\cos(x))$ and $a(x)>0$ with periodic boundary conditions.
By using the FS, we look for a numerical solution to \eqref{CE} represented by
\begin{align}\label{CE_sol}
u_N^r(x_n)=\mathcal{F}^{-1}_n(\alpha_\xi)=\frac{1}{2\pi}\sum_ {\xi=-N/2+1}^{N/2}e^{\mathrm{i}\xi x_n}\alpha_\xi^r.
\end{align}
Once substituting  \eqref{CE_sol} and taking DFT \eqref{FTT} to \eqref{CE}, it becomes
\begin{align}\label{CE_weak2}
    \mathcal{F}_\xi(u_N)_t-\mathcal{F}_\xi\left(a (u_N)_x\right)=0,
\end{align}
for $\xi=-\frac{N}{2}+1,\cdots,\frac{N}{2}$. 
For the temporal method, we utilize the fourth-order Runge-Kutta method. Consequently, \eqref{CE_weak2} is approximated as
\begin{align}\label{CE_RK4}
  \frac{\mathcal{F}_{\xi}(u^{r+1})-\mathcal{F}_{\xi}(u^{r})}{\Delta t}-\frac{1}{6}(\eta_1+2\eta_2+2\eta_3+\eta_4)=0, 
\end{align}
where
\begin{align}\label{CE_eta2}
\begin{split}
\eta_1&= -\mathcal{F}_{\xi}(a\mathcal{F}_{n}^{-1}(\mathrm{i}\xi\mathcal{F}_{\xi}(u^{r}))),\\
\eta_2&= -\mathcal{F}_{\xi}(a\mathcal{F}_{n}^{-1}(\mathrm{i}k\xi(\mathcal{F}_{\xi}(u^{r})+\frac{\Delta t}{2}\eta_1))),\\
\eta_3&= -\mathcal{F}_{\xi}(a\mathcal{F}_{n}^{-1}(\mathrm{i}\xi(\mathcal{F}_{\xi}(u^{r})+\frac{\Delta t}{2}\eta_2))),\\
\eta_4&= -\mathcal{F}_{\xi}(a\mathcal{F}_{n}^{-1}(\mathrm{i}\xi(\mathcal{F}_{\xi}(u^{r})+\Delta t\eta_3))).
\end{split}
\end{align}
Given that \( u^r_N(x) \) in \eqref{CE_sol} consists of \( N \) basis functions and \eqref{CE_RK4} provides \( N \) equations, an \( N \times N \) linear system is established for the unknowns \( \{\alpha^r_\xi\}_{\xi=-N/2+1}^{N/2} \). After solving this system, we can determine \( u^{r+1} \). This process is employed to obtain the reference solution.

Now, we describe how SCLON replicates the procedure of the FS and the Runge-Kutta method in neural networks. To provide variable coefficient functions as input data, we first draw functions \( \tilde{a}(x) \) from GRF \( \sim \mathcal{N}(0,30^2 (-\Delta + 8^2 I)^{-2}) \). Afterward, to ensure they are strictly positive, we transform them into 
\begin{align}
    a(x)=\tilde{a}(x)-\min_x(\tilde{a}(x))+1.
\end{align}
Using the procedure described above, we generate \( P \) input samples. When given one of these \( P \) input samples, denoted by \( a_p(x) \), the operator network \( \mathcal{G}_q \) produces the coefficient \( \{\widehat{\alpha}^{r}_{\xi,p}\}_{r=(q-1)R}^{qR-1} \subset \mathbb{C} \) corresponding to \( a_p(x) \).
With this, the prediction is constructed as
\begin{align}
   \widehat{u}^{r}_{N,p}=\frac{1}{2\pi}\sum_ {\xi=-N/2+1}^{N/2}e^{\mathrm{i}\xi x_n}\widehat{\alpha}_{\xi, p}^{r},
\end{align}
for $(q-1)R\leq r\leq qR-1$.

To continue, in \eqref{CE_RK4}, we replace \( \mathcal{F}_{\xi}(u^{r}) \) with  
$\widehat{\alpha}^{r}_{\xi,p}$ as  
\begin{align}\label{CE_hateta}
\begin{split}
 \widehat{\eta}_{1,p}&= -\mathcal{F}_{\xi}(a_p\mathcal{F}_{n}^{-1}(\mathrm{i}\xi\widehat{\alpha}_{\xi, p}^{r})),\\
\widehat{\eta}_{2,p}&= -\mathcal{F}_{\xi}(a_p\mathcal{F}_{n}^{-1}(\mathrm{i}k\xi(\widehat{\alpha}_{\xi,p}^{r}+\frac{\Delta t}{2}\widehat{\eta}_{1p}))),\\
\widehat{\eta}_{3,p}&= -\mathcal{F}_{\xi}(a_p\mathcal{F}_{n}^{-1}(\mathrm{i}\xi(\widehat{\alpha}_{\xi, p}^{r}+\frac{\Delta t}{2}\widehat{\eta}_{2p}))),\\
\widehat{\eta}_{4,p}&= -\mathcal{F}_{\xi}(a_p\mathcal{F}_{n}^{-1}(\mathrm{i}\xi(\widehat{\alpha}_{\xi, p}^{r}+\Delta t\widehat{\eta}_{3p}))).
\end{split}
\end{align}
Using \eqref{CE_RK4} and \eqref{CE_hateta}, we define a loss for \( \mathcal{G}_q \) as
\begin{align}\label{loss_CE}
 loss_q=  \sum_{p=1}^{P}\sum_{r=(q-1)R}^{qR-1}\sum_{\xi=-N/2+1}^{N/2} \left| \widehat{\alpha}_{\xi, p}^{r+1}-\widehat{\alpha}_{\xi, p}^{r}-\frac{\Delta t}{6}(\widehat{\eta}_{1,p}+2\widehat{\eta}_{2, p}+2\widehat{\eta}_{3,p}+\widehat{\eta}_{4,p})   \right|^2.
\end{align}
Therefore, as \( loss_q \) approaches zero, the values of \( \widehat{u}_{N, p}^{r+1} \) are expected to be closer to \( u_{N, p}^{r+1} \). We continue training the model \( \mathcal{G}_q \) until the loss function, as defined in \eqref{loss_CE}, plateaus during the training process. Subsequently, the sequential method is applied repeatedly until the training of \( \mathcal{G}_Q \) is complete.

\begin{figure}[h!]
\begin{center}
\setlength{\tabcolsep}{0.00001pt}
\makebox[\textwidth][c]{\begin{tabular}{ ccc }
(a) Exact $u(x,t)$& (b) Predicted $\hat{u}(x,t)$& (c) Absolute error\\
\resizebox{0.35\columnwidth}{!}{\includegraphics{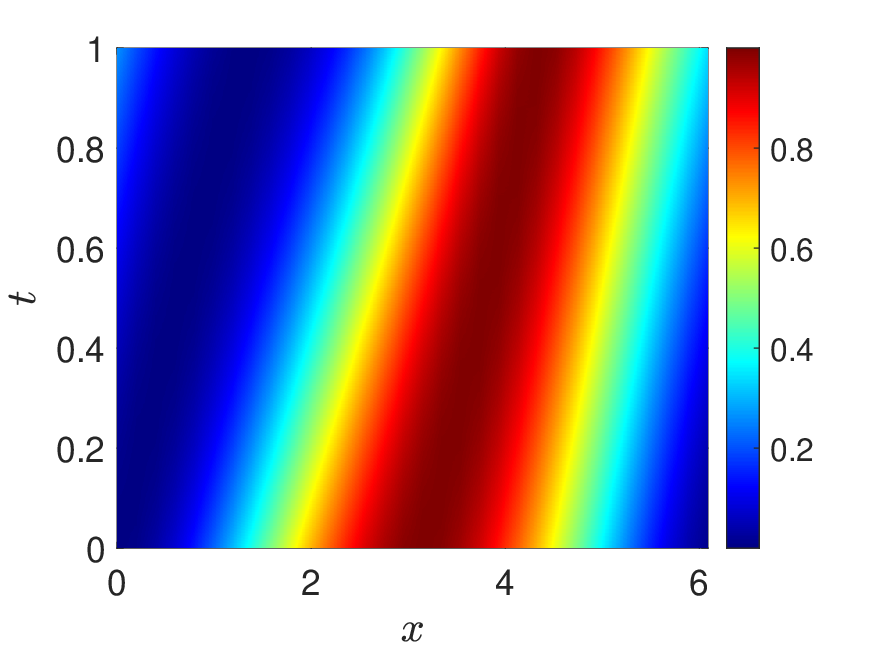}}&\resizebox{0.35\columnwidth}{!}{\includegraphics{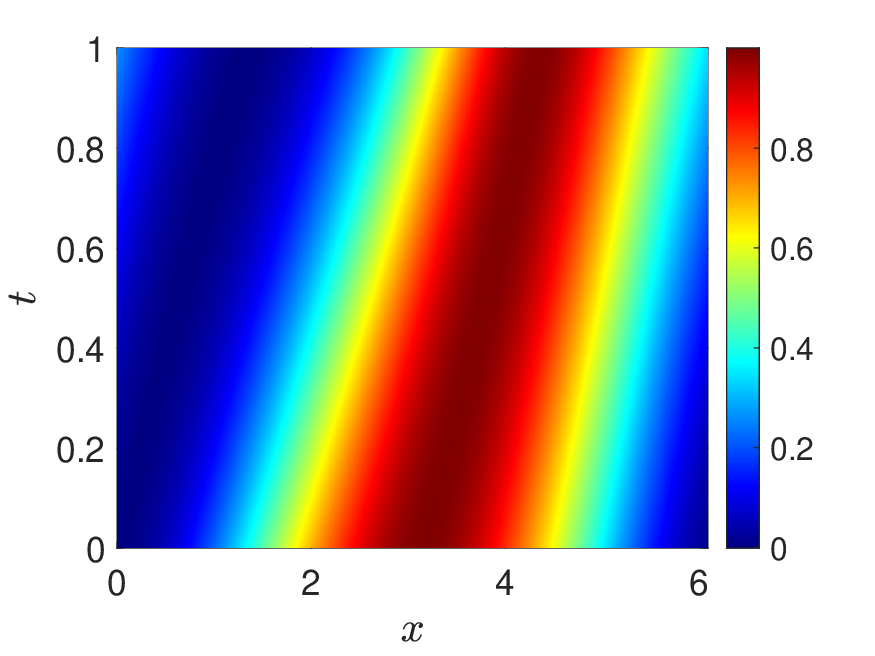}}&\resizebox{0.35\columnwidth}{!}{\includegraphics{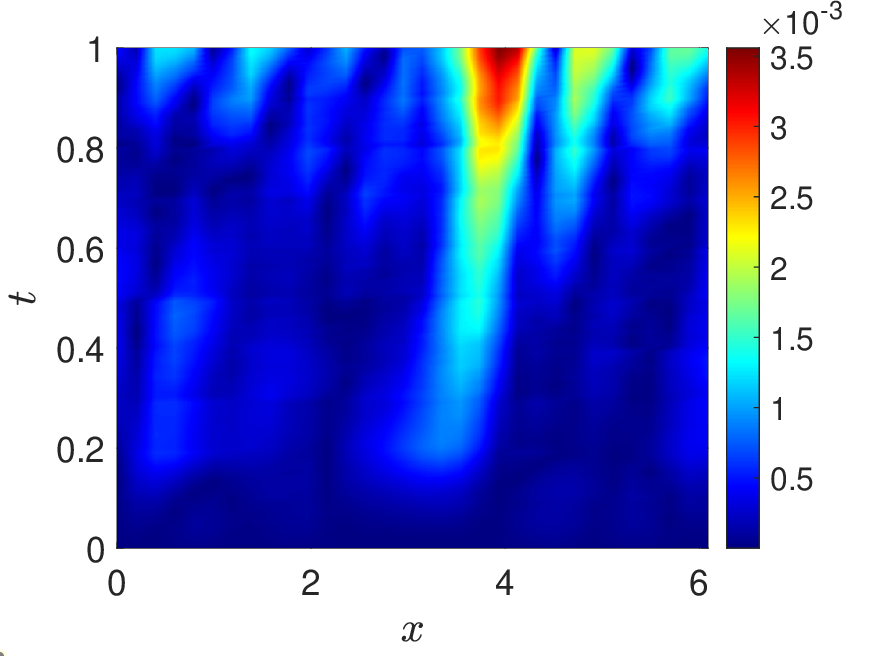}}\\
(d) $$ Exact vs Ours $$&&\\
for $t=0.25$& (e) for $t=0.5$& (f) for $t=1$\\
\resizebox{0.35\columnwidth}{!}{\includegraphics{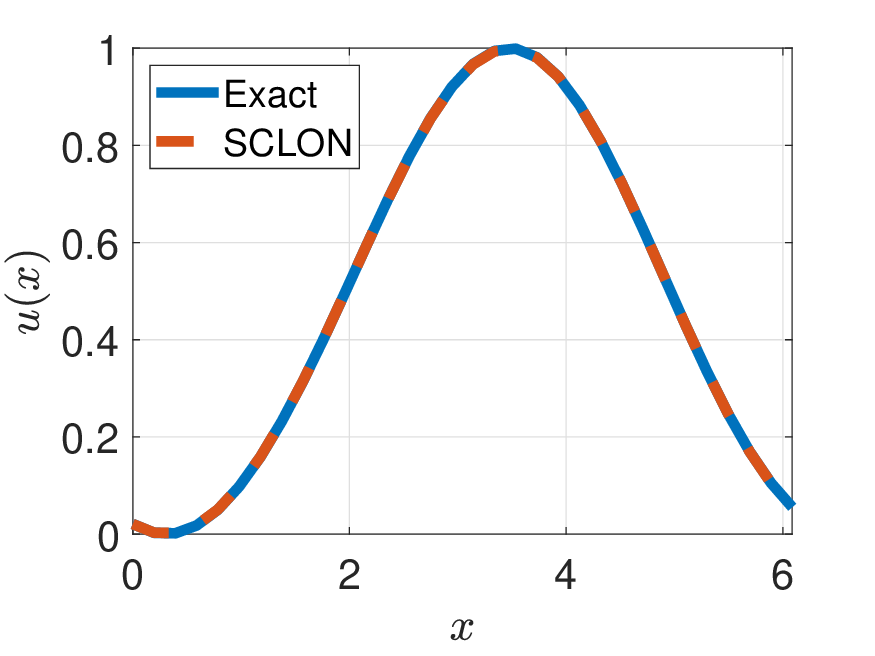}}&\resizebox{0.35\columnwidth}{!}{\includegraphics{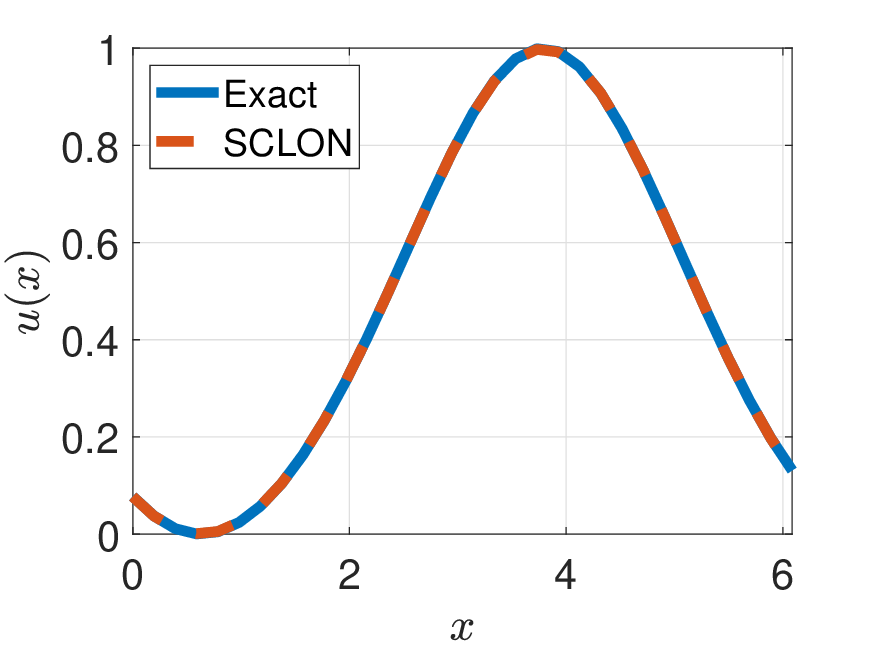}}&\resizebox{0.35\columnwidth}{!}{\includegraphics{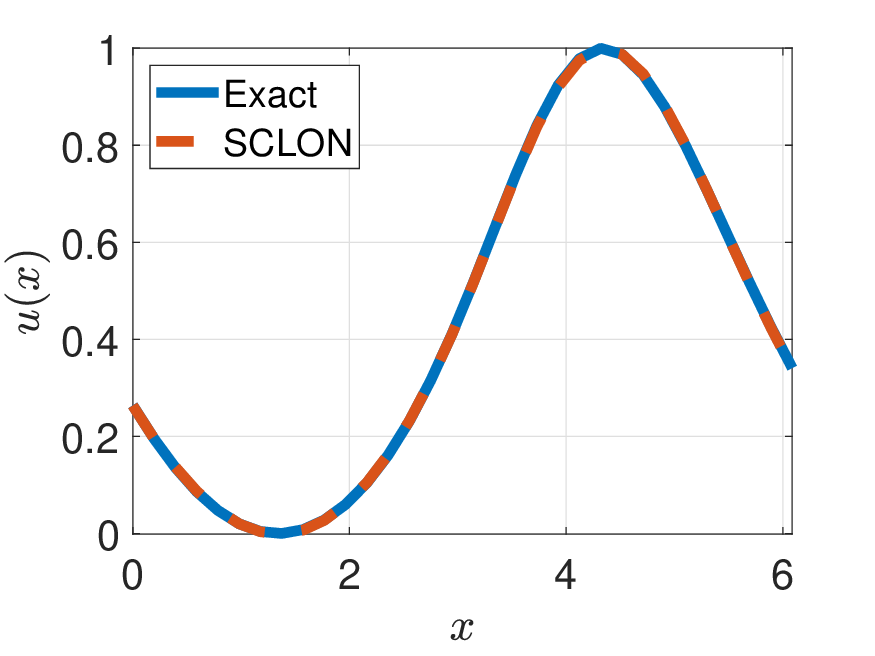}}\\
(g) ${L^\infty} $$error$  for $t=0.25$& (h) for $t=0.5$& (i) for $t=1$\\
\resizebox{0.35\columnwidth}{!}{\includegraphics{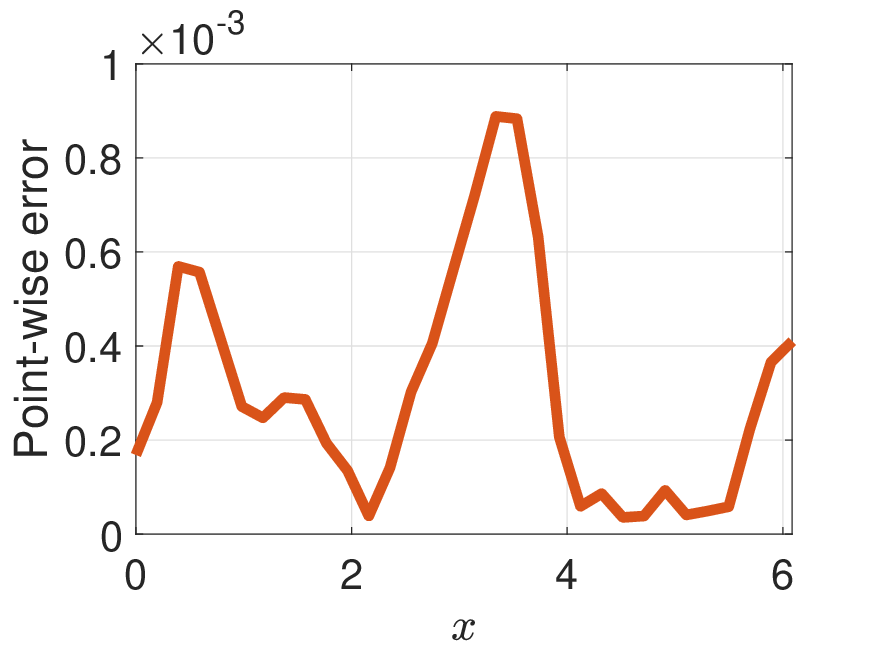}}&\resizebox{0.35\columnwidth}{!}{\includegraphics{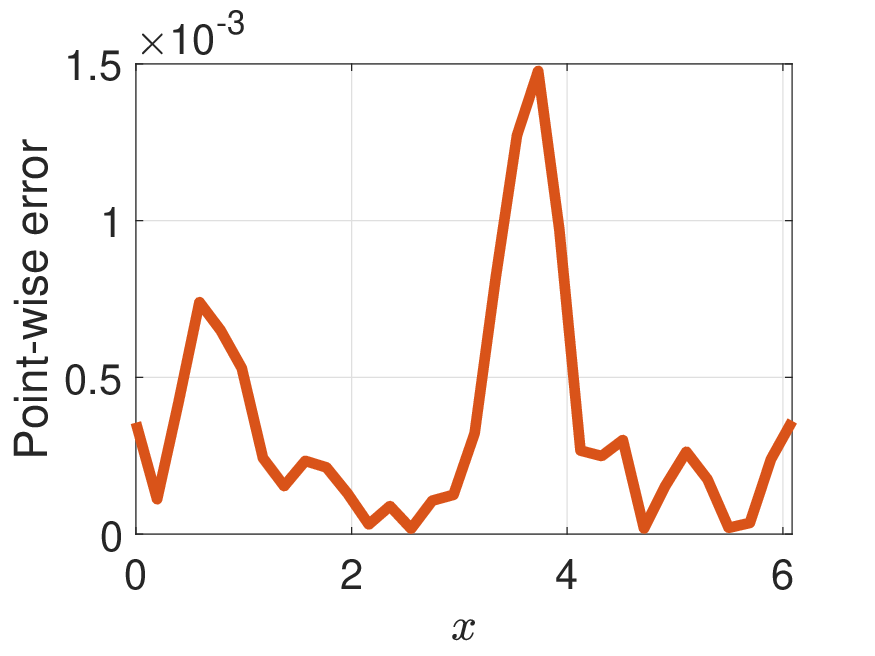}}&\resizebox{0.35\columnwidth}{!}{\includegraphics{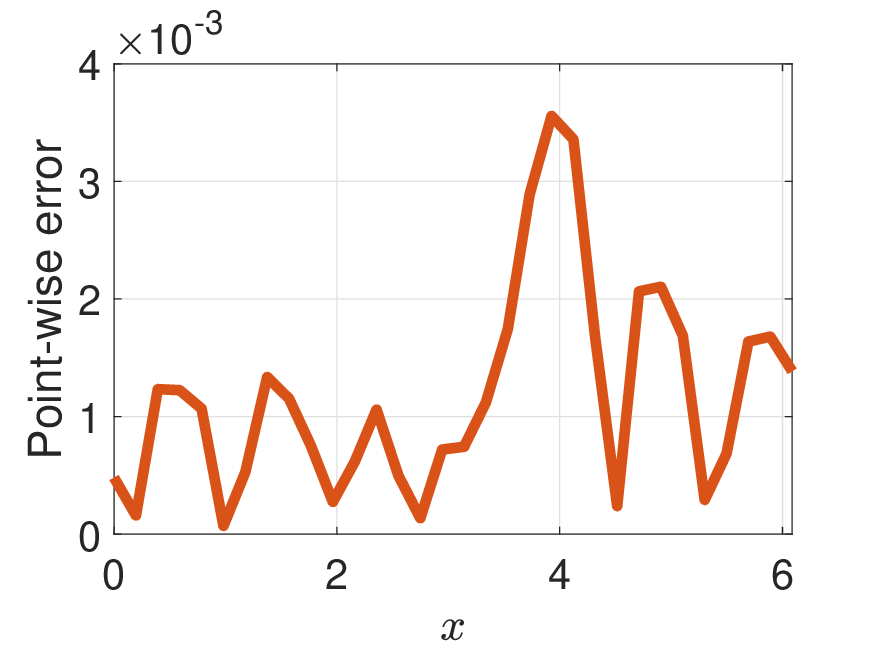}}
\end{tabular}}
\caption{{\bf{Solving an advection  Equation \eqref{CE}. (Top)}} (a) Exact solution versus (b) the prediction of a trained SCLON for a representative example
in the test dataset, and (c) the error in $L^\infty$ norm.  {\bf{(Middle)}} Time slices for Exact solution, the prediction of SCLON(ours) at (d) $t=0.25$, (e) $t=0.5$, and (f) $t=1$. {\bf{(Bottom)}} The error in $L^\infty$ norm between the exact solution and the prediction of SCLON(ours) at (g) $t=0.25$, (h) $t=0.5$, and (i) $t=1$. For this instance, the error of SCLON is measured as 4.276e-04 in MAE, 1.149e-03 in Rel.:$L^2$, 1.532e-03 in $L^\infty$.}
\label{f:adv_eq}
\end{center}
\end{figure}

We test the accuracy of our model predictions using a set of 2,000 new and previously unseen inputs (variable coefficients) generated by the same GRF. These inputs were not part of the model training process. Figure \ref{f:adv_eq} illustrates the high accuracy of our trained SCLON model in predicting the reference PDE solution. 
Figure \ref{f:adv_eq} (a)-(c) showcases both the exact solution and the SCLON predictions across the entire temporal and spatial domain, along with the absolute errors between our method and the reference solution, approximating around \(0.4\%\). Solution profiles at \(t=0.25\), \(0.5\), and \(1\) are depicted in Figure \ref{f:adv_eq} (d)-(f), and a more detailed view is provided in Figure \ref{f:adv_eq} (g)-(i).
Our numerical evidence confirms that the SCLON model serves as an accurate surrogate for advection equations, as highlighted by the impressive low average relative \(L^2\) prediction error of just \(0.217\%\) across all test examples.

\section{Convection-Diffusion equation (CDE) with a boundary layer}\label{sec_DCE}
We utilize our methodology to effectively tackle singular perturbations and boundary layer issues, highlighting one of our key strengths.
To this end, we consider the Diffusion-convection equation with $\nu\ll1$   
\begin{align}\label{DCE}
\begin{split}
&u_t-\nu u_{xx}-u_x=0,\quad\text{for}\quad t>0, x\in \Omega,\\ 
&u=u_0(x),\quad \text{for}\quad t=0,\quad x\in \Omega,\\
&u(-1)=u(1)=0,\quad \text{for}\quad t \geq 0.
\end{split}
\end{align}
The boundary layer problem is a well-known challenge in scientific computing, especially in the context of fluid mechanics. It arises due to the presence of thin regions of high gradients in the solution near boundaries, which can significantly affect the overall behavior of the system. 
These problems are of great significance in scientific research as they occur in various fields, such as fluid mechanics, chemical kinetics, and semiconductor physics.
However, the boundary layer problem is challenging for scientific computing for several reasons. Firstly, the high gradients in the solution require a fine spatial resolution near the boundary, which can lead to significant computational costs. Secondly, the behavior of the boundary layer can be highly sensitive to the boundary conditions and the choice of numerical method used to solve the problem, making it difficult to obtain accurate and reliable results. Finally, the presence of the boundary layer can result in numerical instabilities, which can make it challenging to develop robust and stable numerical methods for solving the problem.
In general, when the target function contains sharp transition, i.e., when the diffusion coefficient, $\nu>0$, is small in our model problem, neural network algorithms often fail to converge to desirable solutions due to the so-called spectral bias phenomenon. 
The general learning process of neural networks relies on a smooth prior, and spectral bias leads to a failure to accurately capture sharp transitions or singular behaviors of the target solution function.
More precisely, while neural networks tend to learn low-frequency or smooth components quickly, they require much time to fit sharp transitions. Hence, without care, neural networks cannot fit the sharp transitions caused by the boundary layer.

We introduce a novel semi-analytic machine learning approach, guided by theory, for effectively capturing the behavior of thin boundary layers. Our framework draws inspiration from the boundary layer theory and asymptotic analysis, which are widely recognized in the applied mathematics community for solving singular perturbed differential equations \cite{MR0402217, mayes2003boundary, MR2148856}. 
For instance, asymptotic basis functions have been added to the basis functions in the Galerkin framework to accurately capture and represent the singular behavior inherent in such solutions \cite{H2020}. 
Inspired by perturbation theory and its use of asymptotic basis functions, we propose the basis-enriched SCLON (BE-SCLON) method to overcome the current limitations of deep learning in resolving thin boundary layers. 
By incorporating boundary layer theory, our BE-SCLON approach enables the prediction of accurate numerical solutions for singular perturbation problems.
More precisely, the essence of the enriched method involves introducing an additional element, known as corrector functions, that represents the boundary layer profile to the finite or spectral element space. Derived from boundary layer analysis, these corrector functions capture the solution within the boundary layer, which includes a profile of the sharp transition.
The concept of enriched spaces in the spectral element method was initially introduced in \cite{hong2018enriched} and subsequently explored in the context of numerical analysis \cite{H2020}. 
The key aspect of the enriched scheme in the spectral element method is to identify the appropriate corrector functions, denoted as $\varphi$, which are defined globally across the physical domain. 
Typically, the corrector function $\varphi$ is obtained through singular perturbation analysis, focusing on the leading order, by zooming in (using stretched variables) near the vicinity of the boundary layer; for more details see e.g., \cite{hong2018enriched, SP_book}
In this section, we establish the SCLON to solve singularly perturbed diffusion-convection equations \eqref{DCE} with $\nu=10^{-6}$. 
It sets off with defining a weak formulation in $H^1_0(\Omega)$ as   
  \begin{align}\label{DCE_weak}
    \int_{\Omega}u_t\phi+ \nu u_x\phi_x-\mu u_x\phi dx=0,
\end{align}
for $0\leq t\leq T$.
From the boundary layer analysis \cite{hong2018enriched}, the boundary layer corrector function is derived as
\begin{align}\label{corrector}
\varphi_{N}=\exp(-(1+x)/\nu)-\left(1-\frac{1-\exp(-2/\nu)}{2}(x+1)\right).
\end{align}
Hence, the LS approximation finds a numerical solution represented by
\begin{align}\label{DCE_sol}
u_N^r=\sum_{n=0}^{N-1}\alpha_n^r \phi_n+\alpha^r_{N} \varphi_{N}.
\end{align}
For the temporal scheme, we adopt the implicit Euler method, which transforms \eqref{DCE_weak} into  
\begin{align}\label{DCE_euler}
    \int_\Omega\frac{u^{r+1}-u^{r}}{\Delta t}\phi_n+\nu u^{r+1}_{x}\phi_{nx}-\mu u^{r+1}_x\phi_n dx=0.
\end{align}
As a result, by successively computing \eqref{DCE_euler}, we obtain a weak solution \eqref{DCE_sol} to \eqref{DCE_euler} for \(1\leq r\leq T/\Delta t\).

In a manner similar to Section \ref{sec_DRE}, our network adheres to the structure defined by the numerical scheme \eqref{DCE_euler}. We first create $P$ initial conditions as input data in the form:
\begin{align}\label{DCE_input}
u_{0p}(x)=(1-x)^4(1+x)\left(\sum_{j=0}^{3}a_{j,p}\phi_{j}\right)
\end{align}
where $a_{j,p}$ are random numbers drawn from a uniform distribution on $[0,1)$ for $p=1,\cdots, P$. In particular, we multiply by $(1-x)^4(1+x)$ to ensure compatibility between the initial condition and the boundary condition at $x=\pm 1$. When our network \( \mathcal{G}_q \) and \eqref{DCE_sol} make predictions for \( \{\widehat{u}_{p}^r\}_{r=(q-1)R+1}^{qR} \) against the input data \( u_{0p} \), the loss can be written as 
\begin{align}\label{loss_DCE}
    loss_q=\sum_{p=1}^{P}\sum_{r=(q-1)R}^{qR-1}\sum_{n=0}^{N}\left|\int_{\Omega}\frac{\widehat{u}^{r+1}_p-\widehat{u}^{r}_p}{\Delta t}\phi_n dx+\nu \int_{\Omega}(\widehat{u}_p^{r+1})_x(\phi_n)_x dx+\int_{\Omega} (\widehat{u}_p^{r})_x\phi_n dx\right|^2.
\end{align}
Lastly, we continue training the network \( \mathcal{G}_q \) until the loss function, as described in \eqref{loss_DCE}, plateaus during the training process.

\begin{figure}[h!]
\begin{center}
\setlength{\tabcolsep}{0.00001pt}
\makebox[\textwidth][c]{\begin{tabular}{ ccc }
(a) Exact $u(x,t)$& (b) Predicted $\hat{u}(x,t)$& (c) Absolute error\\
\resizebox{0.35\columnwidth}{!}{\includegraphics{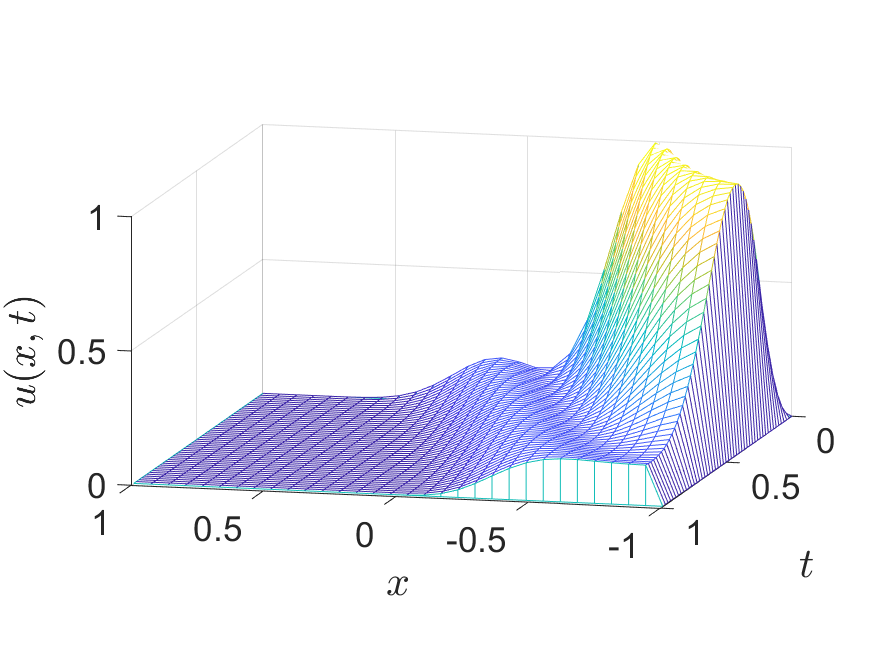}}&\resizebox{0.35\columnwidth}{!}{\includegraphics{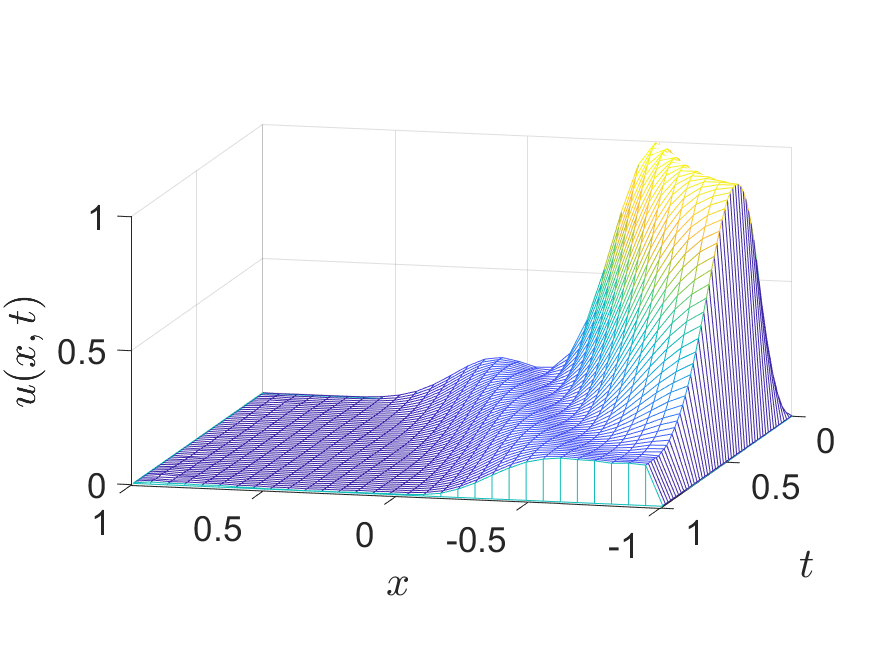}}&\resizebox{0.35\columnwidth}{!}{\includegraphics{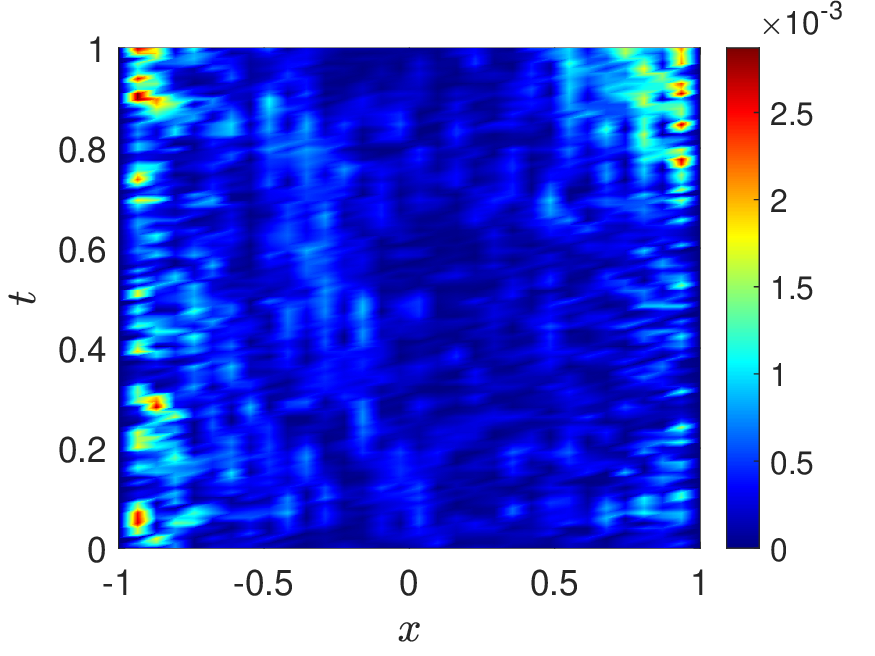}}\\
(d) $$ Exact vs Ours vs PIDoN$$&&\\
for $t=0.25$& (e) for $t=0.5$& (f) for $t=1$\\
\resizebox{0.35\columnwidth}{!}{\includegraphics{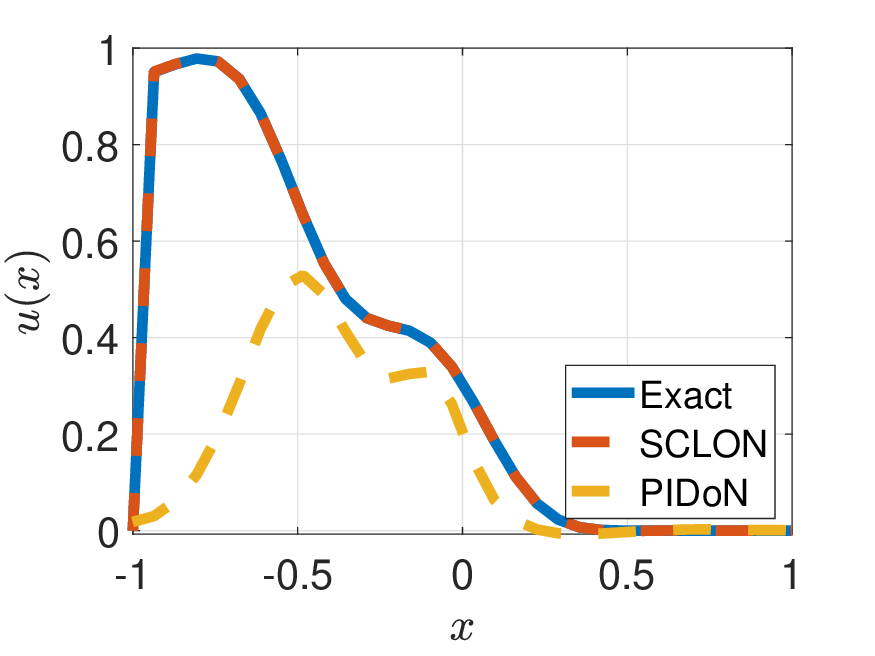}}&\resizebox{0.35\columnwidth}{!}{\includegraphics{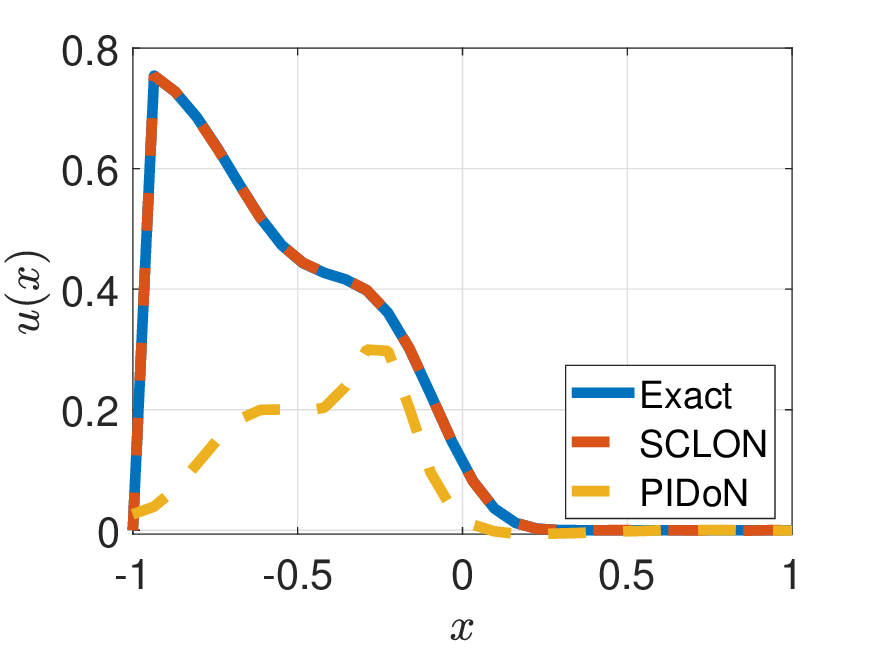}}&\resizebox{0.35\columnwidth}{!}{\includegraphics{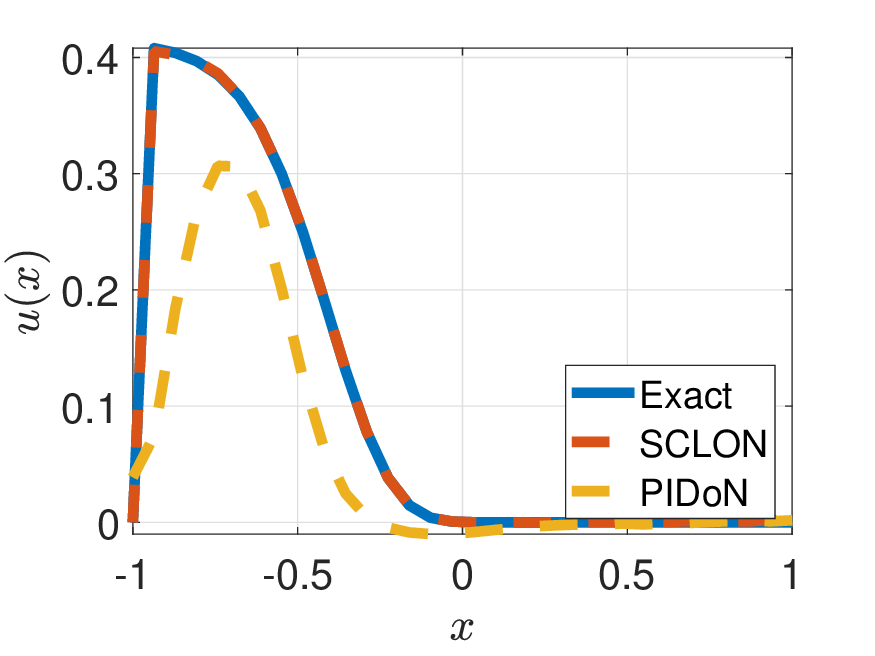}}\\
(g) ${L^\infty} $$error$ for $t=0.25$& (h) ${L^\infty} $$error$ for $t=0.5$& (i) ${L^\infty} $$error$ for $t=1$\\
\resizebox{0.35\columnwidth}{!}{\includegraphics{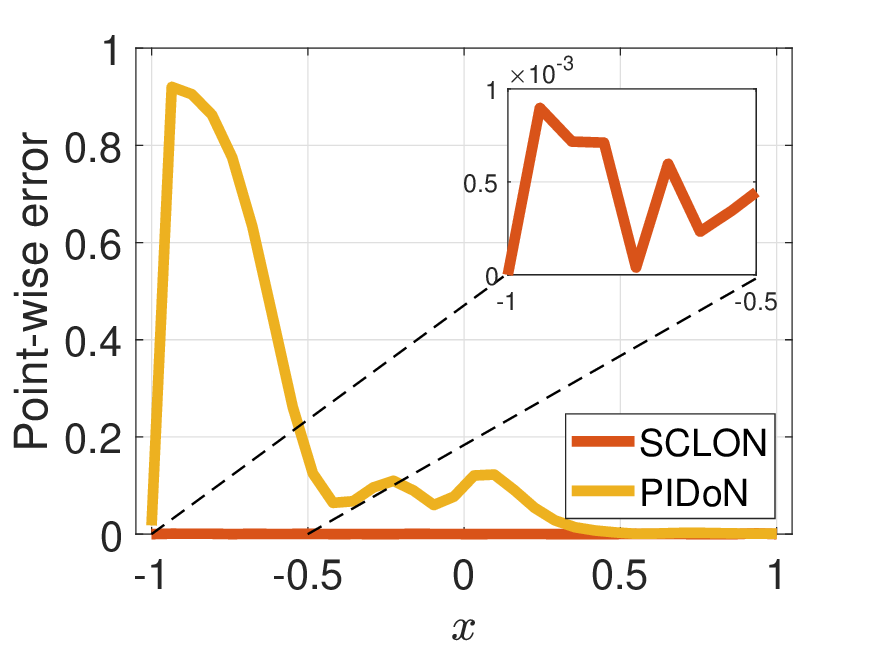}}&\resizebox{0.35\columnwidth}{!}{\includegraphics{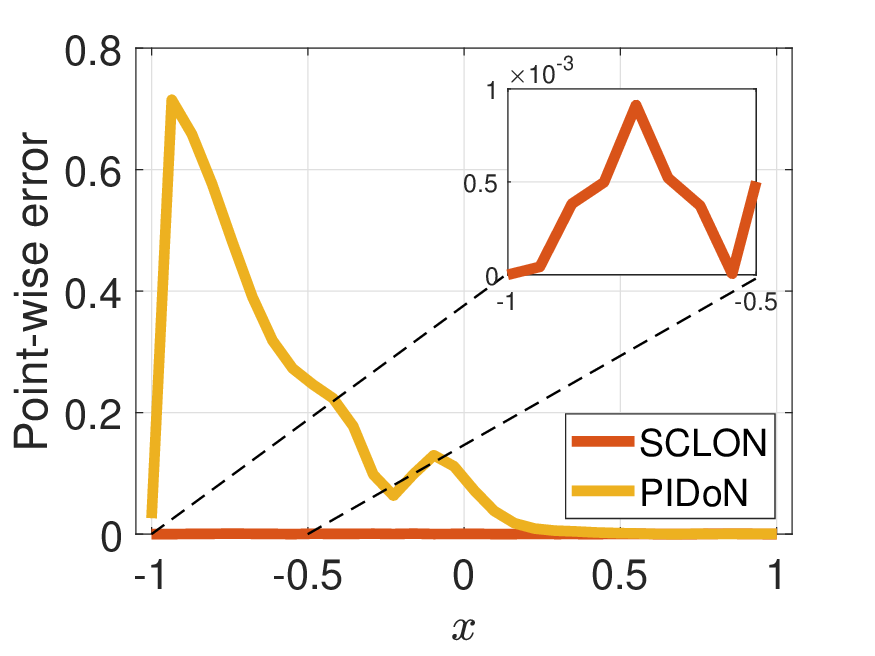}}&\resizebox{0.35\columnwidth}{!}{\includegraphics{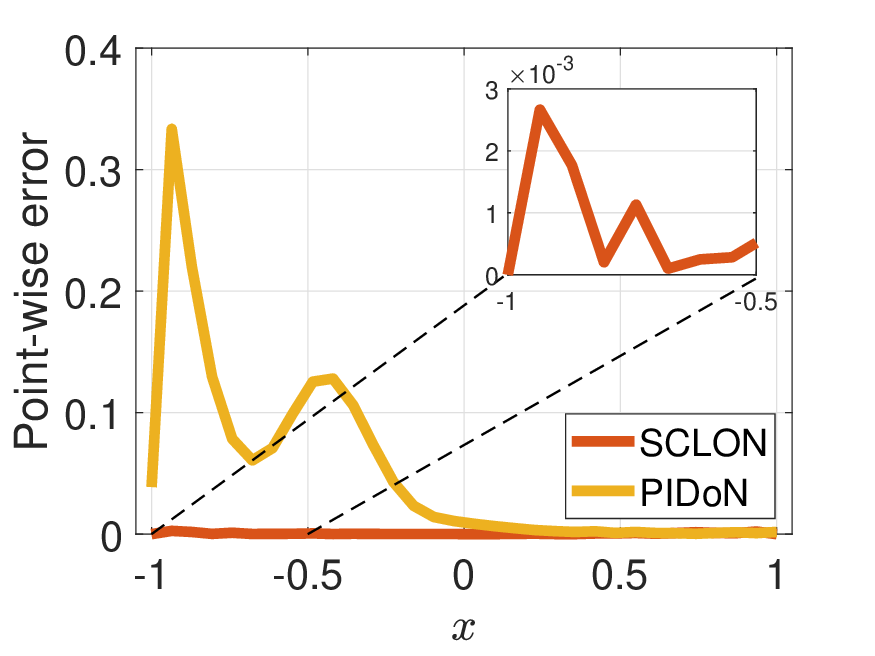}}
\end{tabular}}
\caption{{\bf{Solving a diffusion-convection Equation \eqref{DCE} with $\nu=10^{-6}$. (Top)}} (a) Exact solution versus (b) the prediction of a trained SCLON for a representative example
in the test dataset, and (c) the error in $L^\infty$ norm.  {\bf{(Middle)}} Time slices for Exact solution, the prediction of SCLON(ours), and the one of PIDoN at (d) $t=0.25$, (e) $t=0.5$, and (f) $t=1$. {\bf{(Bottom)}} The error in $L^\infty$ norm between the exact solution and the prediction of SCLON(ours), and between the exact solution and the one of PIDoN at (g) $t=0.25$, (h) $t=0.5$, and (i) $t=1$. For this instance, the error of SCLON is measured as 3.294e-04 in MAE, 1.316e-03 in Rel.:$L^2$, 1.464e-03 in $L^\infty$ whereas the error of PIDoN is 1.370e-01 in MAE, 6.812e-01 in Rel.$L^2$, and 6.539e-01 in $L^\infty$.}
\label{fig:DCE}
\end{center}
\end{figure}

We assess the accuracy of our model predictions using a set of $P=2,000$ new and previously unseen inputs generated by the same procedure described above. Figures \ref{fig:DCE} (a)-(c) display the reference solution and the BE-SCLON prediction for \eqref{DCE} with \(\epsilon = 10^{-6}\) across all times. These figures highlight steep transitions near the boundary layer at \(x=-1\). Remarkably, even without significant mesh refinement, our method effectively captures the sharp transitions of the boundary layer, thanks to the embedded corrector function in the scheme. Notably, when the target function encompasses stiff components, several neural network (NN) algorithms struggle to converge to satisfactory solutions, as they inherently assume smoothness during their learning process. 
Figure \ref{fig:DCE} also provides a comparison of the numerical performance between PIDoN and our method, BE-SCLON, for \eqref{DCE}. 
Specifically, Figures \ref{fig:DCE} (d)-(f) and (g)-(i) offer detailed insights into the PIDoN and BE-SCLON predictions at specific times, respectively, underscoring the superior performance of the BE-SCLON approach. As evidenced in Table \ref{tab:1}, our BE-SCLON model achieves an average relative \(L^2\) error of approximately \(0.27\%\), in stark contrast to the PIDoN's relative \(L^2\) error of \(53.5\%\), representing a significant improvement.

\section{Two-dimensional Kuramoto-Sivashinsky equation}\label{sec_KS}
In this section, we extend the SCLON to tackle two-dimensional nonlinear problems. As a starting point, we examine the Kuramoto-Sivashinsky equations (KSE), given by
\begin{align}\label{KSE}
\begin{split}
&u_t+\Delta u+\Delta^2 u+ |\nabla u|^2=0,\quad\text{for}\quad t>0, x\in \Omega,\\ 
&u=u_0(x),\quad \text{for}\quad t=0,\quad x\in \Omega,
\end{split}
\end{align}
with periodic boundary conditions and $\Omega=(0,2\pi)\times(0,2\pi)$. 
The Kuramoto-Sivashinsky equations (KSE) arise in various physical contexts, including flame propagation, fluid dynamics, and plasma physics. Studying these equations provides insights into the intricate behaviors of nonlinear wave phenomena and pattern formations \cite{KSE01, KSE02}.
Our objective is to employ the proposed SCLON model to learn the solution operator that maps the initial conditions \( u_0(x) \) to the full spatiotemporal solution \( u(x, t) \) of the KSE. It's worth noting that the equation encompasses two-dimensional, highly nonlinear, and high-order PDEs. Consequently, this example suitably demonstrates the SCLON's capability to accurately predict solutions with such characteristics. 

We then employ the two-dimensional FS to obtain a numerical solution written as
\begin{align}
u_N(x_n,y_m)=\frac{1}{(2\pi)^2}\sum_ {\xi_y=-N/2+1}^{N/2}\sum_ {\xi_x=-N/2+1}^{N/2}e^{\mathrm{i}(\xi_x x_n+\xi_yy_m)}\alpha_{\xi_x,\xi_y},\quad n,m=1\cdots,N.
\end{align}
Before applying the two-dimensional FS, we first define our notations. If \( u \in l^2(h\mathbb{Z}^2) \), the DFT in \( \mathbb{R}^2 \) is defined as
\begin{align}\label{2dFTT}
\mathcal{F}_{\xi_x,\xi_y}(u)=h_xh_y\sum_ {n=0}^{N-1}\sum_ {m=0}^{N-1}e^{-\mathrm{i}(\xi_x x_i+\xi_yy_j)}u(x_n,y_m),\quad \xi_x,\xi_y=-\frac{N}{2}+1\cdots,\frac{N}{2}.
\end{align}
In addition, if $\mathcal{F}_{\xi_x,\xi_y}(u)$ is in $L^2([-h/\pi,g/\pi)^2)$, IDFT in $\mathbb{R}^2$ is defined as
\begin{align}\label{idft}
\mathcal{F}_{n,m}^{-1}(\mathcal{F}_{\xi_x,\xi_y}(u))=\frac{1}{(2\pi)^2}\sum_ {\xi_y=-N/2+1}^{N/2}\sum_ {\xi_x=-N/2+1}^{N/2}e^{\mathrm{i}(\xi_x x_n+\xi_yy_m)}\mathcal{F}_{\xi_x,\xi_y}(u),\quad n,m=0\cdots,N-1.
\end{align}
For numerical simulations, we employ the FS combined with the exponential-time-differencing method (ETD) \cite{cox2002exponential} as a temporal numerical scheme for solving \eqref{KSE}. Accordingly, \eqref{KSE} is transformed into equations for \( \mathcal{F}(u^{r+1}) \), given by:
\begin{align}\label{KSE_numerical}
\begin{split}
  \mathcal{F}_{\xi_x,\xi_y}(u^{r+1})=\mathcal{F}_{\xi_x,\xi_y}(u^{r})e^{c\Delta t}&+\left\lbrace G(u^{r})(-4-c\Delta t+e^{c\Delta t}(4-3c\Delta t+(c\Delta t)^2))\right.\\
  &\left.+(2G(\eta_1)+G(\eta_2))(2+c\Delta t+e^{c\Delta t}(-2+c\Delta t))\right.\\
  &\left.+G(\eta_3)(-4-3c\Delta t-(c\Delta t)^2e^{c\Delta t}(4-c\Delta t))\right\rbrace/h^2c^3
  \end{split}
\end{align}
where 
\begin{align}\label{KSE_numerical2}
\begin{split}
    \eta_1&=\mathcal{F}_{\xi_x,\xi_y}(u^{r})e^{c\Delta t/2}+\left(e^{c\Delta t/2}-1 \right)G(u^{r})/c,\\
    \eta_2&=\mathcal{F}_{\xi_x,\xi_y}(u^{r})e^{c\Delta t/2}+\left(e^{c\Delta t/2}-1 \right)G(\eta_1)/c,\\
    \eta_3&=\eta_1e^{c\Delta t/2}+\left(e^{c\Delta t/2}-1 \right)\left(2G(\eta_2)-G(u^{r})\right)/c,
    \end{split}
\end{align}
for 
\begin{align}
    c & =-(\xi_x^2+\xi_y^2)^2-(\xi_x^2+\xi_y^2),\\
    G(u) & =\mathcal{F}_{\xi_x,\xi_y}(\mathcal{F}_n^{-1}(\mathrm{i}\xi_x\mathcal{F}_{\xi_x}(u))^2+\mathcal{F}_m^{-1}(\mathrm{i}\xi_y\mathcal{F}_{\xi_y}(u))^2).
\end{align}

After finding $\mathcal{F}_{\xi_x,\xi_y}(u^{r+1})$ at \eqref{KSE_numerical}, a numerical solution at $t=(r+1)\Delta t$ is obtained by
\begin{align}
u_N^{r+1}(x_n,y_m)=\frac{1}{(2\pi)^2}\sum_ {\xi_y=-N/2+1}^{N/2}\sum_ {\xi_x=-N/2+1}^{N/2}e^{\mathrm{i}(\xi_x x_n+\xi_yy_m)}\mathcal{F}_{\xi_x,\xi_y}(u^{r+1}),\quad n,m=0\cdots,N-1.
\end{align}

We now present how the SCLON replicates the procedure of FS and ETD in the same fashion. We use \( P = 2000 \) initial conditions, which are generated from a GRF distributed as \( \sim \mathcal{N}(0,4^2 (-\Delta + 2^2 I)^{-5/2}) \). This GRF, employed in \cite{PIDON}, satisfies the periodic boundary conditions.

Let \( u_{0p}(x,y) \) be one of the \( P \) input data. We then construct a network, \( \mathcal{G}_q \), which produces \( \{\widehat{\alpha}^{r}_{\xi,p}\}_{r=(q-1)R}^{qR-1} \subset \mathbb{C} \) for time \( t \) ranging from \( t=(q-1)R\Delta t \) to \( t=(qR-1) \Delta t \) given the initial condition \( u_{0p} \). Then, the prediction is constructed as 
\begin{align}
\widehat{u}_N^{r}(x_n,y_m)=\frac{1}{(2\pi)^2}\sum_ {\xi_y=-N/2+1}^{N/2}\sum_ {\xi_x=-N/2+1}^{N/2}e^{\mathrm{i}(\xi_x x_n+\xi_yy_m)}\widehat{\alpha}_{\xi_x,\xi_y}^r,\quad n,m=0\cdots,N-1,
\end{align}
for $(q-1)R\leq r\leq qR-1$.
Afterward, we substitute \( \widehat{u}_N^{r} \) and \( \widehat{\alpha}_{\xi_x,\xi_y}^r \) to \( u_N^r \) and \( \mathcal{F}_{\xi_x,\xi_y}(u^{r}) \), respectively in \eqref{KSE_numerical2} to define a loss function that mimics the procedure of FS and ETD as follows: 

\begin{align} \label{e:KS_loss}
 loss_q=  \sum_{r=(q-1)R}^{qR-1}\sum_{p=1}^{P}\sum_{\xi_y=-N/2+1}^{N/2}\sum_{\xi_x=-N/2+1}^{N/2} &\left| \widehat{\alpha}_{\xi_x,\xi_y, p}^{r+1}-\widehat{\alpha}_{\xi_x,\xi_y, p}^{r}\right.\\
 &-\left\lbrace\left.G( \widehat{u}_p^{r})(-4-c\Delta t+e^{c\Delta t}(4-3c\Delta t+(c\Delta t)^2))\right.\right.\nonumber\\
  &\left.+(2G( \widehat{\eta}_{1,p})+G(\eta_2))(2+c\Delta t+e^{c\Delta t}(-2+c\Delta t))\right.\nonumber\\
  &\left.\left.+G( \widehat{\eta}_{3,p})(-4-3c\Delta t-(c\Delta t)^2e^{c\Delta t}(4-c\Delta t))\right\rbrace/h^2c^3  \right|^2\nonumber
\end{align}
where 
\begin{align*}
\widehat{\eta}_{1,p}&=\mathcal{F}_{\xi_x,\xi_y}(\widehat{u}_p^{r})e^{c\Delta t/2}+\left(e^{c\Delta t/2}-1 \right)G(\widehat{u}_p^{r})/c,\\
\widehat{\eta}_{2,p}&=\mathcal{F}_{\xi_x,\xi_y}(\widehat{u}_p^{r})e^{c\Delta t/2}+\left(e^{c\Delta t/2}-1 \right)G(\widehat{\eta}_{1,p})/c,\\
\widehat{\eta}_{3,p}&=\eta_1e^{c\Delta t/2}+\left(e^{c\Delta t/2}-1 \right)\left(2G(\widehat{\eta}_{2,p})-G(\widehat{u}_p^{r})\right)/c.
\end{align*}
Finally, we train the model \( \mathcal{G}_q \) until the loss function \eqref{e:KS_loss} remains flat during training.

\begin{center}
\begin{figure}[h]
\setlength{\tabcolsep}{0.00001pt}
\makebox[\textwidth][c]{\begin{tabular}{ ccc }
(a) Exact at $t=0.1$ & (b) Exact at $t=0.25$ & (c) Exact at $t=0.5$\\
\resizebox{0.35\columnwidth}{!}{\includegraphics{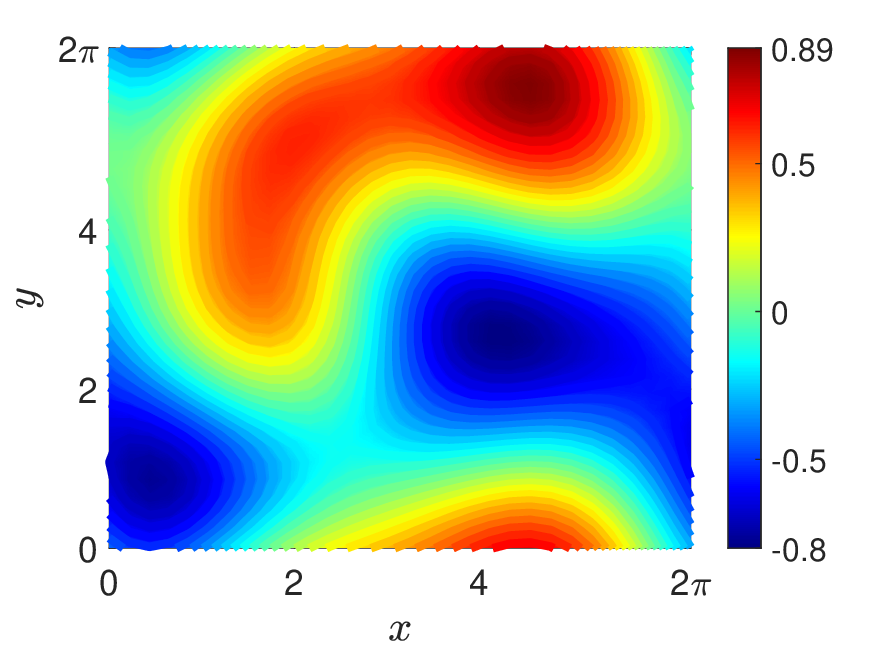}}&\resizebox{0.35\columnwidth}{!}{\includegraphics{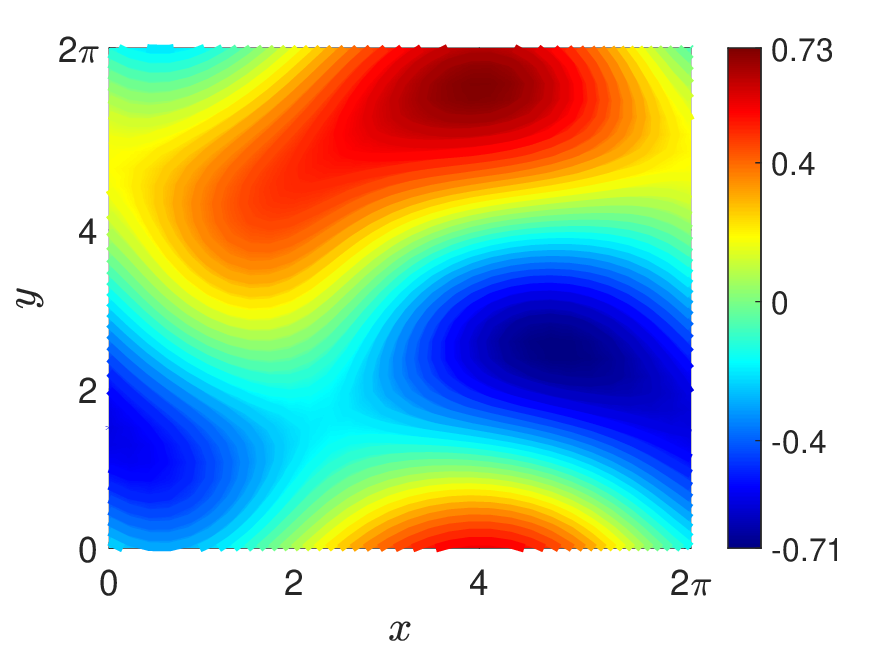}}&\resizebox{0.35\columnwidth}{!}{\includegraphics{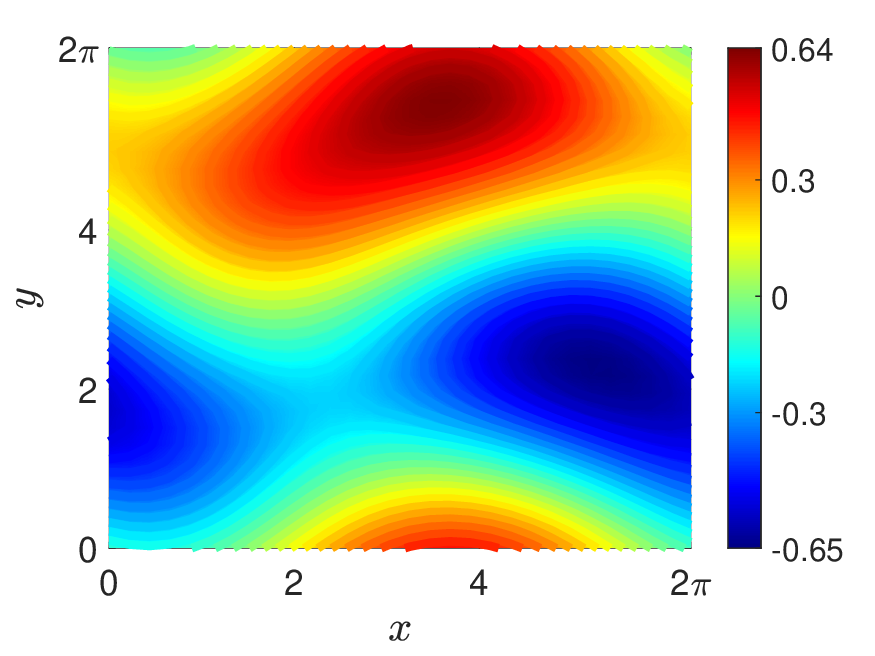}}\\
(d) Prediction at $t=0.1$& (e) Prediction at $t=0.25$ & (f) Prediction at $t=0.5$\\
\resizebox{0.35\columnwidth}{!}{\includegraphics{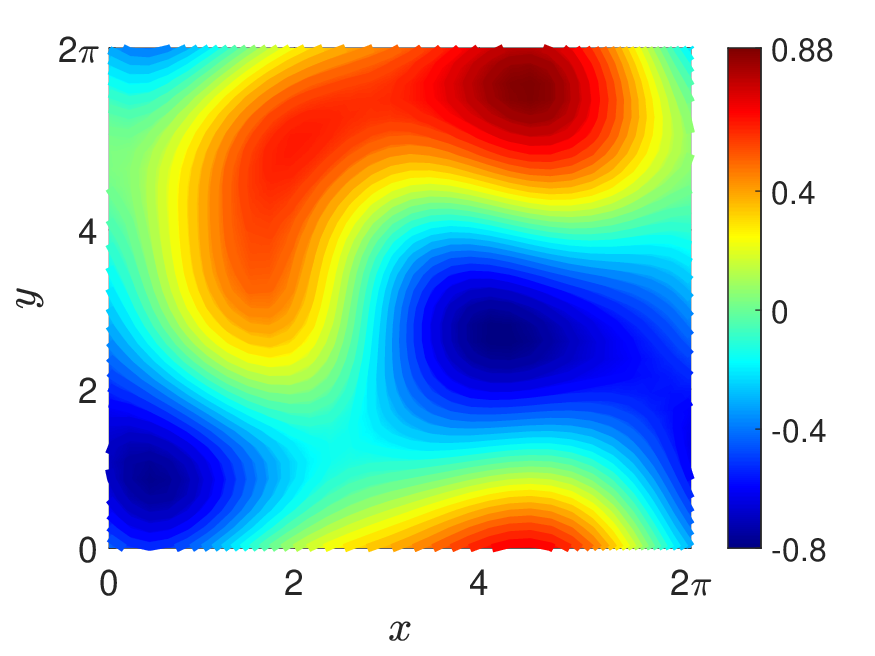}}&\resizebox{0.35\columnwidth}{!}{\includegraphics{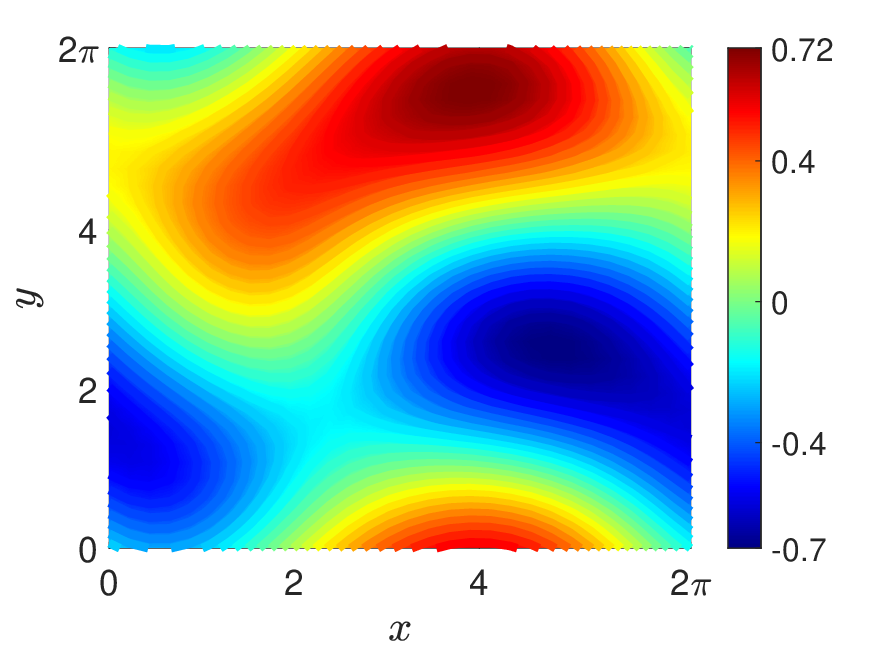}}&\resizebox{0.35\columnwidth}{!}{\includegraphics{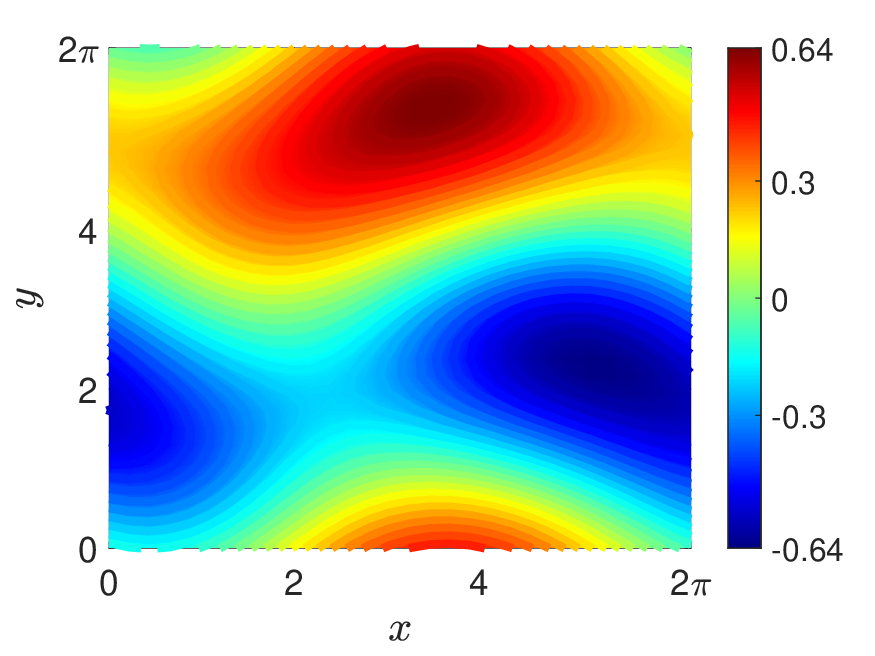}}\\
(g) ${L^\infty} $$error$ at $t=0.1$& (h) ${L^\infty} $$error$ at $t=0.25$& (i) ${L^\infty} $$error$ at $t=0.5$\\
\resizebox{0.35\columnwidth}{!}{\includegraphics{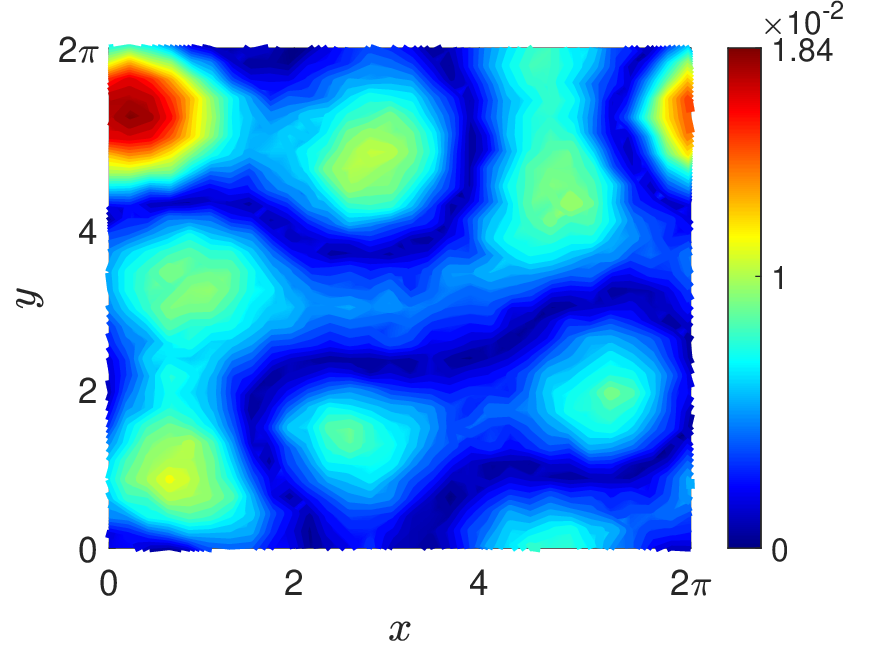}}&\resizebox{0.35\columnwidth}{!}{\includegraphics{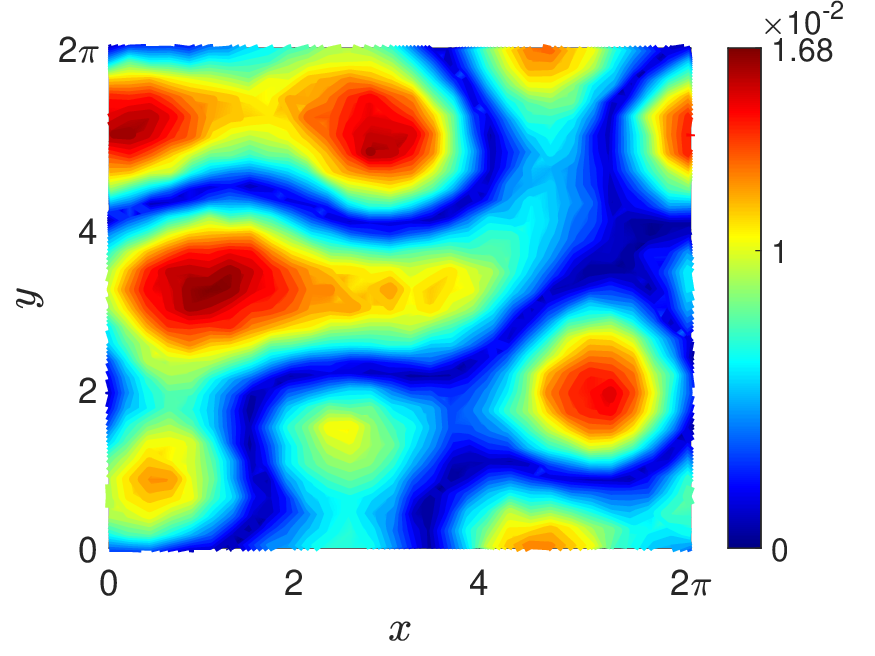}}&\resizebox{0.35\columnwidth}{!}{\includegraphics{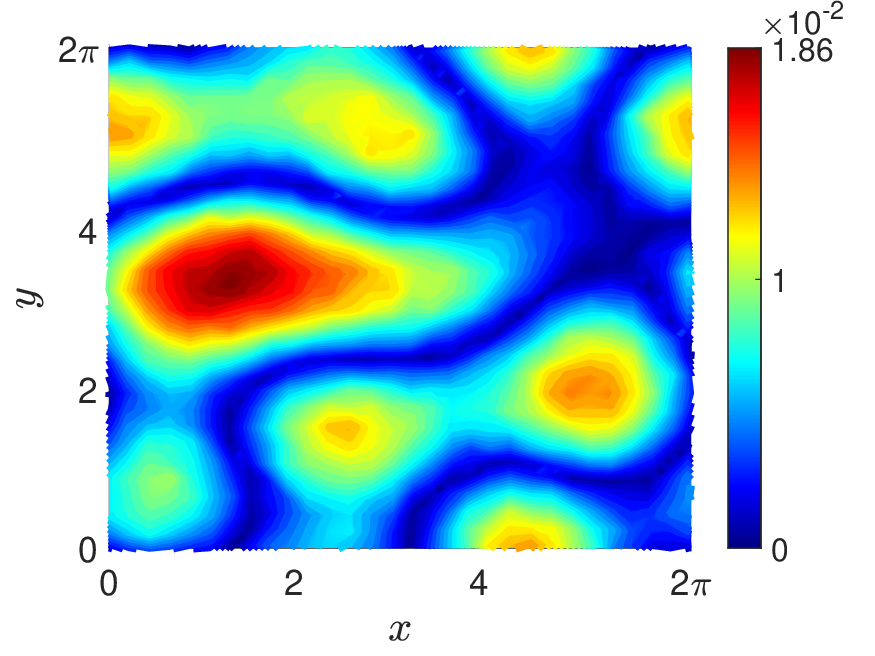}}
\end{tabular}}
\caption{{\bf{2-dimensional Kuramoto-Sivashinsky equations}}\eqref{KSE}. {\bf{(Top)}} Exact solution at $t=0.1$, $0.25$, and $0.5$. {\bf{(Middle)}} the prediction of a trained SCLON for a representative example in the test dataset. {\bf{(Bottom)}} The error in $L^\infty$ norm between the exact solution and the prediction of SCLON. The error for this instance is $5.83e-03$ in MAE, $1.78e-02$ in Rel.$L^2$, and $1.66e-02$ in $L^{\infty}$.}
\label{fig:KS}
\end{figure}
\end{center}

We assess prediction accuracy using new initial conditions that were not part of the model training. The resulting average relative \( L^2 \) prediction across all test dataset examples is \( 1.19\% \) (see Table \ref{tab:1}). This leads us to conclude that the SCLON model serves as an effective surrogate for two-dimensional nonlinear PDEs. Figure \ref{fig:KS} demonstrates the exact solutions, predicted solutions, and \( L^\infty \) errors between them at various time steps for the test initial condition. This example illustrates that the predicted solution aligns well with the reference solution, with the SCLON effectively capturing the nonlinear dynamics.

\begin{figure}[h!]
\begin{center}
\setlength{\tabcolsep}{0.00001pt}
\makebox[\textwidth][c]{\begin{tabular}{ ccc }
(a) Exact at $t=0.1$ & (b) Exact at $t=0.25$ & (c) Exact at $t=0.5$\\
\resizebox{0.35\columnwidth}{!}{\includegraphics{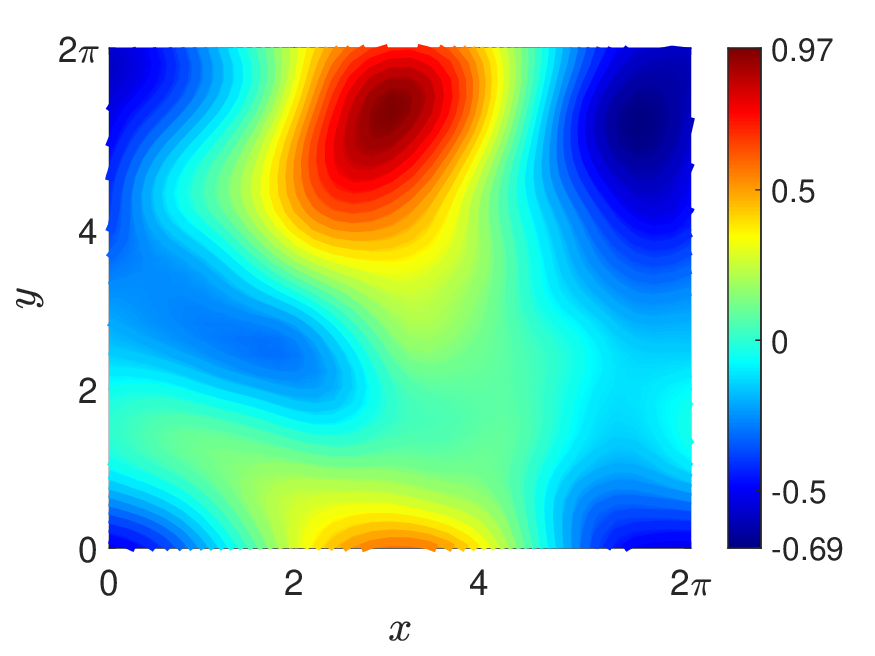}}&\resizebox{0.35\columnwidth}{!}{\includegraphics{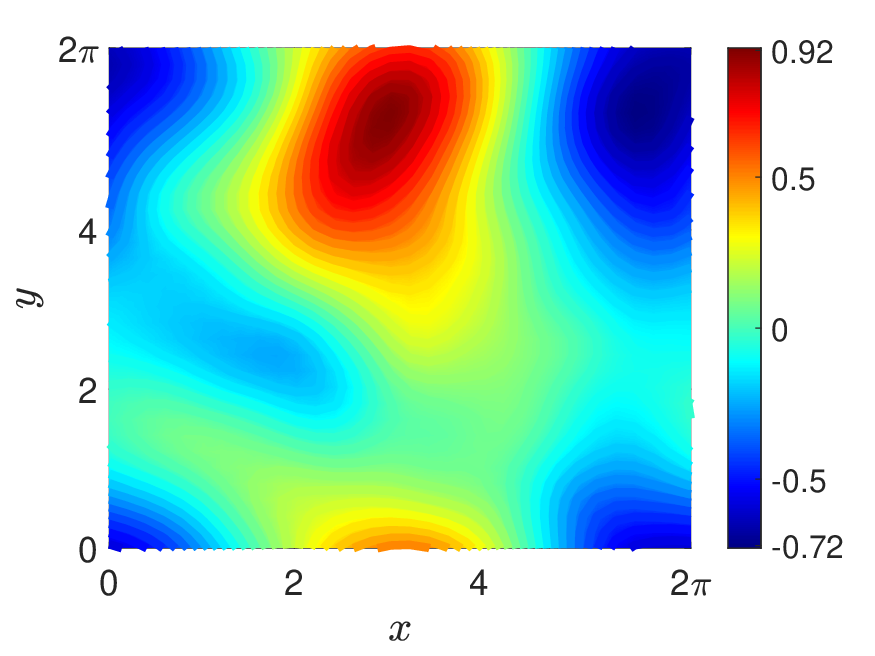}}&\resizebox{0.35\columnwidth}{!}{\includegraphics{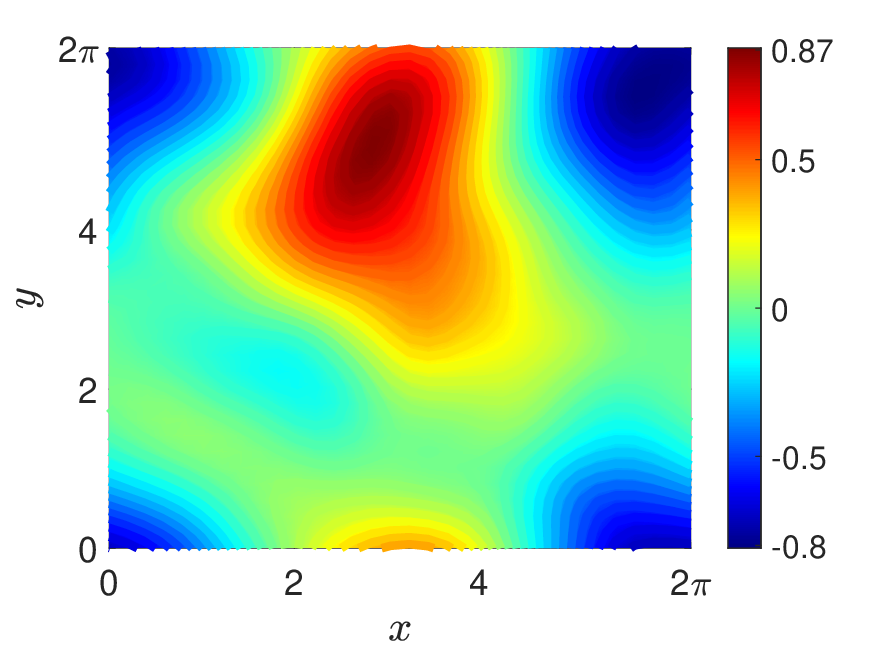}}\\
(d) Prediction at $t=0.1$& (e) Prediction at $t=0.25$ & (f) Prediction at $t=0.5$\\
\resizebox{0.35\columnwidth}{!}{\includegraphics{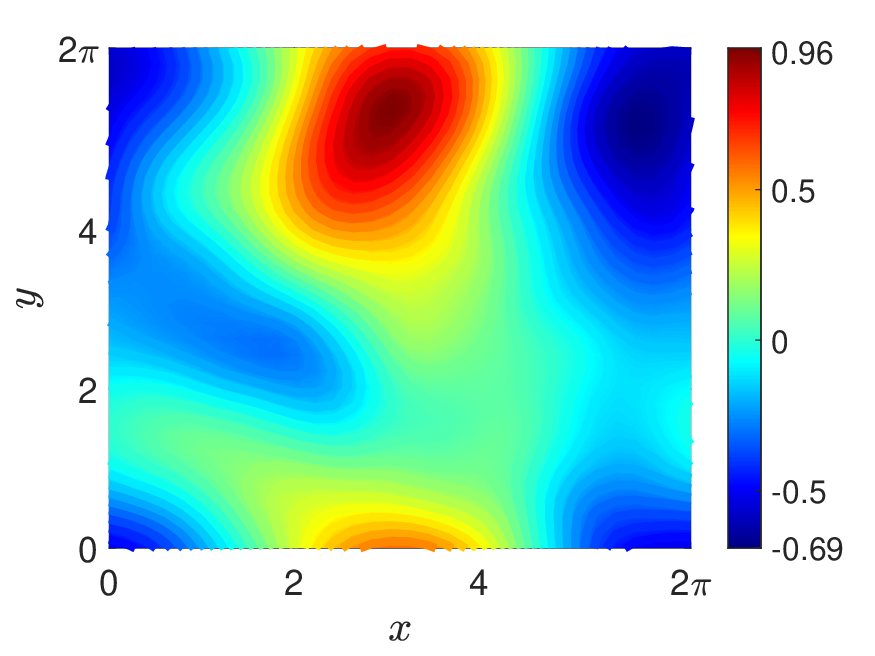}}&\resizebox{0.35\columnwidth}{!}{\includegraphics{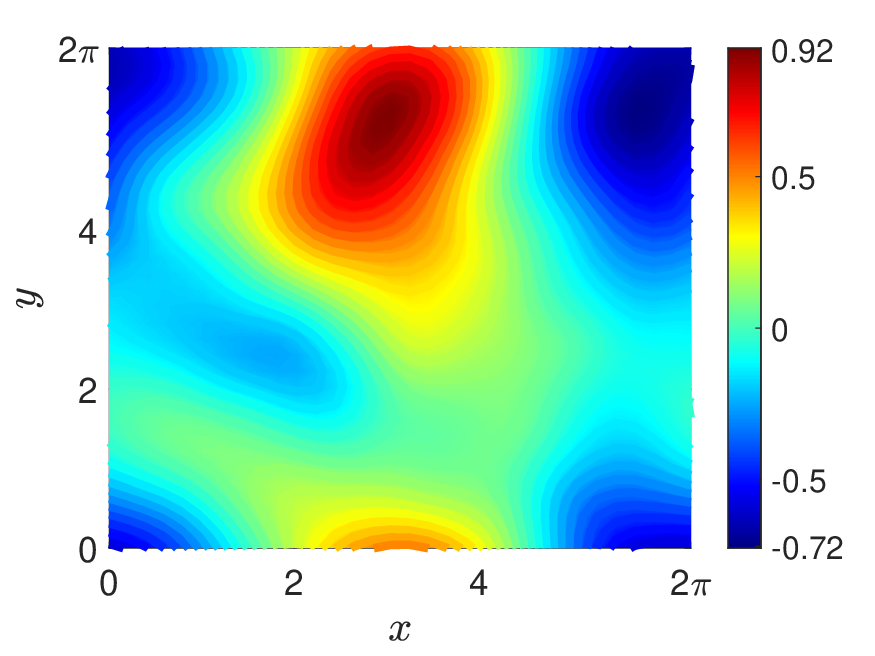}}&\resizebox{0.35\columnwidth}{!}{\includegraphics{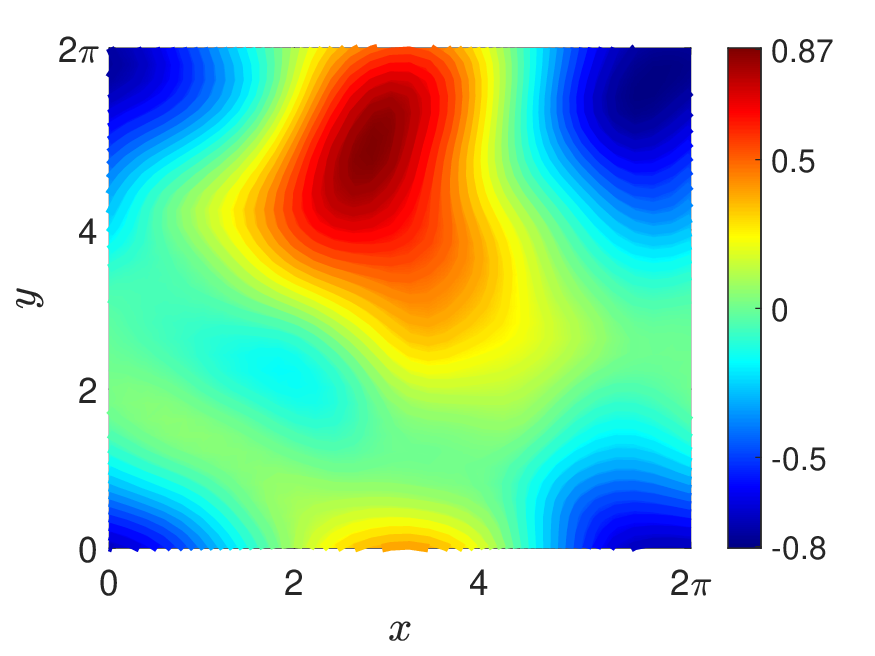}}\\
(g) ${L^\infty} $$error$ at $t=0.1$& (h) ${L^\infty} $$error$ at $t=0.25$& (i) ${L^\infty} $$error$ at $t=0.5$\\
\resizebox{0.35\columnwidth}{!}{\includegraphics{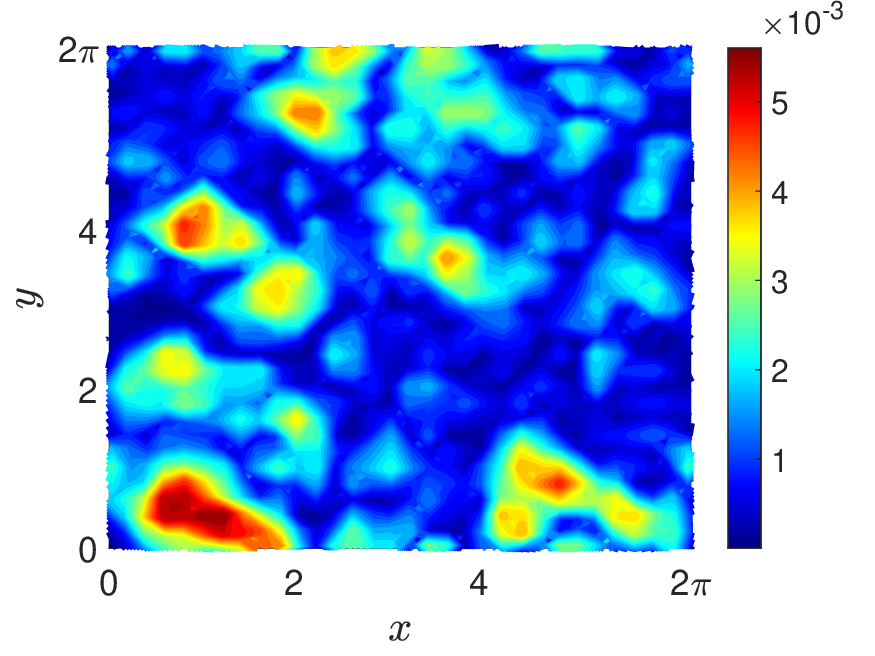}}&\resizebox{0.35\columnwidth}{!}{\includegraphics{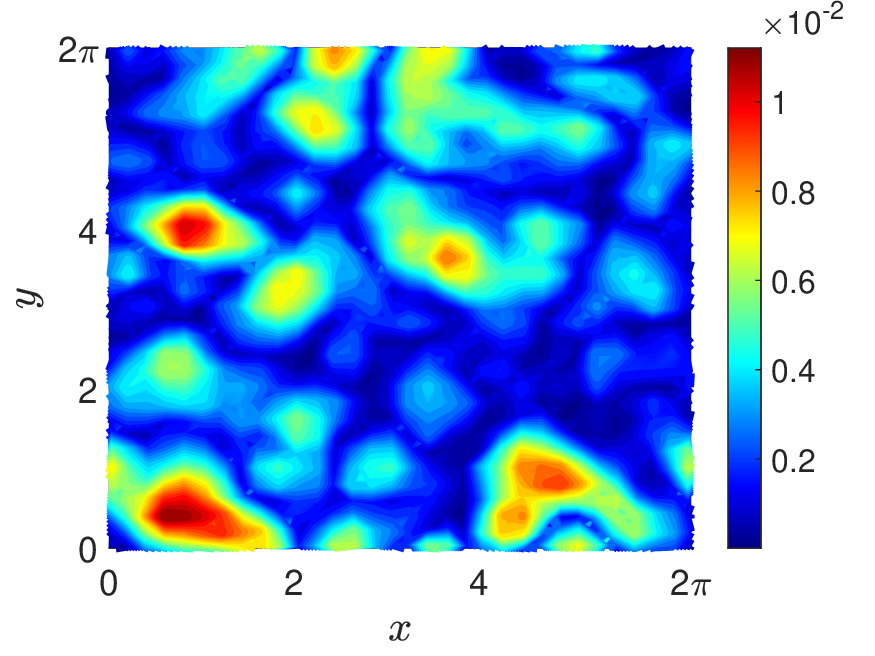}}&\resizebox{0.35\columnwidth}{!}{\includegraphics{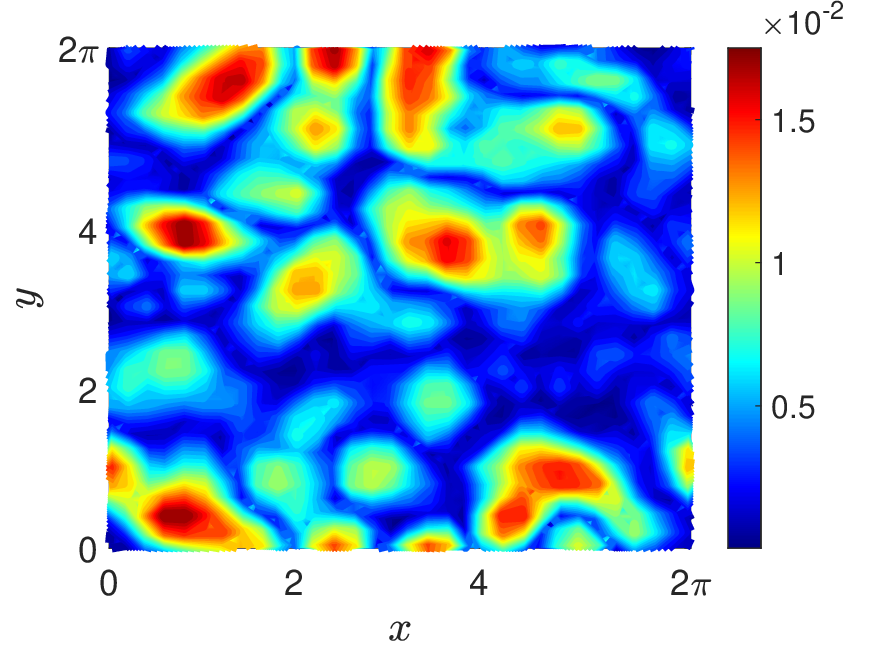}}
\end{tabular}}
\caption{{\bf{2-Dimensional Navier-Stokes Equation}} \eqref{NSE}. We set $Re=200$, $\Delta t=0.01$, and $T_{end}=0.5$. {\bf{(Top)}} Exact solution at $t=0.1$, $0.25$, and $0.5$. {\bf{(Middle)}} The prediction of a trained SCLON for a representative example in the test dataset. {\bf{(Bottom)}} The error in $L^\infty$ norm between the exact solution and the prediction of SCLON. The error for this instance is $2.766e-03$ in MAE, $1.121e-02$ in Rel.$L^2$, and $1.068e-02$ in $L^{\infty}$. }\label{fig:NS}
\end{center}
\end{figure}

\section{Two-dimensional Navier-Stokes equations} \label{sec_NE}
Finally, we take on the challenge of solving the incompressible Navier–Stokes equation (NSE) in its vorticity form for a viscous, incompressible fluid on the unit torus, 
\begin{align}\label{NSE}
\begin{split}
&w_t+\bold{u}\cdot\nabla w-\frac{1}{Re}\Delta w=f,\quad\text{for}\quad t>0, x\in \Omega,\\ 
&\bold{u}=\nabla \times \psi\widehat{\bold{z}}\\
&\Delta \psi=w\\
&w=w_0(x),\quad \text{for}\quad t=0,\quad x\in \Omega,
\end{split}
\end{align}
where $\bold{u}=(u,v)$ is a velocity field, $f$ is the forcing function, and $Re$ is a Reynolds number. 
Our objective is to learn the solution operator that maps the vorticity of the initial condition to the complete solution in time and space.
The vorticity, denoted by \( \omega = \nabla \times {\bf u} \), plays a fundamental role in fluid dynamics. 
It is also crucial in the mathematical analysis of the Navier-Stokes equations. In many instances, it is advantageous to describe the dynamics of a flow in terms of the evolution of its vorticity. 
We employ the two-dimensional FS to obtain a numerical solution written as
\begin{align}
w_N(x_n,y_m)=\frac{1}{(2\pi)^2}\sum_ {\xi_y=-N/2+1}^{N/2}\sum_ {\xi_x=-N/2+1}^{N/2}e^{\mathrm{i}(\xi_x x_n+\xi_yy_m)}\alpha_{\xi_x,\xi_y},\quad n,m=1\cdots,N.
\end{align}
To numerically solve \eqref{NSE}, we employ the Crank-Nicolson and Heun's methods for the temporal domain, and the Fourier spectral method for the spatial domain \cite{chandler2013invariant}. Both methods consist of two steps; for more details, see \cite{chandler2013invariant}.
We start by the first step:
\begin{align}\label{NSE_weak1}
   \frac{ \mathcal{F}_{\xi_x,\xi_y}(\widetilde{w})-\mathcal{F}_{\xi_x,\xi_y}({w}^{r})}{\Delta t}+\frac{|\bold{k}|^2}{2Re}(\mathcal{F}_{\xi_x,\xi_y}(\widetilde{w})+\mathcal{F}_{\xi_x,\xi_y}({w}^{r})+\mathrm{i}\bold{k}\cdot\left(\mathcal{F}_{\xi_x,\xi_y}({u^{r}w^{r}}),\mathcal{F}_{\xi_x,\xi_y}({v^{r}w^{r}})\right)-\mathcal{F}_{\xi_x,\xi_y}({f})=0.
\end{align}
After solving \eqref{NSE_weak1} for \( \mathcal{F}_{\xi_x,\xi_y}(\widetilde{w}) \), we take the IDFT \eqref{idft} on \( \mathcal{F}_{\xi_x,\xi_y}(\widetilde{w}) \) to obtain \( \widetilde{w} \). Next, we solve for \( \widetilde{\bold{u}}=(\widetilde{u},\widetilde{v}) \) using the equation: 
\begin{align}
&\Delta \widetilde{\psi}=\widetilde{w}\\
   &\widetilde{\bold{u}}=\nabla \times \widetilde{\psi}\widehat{\bold{z}}.
\end{align}

Subsequently, $\widetilde{w}$ and $\widetilde{\bold{u}}$ are used to solve the second step,
\begin{align}\label{NSE_weak2}
          \frac{\mathcal{F}_{\xi_x,\xi_y}({w}^{r+1})-\mathcal{F}_{\xi_x,\xi_y}({w}^{r})}{\Delta t}=&-\frac{|\bold{k}|^2}{2Re}(\mathcal{F}_{\xi_x,\xi_y}({w}^{r+1})+\mathcal{F}_{\xi_x,\xi_y}({w}^{r})\\
          &-\frac{\mathrm{i}\bold{k}}{2}\cdot\left((\mathcal{F}({u^{r}w^{r}}),\mathcal{F}_{\xi_x,\xi_y}({v^{r}w^{r}}))+(\mathcal{F}_{\xi_x,\xi_y}(\widetilde{u}\widetilde{w}),\mathcal{F}_{\xi_x,\xi_y}(\widetilde{v}\widetilde{w}))\right)+\mathcal{F}_{\xi_x,\xi_y}({f}),\nonumber
\end{align}
from which the Fourier coefficient $\mathcal{F}_{\xi_x,\xi_y}({w}^{r+1})$ at $t=(r+1)\Delta t$ is obtained. As a result, $w_{N+1}^{r+1}$ can be constructed as
\begin{align}
w_N^{r+1}(x_n,y_m)=\frac{1}{(2\pi)^2}\sum_ {\xi_y=-N/2+1}^{N/2}\sum_ {\xi_x=-N/2+1}^{N/2}e^{\mathrm{i}(\xi_x x_n+\xi_yy_m)}\mathcal{F}_{\xi_x,\xi_y}(w^{r+1}),\quad n,m=0\cdots,N-1.
\end{align}

Following the procedure of the numerical scheme, we design the SCLON to replicate this approach. 
First, we generate \( P=2000 \) initial conditions \( u_0(x,y) \) as training input data. 
These are derived from the GRF 
\( \sim \mathcal{N}(0,9^2 (-\Delta + 3^2 I)^{-5/2}) \)
as employed in \cite{PINO}, and they satisfy the periodic boundary conditions.

Let \( u_{0p}(x,y) \) be one of the \( P \) inputs data. We then design a network, \( \mathcal{G}_q \) that produces the coefficients \( \{\widehat{\alpha}^{r}_{\xi_X,\xi_y,p}\}_{r=(q-1)R}^{qR-1} \) in the complex domain for times ranging from \( t=(q-1)R \Delta t \) to \( t=(qR-1)\Delta t \), based on the initial condition \( u_{0p} \). Subsequently, the predicted coefficients \( \widehat{\alpha}^{r}_{\xi_x,\xi_y,p} \) are utilized to construct the following:
\begin{align}
\widehat{w}_{N,p}^{r}(x_n,y_m)=\frac{1}{(2\pi)^2}\sum_ {\xi_y=-N/2+1}^{N/2}\sum_ {\xi_x=-N/2+1}^{N/2}e^{\mathrm{i}(\xi_x x_n+\xi_yy_m)}\widehat{\alpha}_{\xi_x,\xi_y}^r,\quad n,m=0\cdots,N-1.
\end{align}
We then compute  $\widehat{\bold{u}}^{r}=(\widehat{u}^{r},\widehat{v}^{r})$ from
\begin{align}
&\Delta \widehat{\psi}=\mathcal{F}^{-1}(\widehat{\alpha}^r_{\xi_x,\xi_y,p}),\\
   &\widehat{\bold{u}}^r=\nabla \times \widehat{\psi}\widehat{\bold{z}}.
\end{align} 
Next, in \eqref{NSE_weak1}, we replace \( \widehat{w}_{N,p}^{r} \) with \( w_N^r \) and \( \widehat{\alpha}_{\xi_x,\xi_y,p}^r \) with \( \mathcal{F}_{\xi_x,\xi_y}(w^{r}) \), resulting in
\begin{align}
   \frac{ \mathcal{F}_{\xi_x,\xi_y}(\widetilde{w})-\widehat{\alpha}^{r}_{\xi_x,\xi_y}}{\Delta t}=-\frac{|\bold{k}|^2}{2Re}(\mathcal{F}_{\xi_x,\xi_y}(\widetilde{w})+\widehat{\alpha}^{r}_{\xi_x,\xi_y}-\mathrm{i}\bold{k}\cdot\left(\mathcal{F}_{\xi_x,\xi_y}({ \widehat{u}^r \widehat{w}^r}),\mathcal{F}_{\xi_x,\xi_y}( \widehat{v}^r \widehat{w}^r)\right)+\mathcal{F}_{\xi_x,\xi_y}({f}),
\end{align}
to find $\mathcal{F}_{\xi_x,\xi_y}(\widetilde{w})$. 
After taking the IDFT \eqref{idft} on \( \mathcal{F}_{\xi_x,\xi_y}(\widetilde{w}) \) to obtain \( \widetilde{w} \), we solve for \( \widetilde{\bold{u}} = (\widetilde{u},\widetilde{v}) \) such that
\begin{align}
&\Delta \widetilde{\psi}=\widetilde{w}\\
   &\widetilde{\bold{u}}=\nabla \times \widetilde{\psi}\widehat{\bold{z}}.
\end{align}
Consequently, once substituting \( \widehat{\alpha}_{N,p}^{r} \), \( \widehat{\alpha}_{N,p}^{r+1} \), $\widetilde{u}$, $\widetilde{v}$ to \eqref{NSE_weak2}, the loss can be defined as
\begin{align} \label{e:NSE_loss}
         loss_q=&\sum_{r=(q-1)R}^{qR-1}\sum_{p=1}^{P}\sum_{\xi_y=-N/2+1}^{N/2} \sum_{\xi_x=-N/2+1}^{N/2}\left|\frac{\widehat{\alpha}^{r+1}_{\xi_x,\xi_y,p}-\widehat{\alpha}^{r}_{\xi_x,\xi_y,p}}{\Delta t}+\frac{|\bold{k}|^2}{2Re}(\widehat{\alpha}^{r+1}_{\xi_x,\xi_y,p}+\widehat{\alpha}^{r}_{\xi_x,\xi_y,p})\right.\\
         &\left.+\frac{\mathrm{i}\bold{k}}{2}\cdot\left((\mathcal{F}_{\xi_x,\xi_y}({\widehat{u}^r_p\widehat{w}^r_p}),\mathcal{F}_{\xi_x,\xi_y}({\widehat{v}^r_p\widehat{w}^r_p}))+(\mathcal{F}_{\xi_x,\xi_y}({\widetilde{u}\widetilde{w}}),\mathcal{F}_{\xi_x,\xi_y}(\widetilde{v}\widetilde{w}))\right)-\mathcal{F}_{\xi_x,\xi_y}(f)\right|^2.\nonumber
\end{align}
We train the model \( \mathcal{G}_q \) until the loss function \eqref{e:NSE_loss} shows negligible changes during the training. After training, we anticipate that \( \widehat{w}_{N,p}^{r+1} \) 
should more closely approximate \( w_{N,p}^{r+1} \).

To assess the accuracy of our predictions, we test the SCLON using new initial conditions that were not used during the model training process. Specifically, in this example, we examine the chaotic Kolmogorov flow with \( Re = 200 \) and \( f = -n\cos(n y) \), where \( n = 1 \). We benchmark our method against the recent advancement, the PINO, proposed by Li et al. \cite{PINO}, as detailed in Table \ref{tab:2}. Like our approach, this model is trained without paired input-output data and relies solely on the knowledge of the governing equation and its initial or boundary conditions, aligning it closely with our task. The resulting average relative $L^2$ prediction over the test dataset is $2.23\%$, as shown in Table \ref{tab:2}, while the PINO model reports a relative $L^2$ error of $47.8\%$ in \cite{PINO}. Figure \ref{fig:NS} presents the exact solutions, predicted solutions, and \( L^\infty \) errors between them at various time steps for the test initial condition. This example underscores the close alignment of the predicted solution with the reference, demonstrating the SCLON's ability to effectively capture the nonlinear dynamics.

\section{Concluding Remark}
In this paper, we introduce the SCLON, a deep learning framework uniquely designed to approximate nonlinear operators in infinite-dimensional Banach spaces using Fourier-spectral or Legendre-spectral approximations.
This work advances the field of operator learning methods, aiming to bridge the gap between traditional PDE solvers and contemporary deep learning techniques. Our method demonstrates simplicity and robust efficacy for solving parametric PDEs, achieving notable advancements in predictive accuracy, generalization performance, and data efficiency over prevailing techniques.
Importantly, the SCLON presents an ability to learn the solution operator of parametric PDEs without the need for paired input-output training data, effectively predicting solutions even with pronounced singular behavior in thin boundary layers.
This capability presents a significant stride forward, offering a new avenue for simulating nonlinear, singular, and chaotic phenomena across a broad spectrum of scientific and engineering contexts.

Despite the initial promising results presented in this work, several technical questions remain open and require further investigation.
For example, it remains unclear what the convergence properties of the SCLON approach are. 
Recently, Ko et al. \cite{ko2022convergence} proposed a convergence analysis for the spectral coefficient learning method for stationary PDEs, which could potentially be adapted to provide a convergence result for the SCLON method for learning solution operators.
An additional question that arises is the potential for prediction on complex domains. While the SCLON method is based on the spectral method for solving PDEs, its applicability may be limited to certain types of domains, such as circular or spherical domains. For general smooth or polygonal domains, the method can be extended to include finite element coefficient learning. These challenges can be addressed in the future by developing more specialized architectures such as graph neural networks that are specifically designed to capture the dynamic behavior of a given PDE, as well as by implementing more effective numerical algorithms such as finite element or finite volume methods for training and optimization.

\section*{Acknowledgments}
The work of Y. Hong was supported by Basic Science Research Program through the National Research Foundation of Korea (NRF) funded by the Ministry of Education (NRF-2021R1A2C1093579) and Korean Government (MSIT) (2022R1A4A3033571). This study was supported by the National Research Foundation of Korea (NRF) grant funded by the Korean government (MSIT) (No. 2022R1C1C1009387, No. 2022R1A4A3033320). This study was also supported by the National Supercomputing Center with Supercomputing Resources, including technical support (KSC-2022-CRE-0213).

\bibliographystyle{unsrt}
\bibliography{ULGNet_ref}  

\clearpage
\appendix
\section*{Appendix}
\section{Nomenclature}
\begin{table}[h!]\label{notation}
\begin{center}
\begin{tabular}{c l}
 \hline
 Notation & Description \\ \addlinespace
 \hline
 $u(\cdot)$, $w(\cdot)$ &  A solution to a parametric PDE\\\addlinespace
 $\widehat{u}(\cdot)$,$\widehat{w}(\cdot)$ & a prediction of a solution to PDE \\\addlinespace
  $\alpha$ & A coefficient of a basis function (see \eqref{sm})\\\addlinespace
 $\widehat{\alpha}(\cdot)$ & An output of a network which predicts $\alpha$ (see \eqref{network})  \\
\addlinespace
  $x_n$ & The $n$th nodal point on a spatial domain.  \\\addlinespace
  $\xi_n$ & The $n$th wave number in Fourier space.  \\\addlinespace
 $\phi_n(x)$ & A basis function of $n$th order consisting of Legendre polynomials (see \eqref{Le_bases})  \\\addlinespace
 $e^{\rm{i}\xi_n x}$ & A Fourier basis function \\ 
\addlinespace
 $P$ & The number of input data \\ 
 \addlinespace
 $T$ & The final time \\ 
 \addlinespace
 $\Delta t$ & A time step \\ 
\addlinespace
 $p$ & An index of input data between $1$ and $P$ \\ 
 \addlinespace
 $N$ & The number of basis functions \\ 
\addlinespace
 $Q$ & The number of models \\ 
\addlinespace
 $q$ & An index of a model between $1$ and $Q$ \\ 
 \addlinespace
  $\mathcal{G}_q$ & A representation of the model of $q$th \\
  \addlinespace
 $R$ & The number of time steps on a time segment \\ 
 \addlinespace
 $q$ & An index of a time step between $1$ and $R$ \\ 
 \addlinespace
 $\mathcal{F}$ & Discrete Fourier transform (DFT) \\ 
  \addlinespace
  $\mathcal{F}^{-1}$ & Inverse discrete Fourier transform (IDFT) \\ 
 \hline
\end{tabular}
\end{center}
\end{table}
\section{Input data generation}\label{input_generate}
In all experiments except CDE\eqref{DCE}, input functions were randomly generated by mean-zero Gaussian random fields (GRF) following Normal distributions as below \cite{williams2006gaussian,PIDON,PINO} : 
\begin{center}
\begin{tabular}{c|c c c cc } 
 \hline
cases & Type of inputs&$P_{train}$&$P_{test}$&$N$ &Distribution  \\ 
 \hline\vspace{2mm}
  Diffusion-reactioin\eqref{DRE} &Forcing functions&2000&2000&50&$\mathcal{N}(0,25^2)$\\ 
\vspace{2mm}
 Viscous Burgers\eqref{VBE} &initial conditions&2000&2000&32&$\mathcal{N}(0,25^2 (-\Delta + 5^2 I)^{-2})$ \\ 
\vspace{2mm}
Advection\eqref{CE}&Variable coefficients&2000&2000&32& $\mathcal{N}(0,30^2 (-\Delta + 8^2 I)^{-2})$\\
\vspace{2mm}
Convection-diffusion\eqref{DCE} &initial conditions&2000&2000&32& uniform distribution on $[0,1)$ \\ 
\vspace{2mm}
 2D Kuramoto
Sivashinsky\eqref{KSE} &initial conditions&3000&1000&30&$\mathcal{N}(0,4^2 (-\Delta + 2^2 I)^{-5/2})$  \\ 
\vspace{2mm}
  2D Navier-Stokes\eqref{NSE} &initial conditions&2000&1000&32&$\mathcal{N}(0,9^2 (-\Delta + 3^2 I)^{-5/2})$   \\ 
\hline
\end{tabular}
\end{center}
There are two kinds of input data: the first one is for training denoted by $P_{train}$, and the other is for test denoted by $P_{test}$. The size of input data for one dimensional domain is $P\times N$, and the one is $P\times N\times N$ for two dimensional domain.

Only for DCE \eqref{DCE}, we drew the coefficients $a_{j,p}$ in \eqref{DCE_input} from an uniform distribution on $[0,1)$. Accordingly, the size of input data is $P\times N$. For more detail on the input data, refer to \eqref{DCE_input}.

\section{Network architecture and hyper-parameter settings}
In all examples provided in this paper, networks were convolutional neural networks (CNN) equipped with Swish activation function. Regrading an optimizer,  Limited-memory BFGS (L-BFGS) was employed. The hyper-parameters used in each examples are articulated in the table. We note that due to the definition of the notations, $T=\Delta t QR$. and $t_q-t_{q-1}=\Delta t R$. 
\begin{center}
\begin{tabular}{c|c c   c c c c c c } 
 \hline
cases & BC &Basis&$T$&$\Delta t$&$Q$&$R$&Width &Depth \\ 
 \hline\vspace{2mm}
  viscous Burgers\eqref{VBE} &periodic& Fourier&1&0.01&10&10&32& 3 \\ 
\vspace{2mm}
 Diffusion-reaction\eqref{DRE} &Dirichlet& Legendre type&1&0.01&10&10&50&5\\ 
\vspace{2mm}
Advection\eqref{CE}&periodic&Fourier&1&0.01&10&10&32& 5 \\
\vspace{2mm}
 Convection-diffusion\eqref{DCE}&Dirichlet& Legendre type&1&0.01&10&10&32& 5 \\
\vspace{2mm}
 2D Kuramoto
Sivashinsky\eqref{KSE} &periodic&Fourier&0.5&0.01&10& 5 &$30^2$ &1 \\ 
\vspace{2mm}
  2D Navier-Stokes\eqref{NSE} &periodic&Fourier&0.5&0.01&10& 5 &$32^2$&1  \\ 
\hline
\end{tabular}
\end{center}
When training PIDoN and PINO to create reference data, hyper-parameters were employed as in \cite{PIDON} and \cite{PINO}, respectively. 

\section{Performance metrics}\label{metrics}
So as to measure errors, we employ three kinds of metrics as follows (see table\ref{tab:1}). The first one is the mean absolute error (MAE) defined by
\begin{align}\label{MAE}
MAE(u,\widehat{u}):= \frac{1}{RPN}\sum_{p=1}^{P}\sum_{r=1}^{R}\sum_{n=0}^{N+1}|u^{r}_p(x_n)-\widehat{u}^{r}_p(x_n)|,
\end{align}
where $R$ is the number of time steps, $P$ is the number of input data, and $N$ is the number of basis functions (for more details on the notation, refer to table \ref{notation}).

The next is relative $L^2$ norm written as
\begin{align}\label{rel_l2}
Rel.L^2(u,\widehat{u}):=  \frac{1}{P}\sum_{p=1}^{P}\sqrt{\frac{\sum_{r=1}^{R}\sum_{n=0}^{N+1}|u^{r}_p(x_n)-\widehat{u}^{r}_p(x_n)|^2}{\sum_{r=1}^{R}\sum_{n=0}^{N+1}|u^{r}_p(x_n)|^2}}.
\end{align}

The last is $L^\infty$ norm defined by
\begin{align}\label{Linfty}
L^\infty(u,\widehat{u}):= \frac{1}{RP}\sum_{p=1}^{P}\sum_{r=1}^{R}\max_{x_n\in\Omega}|u^{r}_p(x_n)-\widehat{u}^{r}_p(x_n)|.
\end{align}

\end{document}